\documentclass{article}

\PassOptionsToPackage{numbers, compress}{natbib}

\usepackage[preprint]{neurips_2026}

\usepackage[utf8]{inputenc} 
\usepackage[T1]{fontenc}    
\usepackage[colorlinks=true,linkcolor={blue!70!black},citecolor={blue!70!black},urlcolor={blue!70!black},hypertexnames=false]{hyperref}
\usepackage{url}            
\usepackage{booktabs}       
\usepackage{amsfonts}       
\usepackage{nicefrac}       
\usepackage{microtype}      
\usepackage{xcolor}         
\usepackage{graphicx}       
\usepackage{subcaption}     
\usepackage{amsmath,amssymb,amsthm} 
\usepackage{algorithm}
\usepackage{algorithmic}
\usepackage{multirow}

\theoremstyle{definition}

\theoremstyle{remark}

\title{Training-Free Generation of Protein Sequences from Small Family Alignments via Stochastic Attention}

\author{%
  Jeffrey D. Varner \\
  R.F. Smith School of Chemical and Biomolecular Engineering\\
  Cornell University, Ithaca, NY 14850 \\
  \texttt{jdv27@cornell.edu} \\
}

\begin{document}

\maketitle

\begin{abstract}
Generating novel protein sequences that respect a family's statistical constraints typically requires training deep generative models on thousands to millions of examples. Yet most protein families are small: the median Pfam seed alignment contains only 22 sequences, a regime where learned models overfit or collapse. We propose \emph{stochastic attention} (SA), a training-free sampler that treats the modern Hopfield energy over stored sequences as a Boltzmann distribution and draws samples via Langevin dynamics. The score function is the residual of a single softmax attention operation, eliminating the need for a trained score network, pretraining data, or graphics processing units (GPUs). Across eight Pfam families spanning 37 to 420 sequences and 23 to 262 residues, SA generates sequences with low composition divergence, novelty, and structural plausibility supported by ESMFold and AlphaFold2. Compared with profile hidden Markov models (HMMs), EvoDiff, and the multiple sequence alignment (MSA) Transformer, SA is the only tested method to simultaneously achieve low composition divergence, genuine novelty, and sequence identity within each family's nearest-neighbor identity range; the others drift outside this range or produce near-copies. The critical inverse temperature is predicted from principal component analysis (PCA) dimensionality alone, enabling fully automatic operation from a seed alignment. In two domains with deep mutational scanning data, SA-generated substitutions are enriched for experimentally tolerated mutations beyond a position-matched null, and an independent language model (ESM2-650M) scores them within the natural range. Stochastic attention thus opens training-free sequence generation to the long tail of protein families too small for deep learning.
\end{abstract}

\bigskip
\noindent\fbox{\parbox{\textwidth}{\small\textbf{Significance Statement.}
Most protein families are small, with a median of only 22 sequences in their Pfam seed alignments, too few to train a deep generative model. We show that a single attention operation over a family alignment defines an energy landscape whose Boltzmann distribution can be sampled by Langevin dynamics without model training, GPUs, or external data. The resulting method, stochastic attention, generates novel protein sequences that preserve family-level amino acid statistics and are predicted by independent structure predictors to adopt the target-family fold across eight families. A linear relationship between representation dimensionality and critical temperature enables fully automatic operation from a seed alignment, opening training-free generation to the many protein families too small for deep learning.}}
\bigskip

\section{Introduction}

Deep learning has transformed single-chain protein structure prediction, most visibly through AlphaFold2~\cite{jumperHighlyAccurateProtein2021} and RoseTTAFold~\cite{baekRoseTTAFold2021}. The frontier has since shifted toward \emph{designing} novel protein sequences with desired properties~\cite{kuhlmanBradleyAdvances2019,huangComingAgeDeNovo2016}. Generative approaches to this problem now range from classical profile hidden Markov models (pHMMs)~\cite{hmmer3,kroghHiddenMarkovModels1994} and Potts models~\cite{morcosDirectCouplingAnalysis2011,ekeberg2013improved,trinquierEfficientGenerativeModeling2021} to deep generative models including variational autoencoders~\cite{kingmaWellingAutoEncoding2014,riesselman2018deep}, autoregressive language models~\cite{ferruzProtGPT2DeepUnsupervised2022,madaniProGenLanguageModeling2023}, alignment-processing transformers~\cite{raoMSATransformer2021}, MSA-conditioned diffusion~\cite{alamdariProteinGeneration2023}, and structure-conditioned methods such as ProteinMPNN~\cite{dauparasProteinMPNN2022} and RFdiffusion~\cite{watsonRFdiffusion2023}. Yet all sequence-level approaches share a common requirement: large training corpora. Language models are pretrained on millions of UniRef50 sequences; even MSA-conditioned methods require either extensive pretraining or large input alignments. This creates a gap for small families. The median Pfam seed alignment, the curated set of representative sequences that alignment-based methods take as input, contains only 22 sequences. Nearly half of all families have fewer than 20 sequences (Appendix~\ref{si:pfam-census}), below the minimum viable training set for any deep generative architecture. In this scarce-data regime, a few dozen sequences cannot constrain the thousands to millions of parameters in a deep generative model, while pHMMs can emit sequences with low identity to the source family. What is needed is an approach that can extract the statistical structure of a protein family from whatever data is available, even a few dozen sequences, without learning parameters that could lead to overfitting.

The connection between attention and associative memory supplies this approach. Ramsauer et al.~\cite{ramsauerHopfieldNetworksAll2021} showed that a single softmax attention operation is equivalent to one step of the retrieval dynamics on the modern Hopfield energy, building on the dense associative memory framework of Krotov and Hopfield~\cite{krotovDenseAssociativeMemory2016}. This connection endows any set of stored examples with a well-defined energy landscape: the examples become memory patterns, and the Boltzmann distribution over this landscape assigns high probability to states that resemble the stored set. The energy itself has a simple structure. A quadratic term pulls a candidate state toward the origin, while a competing log-sum-exp term over its similarities to the stored patterns pulls it toward whichever memories it most resembles. An inverse temperature controls how sharply the nearest pattern dominates. The gradient of the log-sum-exp is precisely the softmax-weighted combination of stored patterns, i.e., a single attention operation. The score function of the Boltzmann distribution, the gradient that guides sampling toward higher-probability states, is therefore the \emph{residual} of attention: the difference between the attention output and the current state. Diffusion-based generative models such as RFdiffusion~\cite{watsonRFdiffusion2023} obtain such a gradient by training a neural network on data; here it is instead exact and available in closed form, so Langevin sampling can proceed without learning any parameters or approximating gradients. We build on this result to propose \emph{stochastic attention} (SA), a training-free protein sequence generator that combines the exact Hopfield score function with Langevin dynamics. Our prior work~\cite{alswaidanVarnerStochasticAttention2026} introduced SA and validated it on synthetic data, MNIST, and a single protein family. The present study extends that proof of concept from a single illustrative family to a systematic evaluation across diverse families, adding the scaling analysis, baseline benchmarking, and biological validation that the earlier work did not attempt.

In this study, we explored whether stochastic attention can serve as a practical generative model for small protein families. We systematically evaluated SA across eight diverse Pfam families spanning a ten-fold range of family sizes and a seven-fold range of sequence lengths, assessing compositional fidelity, sequence novelty, and structural plausibility using both ESMFold and AlphaFold2. We characterized how generation quality scaled with family size, uncovering an empirical relationship that predicts the critical temperature from alignment statistics alone and eliminates the need for a temperature sweep. We tested robustness in a scarce-data regime down to $K{=}20$ sequences per family and compared SA against four established baselines (profile HMMs, EvoDiff, MSA Transformer, and a Potts model). We also examined whether the generated sequences preserved the pairwise residue covariation that reflects structural contacts and functional couplings. Finally, we sought independent validation of biological plausibility through ESM2-650M pseudo-perplexity scoring and cross-referencing against published deep mutational scanning datasets. Across all eight families, SA generated novel sequences that preserved family-level amino acid statistics and were predicted by both ESMFold and AlphaFold2 to adopt the target-family fold, a combination not matched by the learned baselines tested here. Together these results indicate that the statistical structure of a protein family can itself support sequence generation, without iterative parameter optimization, backpropagation, or external pretraining, when the sampler respects the geometry of the family's sequence space.

\section{Results}

\begin{figure}[p]
\centering
\includegraphics[width=\textwidth]{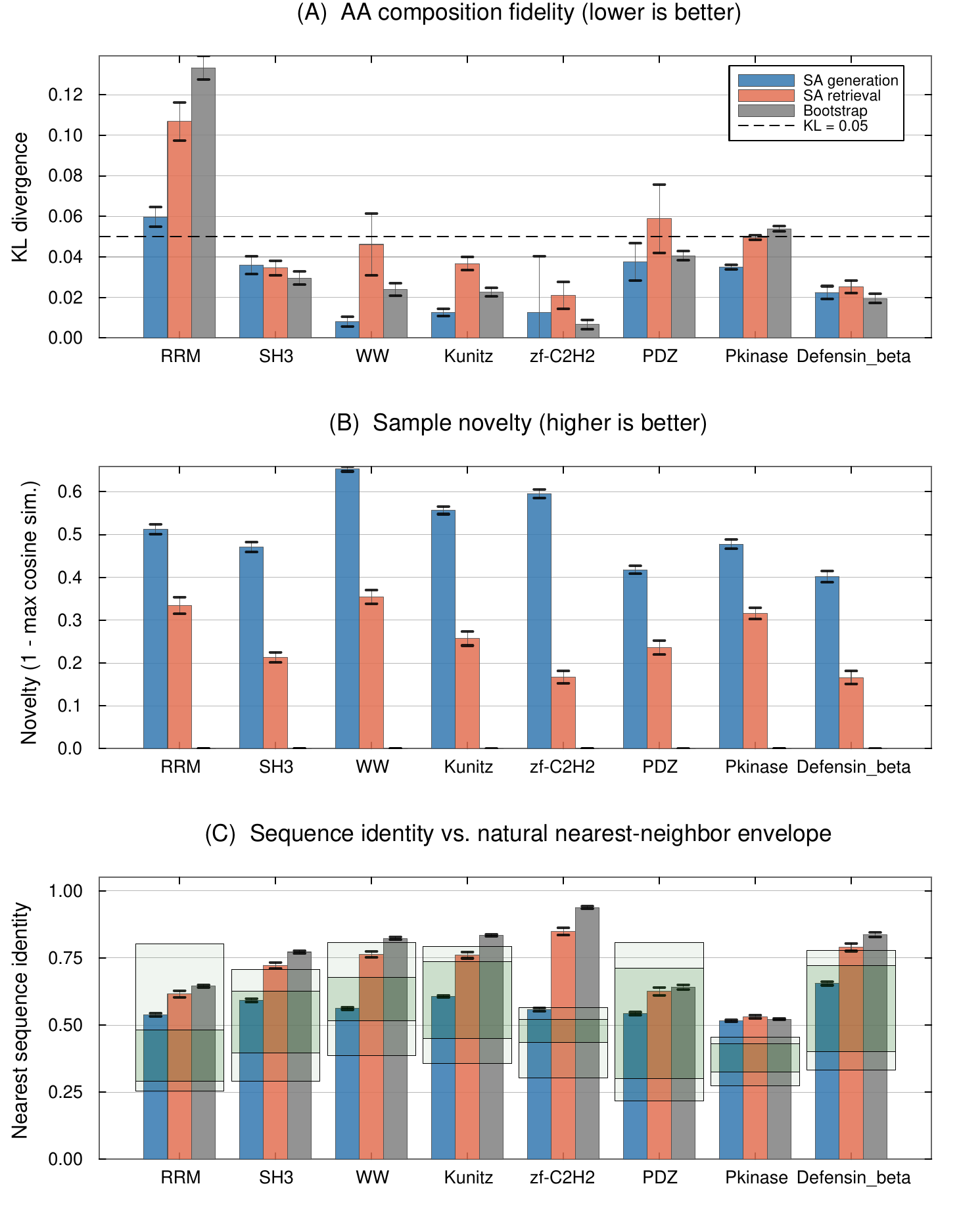}
\caption{Cross-family comparison of SA generation, SA retrieval, and bootstrap across eight Pfam families. \textbf{(A)}~KL divergence of amino acid composition (lower is better); dashed line at KL${=}0.05$. \textbf{(B)}~Novelty in PCA space ($1 - \max_k \cos(\hat{\xi}, \mathbf{m}_k)$; higher is better). Bootstrap samples have zero novelty by construction. \textbf{(C)}~Nearest sequence identity to a stored pattern; the green boxes show each family's natural within-family nearest-neighbor identity envelope (dark: 10th--90th percentile; light: full min--max range; SI Appendix, Table~\ref{tab:nn-band}). The SA-generated identity lies within the envelope in seven of eight families; in Pkinase it rises above the natural maximum. Error bars: $\pm 1$ SE (30 chains $\times$ 5 samples).}
\label{fig:cross-family}
\end{figure}

We ran SA on all eight Pfam families (Table~\ref{tab:families}) and observed that the method consistently generated novel protein sequences while preserving family-level amino acid statistics, across a ten-fold range of family sizes and a seven-fold range of sequence lengths (Fig.~\ref{fig:cross-family}; SI Appendix, Table~\ref{tab:results}). We first assessed sampler accuracy by comparing the outputs of the unadjusted Langevin algorithm (ULA) and its Metropolis-adjusted variant (MALA): acceptance rates exceeded 99.6\% in every family. The Metropolis correction rejected fewer than 0.4\% of proposed moves at this step size (SI Appendix, Table~\ref{tab:sampling-diag}). Thus, the unadjusted and adjusted updates rarely differed in this regime, and we adopted the computationally cheaper ULA in all subsequent experiments. Generated sequences were not near-duplicates of stored patterns: zero sequences across all eight families exceeded 90\% identity to any stored sequence (SI Appendix, Table~\ref{tab:identity-dist}), and the generated ensemble was insensitive to initialization strategy (SI Appendix, Table~\ref{tab:random-init}). In the generation regime, the sampler operated at inverse temperature $\beta \approx 2\beta^*$, where the critical inverse temperature $\beta^*$ is the value at which the energy landscape changes character: below $\beta^*$ the stored patterns blur into a single broad well, while above it the landscape resolves into distinct basins around individual patterns (SI Appendix, Section~\ref{si:beta-star-theory}). At $\beta \approx 2\beta^*$, the generated amino acid composition closely matched each natural family's: the Kullback--Leibler (KL) divergence between the generated and natural amino acid compositions, which falls to zero only when two distributions are identical, stayed at or below $0.06$ in every family and below $0.02$ in three of them. At the same time, novelty ranged from $0.40$ (Defensin\_beta) to $0.65$ (WW), indicating departure from any single stored memory. All generated sequences were distinct in every family, with mean pairwise sequence identity of $35$--$65\%$ (equivalently, mean pairwise Hamming distance $0.36$--$0.65$; SI Appendix, Table~\ref{tab:nn-band}). Thus, the sampler produced a diverse ensemble rather than collapsing to a single mode. Nearest sequence identity to a stored pattern was $51$--$66\%$, within each family's empirical within-family nearest-neighbor identity envelope (each natural sequence's identity to its closest neighbor in the same family; SI Appendix, Table~\ref{tab:nn-band}) in seven of eight families. In Pkinase, generated identity ($0.515$) exceeded the maximum nearest-neighbor identity among natural Pkinase sequences ($0.454$), consistent with that family's consensus-proximal generation. Together, generated sequences remained within the observed sequence-similarity envelope while departing from stored patterns. By contrast, bootstrap samples had zero novelty by construction and Gaussian perturbation produced negligible novelty ($<0.02$), indicating that small additive noise is insufficient to escape the basin of the nearest stored pattern.

Position-specific analysis of the Kunitz domain revealed why SA preserves family statistics (SI Appendix, Fig.~\ref{fig:sequence-analysis}). Sequences generated by stochastic attention preserved all 11 strongly conserved positions of the 53-column seed alignment (those exceeding 90\% conservation), including all six cysteines forming the three disulfide bonds essential to the Kunitz fold, while introducing 17--25 substitutions concentrated at the variable positions. Per-position Shannon entropy correlated between the SA-generated ensemble and the stored alignment (Pearson $r{=}0.968$). Thus, the method identified which positions tolerate variation and which do not. This behavior followed from the Boltzmann sampling mechanism: at strongly conserved positions, all stored patterns agreed and the softmax weights concentrated probability on the consensus residue; at variable positions, stored patterns diverged and the sampler drew from the local distribution of tolerated amino acids.

Subsampling the WW domain at $K \in \{20, 50, 100, 200, 400\}$ with five independent replicates per condition characterized how generation quality changed with decreasing family size (SI Appendix, Fig.~\ref{fig:scaling}). Stochastic attention maintained low KL divergence across the entire range ($0.019 \pm 0.008$ at $K{=}20$, improving to $0.010 \pm 0.002$ at $K{=}400$), while the PCA dimensionality grew from $d{=}17$--$18$ components at $K{=}20$ to $d{=}182$--$183$ at $K{=}400$ and the critical inverse temperature tracked this growth ($\beta^*{=}2.7$ to $5.5$). Despite the roughly ten-fold increase in effective dimensionality, compositional fidelity remained stable, suggesting that the entropy inflection criterion adapted the operating temperature to the geometry of the memory matrix at each scale. Novelty increased with $K$ (from $0.47$ to $0.74$), reflecting the richer landscape of a larger memory set. Nearest sequence identity decreased correspondingly (from $0.71$ to $0.56$), indicating that the sampler explored further from stored patterns when more patterns were available to collectively shape the energy landscape. Replication across three additional families at $K{=}20$ (SH3, Kunitz, zf-C2H2) confirmed that this robustness is not family-specific (SI Appendix, Table~\ref{tab:multifamily-k20}).

\begin{figure}[p]
\centering
\includegraphics[width=\textwidth]{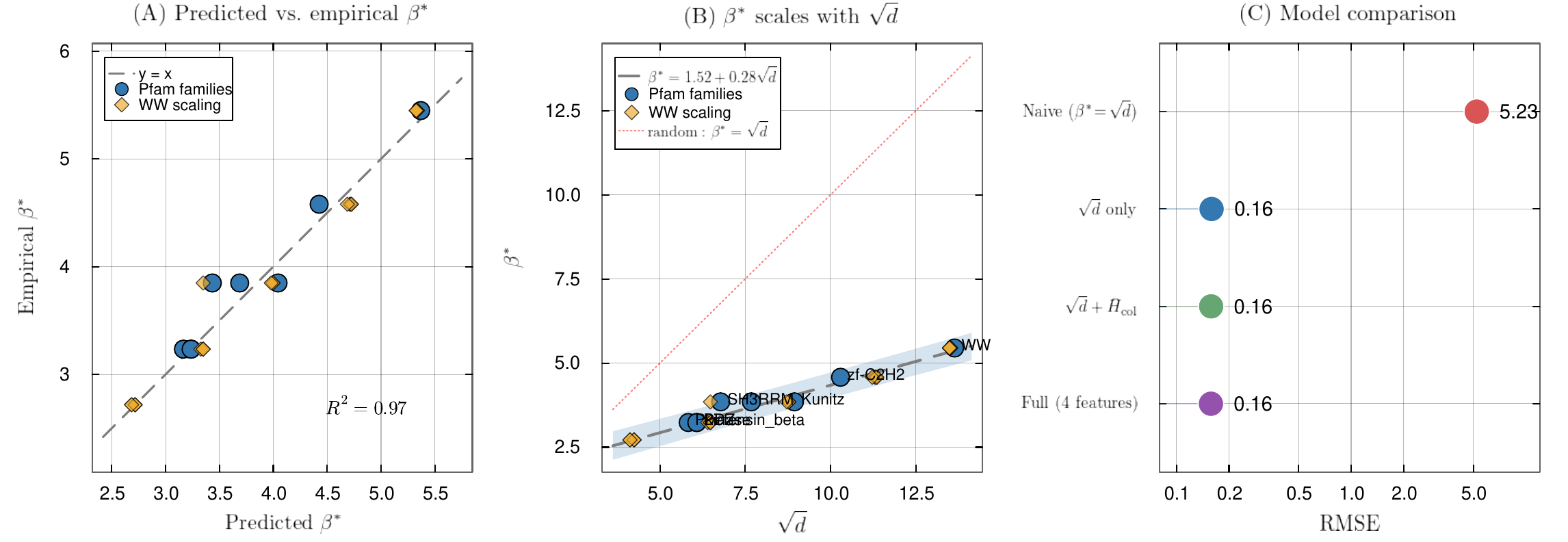}
\caption{Predicting the critical temperature from MSA statistics. \textbf{(A)}~Predicted vs.\ empirical $\beta^*$ for the simple linear model $\beta^* \approx 1.52 + 0.28\sqrt{d}$ ($R^2{=}0.97$, $n{=}33$). Blue circles: eight Pfam families; gold diamonds: WW domain scaling replicates ($K \in \{20,\ldots,400\}$, five repeats each). \textbf{(B)}~$\beta^*$ as a function of $\sqrt{d}$, with the fitted regression (dashed) and the unit-coefficient reference $\beta^* = \sqrt{d}$ (dotted red). The fitted intercept and reduced slope reflect structured correlations in real protein families. \textbf{(C)}~Model comparison (log-scale RMSE). The fitted two-parameter model (RMSE${=}0.16$) is $33\times$ more accurate than the unit-coefficient reference $\beta^* = \sqrt{d}$ (RMSE${=}5.23$); adding MSA features provides negligible improvement.}
\label{fig:beta-prediction}
\end{figure}

The scaling study revealed a consistent pattern in how the sampler self-tunes: the more independent variation a family contained, the higher the inverse temperature the sampler selected for it. The critical inverse temperature $\beta^*$ tracked the family's PCA dimensionality $d$, the number of principal components needed to capture 95\% of its sequence variation. This prompted us to ask whether $\beta^*$ could be predicted from alignment statistics alone, without a temperature sweep (Fig.~\ref{fig:beta-prediction}). A two-parameter model linear in $\sqrt{d}$ predicted $\beta^*$ from PCA dimensionality alone. Fitted on the eight independent Pfam families, it accounted for $95\%$ of the variance in $\beta^*$ ($R^2{=}0.95$) and generalized under leave-one-family-out cross-validation (SI Appendix, Section~\ref{si:beta-star-theory}). Pooling all 33 data points (the 8 families and 25 WW scaling replicates) gave a nearly identical fit, $\beta^* \approx 1.52 + 0.28\sqrt{d}$ (bootstrap standard error [SE]: intercept $\pm 0.07$, slope $\pm 0.007$; $R^2{=}0.97$). The square-root dependence on $d$ follows from concentration of measure on the unit sphere $\mathbb{S}^{d-1}$: the similarity between a random probe and the stored patterns scales as $\mathcal{O}(1/\sqrt{d})$, so the inverse temperature needed to resolve those similarities grows as $\sqrt{d}$. The fitted slope, however, was smaller than geometry alone predicts: $0.28$, against the ${\approx}1.6$ expected for random patterns under a Gaussian random-score model. That gap is a signature of real sequence structure. Each stored pattern has unit inner product with itself but lower similarity to the other patterns (a gap of $0.55$--$0.82$ to its next-nearest pattern), so when the patterns themselves serve as probes the softmax concentrates on the self-pattern at a lower $\beta$ than the random-score model predicts (SI Appendix, Section~\ref{si:beta-star-theory}). Adding further predictors (mean column entropy, effective number of sequences, spectral concentration) produced negligible improvement ($\Delta R^2 < 0.001$), indicating that PCA dimensionality captured most of the relevant variation. This regression provides a practical shortcut: for a new family, $\beta^*$ can be estimated from its PCA dimensionality and used to set the generation temperature without a sweep.

\begin{figure}[p]
\centering
\includegraphics[width=\textwidth]{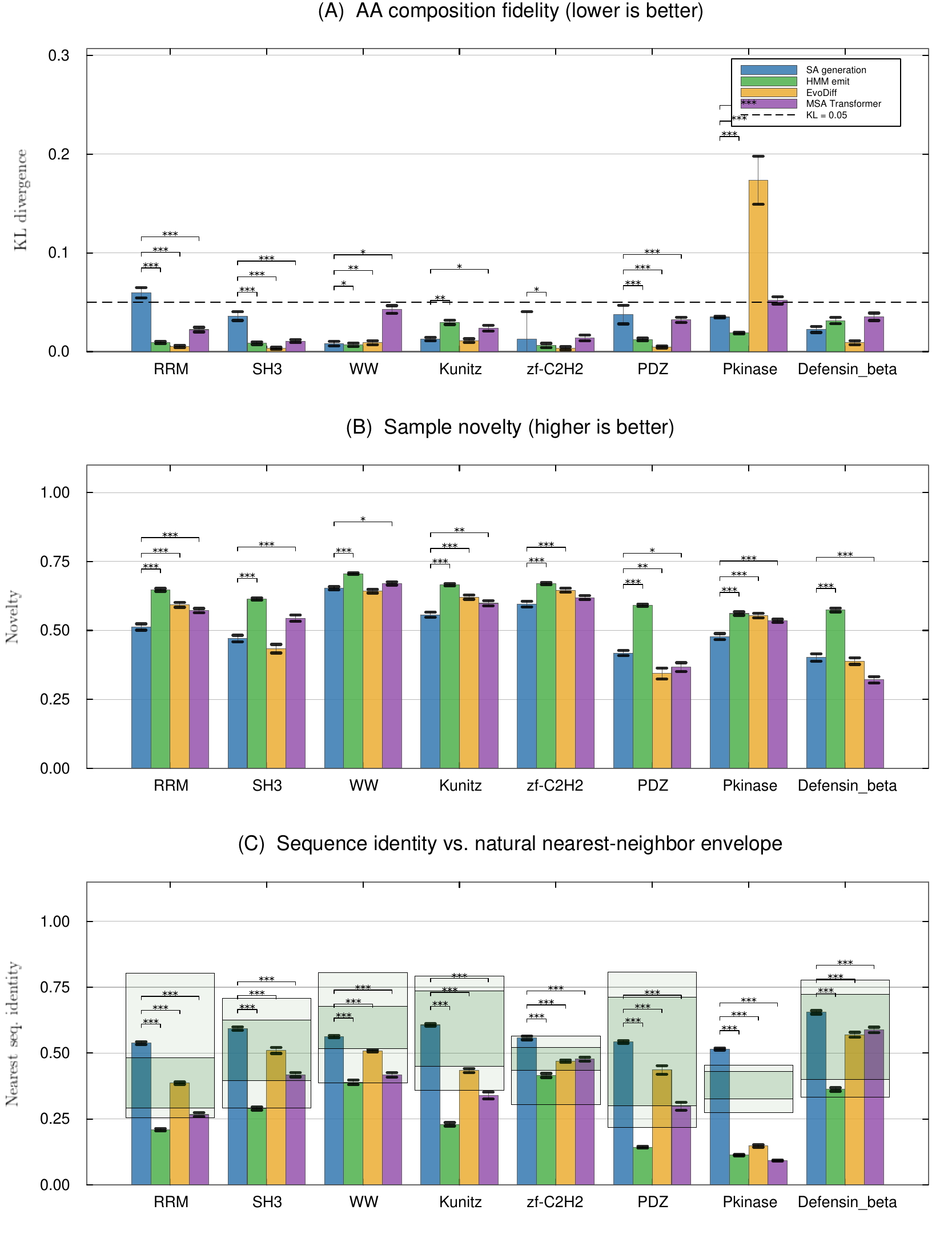}
\caption{Head-to-head comparison of SA generation against three learned baselines across all eight Pfam families. \textbf{(A)}~KL divergence of amino acid composition (bootstrap SE; $n_{\mathrm{boot}}{=}1000$); dashed line at KL${=}0.05$. \textbf{(B)}~Novelty in PCA space. \textbf{(C)}~Nearest sequence identity to a stored pattern; the green boxes show each family's natural within-family nearest-neighbor identity envelope (dark: 10th--90th percentile; light: full min--max range; SI Appendix, Table~\ref{tab:nn-band}). Panels (B) and (C): per-chain means $\pm 1$ SE. Significance brackets: Wilcoxon rank-sum tests ($^{***}p < 0.001$, $^{**}p < 0.01$, $^{*}p < 0.05$). Per-family numerical values are reported in SI Appendix for the learned baselines (Table~\ref{tab:baseline-metrics}) and for SA generation and the reference conditions (Table~\ref{tab:results}).}
\label{fig:baseline-comparison}
\end{figure}

Having characterized how SA self-tunes in isolation, we next compared it against three established learned baselines, profile HMM emit, EvoDiff (MSA-conditioned diffusion), and the MSA Transformer (Gibbs masked language model sampling), across all eight families (Fig.~\ref{fig:baseline-comparison}); a physics-based Potts model was compared separately. Profile HMM emit and the MSA Transformer achieved high novelty scores, but their nearest sequence identity to stored patterns fell to $9$--$59\%$ across the eight families. Because the identity reported for each generated sequence is a nearest-neighbor quantity, the comparable natural quantity is the within-family nearest-neighbor identity, each natural sequence's identity to its closest neighbor in the same family. By this measure, the drifted baselines fell below the lowest nearest-neighbor identity of any natural sequence: profile HMM emit reached $11$--$21\%$ in Pkinase, PDZ, and RRM, whereas no natural sequence in those families is less than $22$--$28\%$ identical to its nearest neighbor (SI Appendix, Table~\ref{tab:nn-band}). Each family's natural within-family nearest-neighbor identity envelope is shown in green (Fig.~\ref{fig:baseline-comparison}C); a sequence falling below its family's box may have moved outside the family, whereas one rising above it is a near-copy of stored patterns with insufficient novelty. EvoDiff balanced the metrics more effectively in shorter families but required pretraining on millions of UniRef50 MSAs and scaled poorly with sequence length (${\sim}14$ hours for 150 Pkinase sequences on CPU, though GPU inference would reduce this, versus seconds for SA on the same hardware; SI Appendix, Table~\ref{tab:timing}). Taken together, none of the three learned baselines combined compositional fidelity, novelty, and family-consistent sequence identity: the profile HMM and MSA Transformer fell outside the family's identity range, while EvoDiff achieved the balance only in shorter families and only with large-scale pretraining.

Stochastic attention simultaneously achieved compositional fidelity, novelty, and family-consistent sequence identity (Fig.~\ref{fig:baseline-comparison}). It held KL divergence below $0.05$ in seven of eight families. Its novelty of $0.40$--$0.65$ indicated departure from stored patterns. Sequence identity stayed at $51$--$66\%$, within each family's empirical within-family nearest-neighbor identity envelope for seven of eight families, with Pkinase marginally above (SI Appendix, Table~\ref{tab:nn-band}). This combination (compositionally faithful, novel, and within the family's sequence-similarity envelope) distinguished SA from the baselines tested (which differ from SA in architecture and training requirements). A physics-based model that fits that structure explicitly gives a sharper test. We therefore compared SA against a Potts model fitted via pseudo-likelihood maximization (plmDCA) across all eight families (SI Appendix, Table~\ref{tab:potts}). Potts models explicitly fit pairwise coupling parameters $J_{ij}(a,b)$ between all position pairs, giving them direct access to coevolutionary signals. The comparison between the Potts model and SA revealed comparable compositional fidelity and output quality in six of eight families, with SA achieving higher novelty in five of eight. However, the Potts model showed limitations in the most data-starved families. In Pkinase its novelty was $0.22$, against $0.48$ for SA, showing the Gibbs sampler remained near stored patterns. In Defensin\_beta the Potts method KL of $0.112$ was the highest of any method (consistent with its high parameter count). The number of coupling parameters scales as $L(L{-}1)q^2/2$ for alignment length $L$ and alphabet size $q{=}20$, yielding data-to-parameter ratios as low as $2.5 \times 10^{-6}$. That SA achieved comparable or better output without fitting any parameters suggests that the Hopfield energy implicitly captures some of the correlational structure that pairwise coevolutionary models fit explicitly, but through the softmax attention operation rather than through parameter estimation.

Pairwise mutual information (MI) analysis tested whether SA preserved residue covariation, which reflects structural contacts and functional couplings that single-site statistics cannot capture~\cite{morcosDirectCouplingAnalysis2011} (SI Appendix, Table~\ref{tab:mi}; Figs.~\ref{fig:mi-scatter}--\ref{fig:mi-barplot}). Pearson correlations between stored and SA-generated MI vectors ranged from $r{=}0.53$ to $0.92$ across seven of eight families, with Pkinase ($r{=}0.05$) the lone exception. The Pkinase alignment was shallow, only $K{=}37$ sequences over $L{=}262$ positions, which could in principle make its covariation hard to estimate. However, this was likely not the case because the Potts model recovered the covariation on the identical alignment ($r{=}0.95$). The failure lay instead in SA generation, which in Pkinase stayed closer to the family consensus than in any other family (mean identity to the consensus $0.73$, the highest observed; SI Appendix, Section~\ref{si:consensus-reg}). Because near-consensus sequences vary little from one another, they carried almost none of the position-to-position covariation that MI measures. Thus, stochastic attention preserved covariation in the compact families but not in the longest, most data-starved one. Further, stochastic attention exceeded profile HMMs (whose MI correlations spanned $r{=}0.07$--$0.83$) in seven of eight families. Ranked by covariation fidelity, the four methods followed the order HMM $<$ SA $<$ Potts $\leq$ EvoDiff in three of eight families. This order tracked how explicitly each method represents covariation: profile HMMs model only position-independent frequencies, stochastic attention captures couplings implicitly through the Hopfield energy, Potts models fit pairwise couplings explicitly, and EvoDiff adds deep coevolutionary priors from large-scale pretraining. That stochastic attention outranked the position-independent baseline suggested the Hopfield energy captured correlational structure single-site models cannot encode, without fitting any pairwise parameters.

\begin{figure}[p]
\centering
\includegraphics[width=\textwidth]{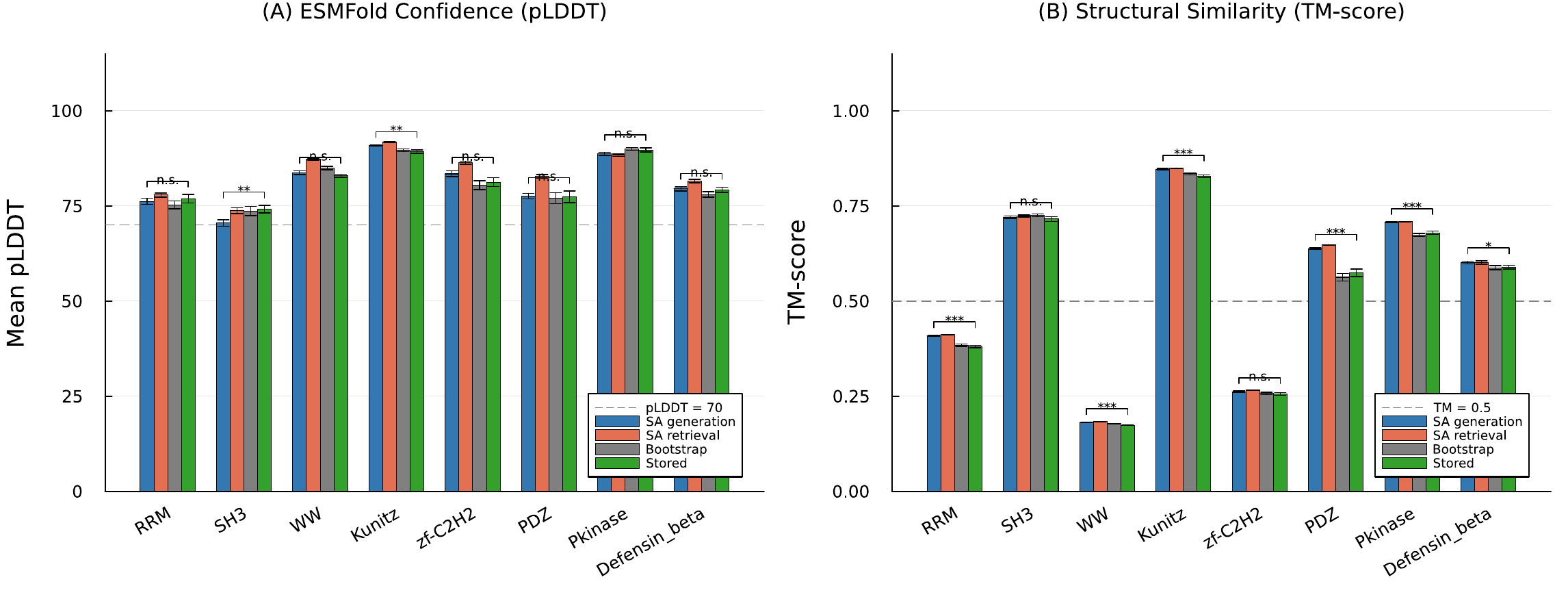}
\caption{Structure validation using ESMFold. \textbf{(A)}~Mean pLDDT across all eight families for SA generation, SA retrieval, bootstrap, and stored sequences; dashed line at pLDDT${=}70$. \textbf{(B)}~TM-score to the experimentally determined reference structure; dashed line at TM${=}0.5$. Significance: Wilcoxon rank-sum tests ($^{***}p < 0.001$, $^{**}p < 0.01$, $^{*}p < 0.05$, n.s.\ = not significant). Error bars: $\pm 1$ SE ($n{=}50$).}
\label{fig:structure-validation}
\end{figure}

We assessed structural plausibility of SA-generated sequences with two independent structure predictors, ESMFold~\cite{linEvolutionaryScaleModeling2023} and AlphaFold2~\cite{jumperHighlyAccurateProtein2021}. We predicted folded structures for 50 sequences per method per family (Fig.~\ref{fig:structure-validation}). We compared SA-generated and stored natural sequences on two structural metrics. For each metric, a Wilcoxon rank-sum test assessed significance and Cohen's $d$ measured the effect size. The first structural metric, the predicted local distance difference test (pLDDT) score, is the predictor's own per-residue confidence that each position was placed correctly, reported from 0 to 100. Under ESMFold, sequences generated by stochastic attention achieved mean pLDDT scores statistically indistinguishable from stored sequences in six of eight families (no significant difference, $p > 0.05$). The pLDDT was significantly higher in Kunitz ($90.8$ vs.\ $89.3$, $p{=}0.001$, $d{=}0.67$). At least 86\% of SA-generated sequences exceeded the pLDDT${=}70$ confidence threshold in seven of eight families, with SH3 the lone exception ($54\%$). The second structural metric, the template modeling (TM) score, ranges from 0 to 1 and rises as a predicted structure superimposes more closely on the reference. In six of eight families, the predicted structures of SA-generated sequences matched the reference significantly better than stored sequences ($p < 0.05$; $d > 1.0$ in RRM, WW, PDZ, and Pkinase). To test whether these elevated TM-scores reflected a predictor preference for consensus-like sequences, we asked whether predicted structural quality tracks consensus proximity within each family. The correlation was weak and inconsistent (per-family Spearman $\rho$ between TM-score and identity to the consensus ranged from $-0.13$ to $0.53$, pooled $\rho{=}0.17$; pLDDT pooled $\rho{=}0.23$; SI Appendix, Section~\ref{si:consensus-reg}). Thus, the elevated TM-scores were likely not an artifact of consensus proximity. As an independent cross-validation, the AlphaFold2 predictions corroborated the ESMFold findings while resolving the single negative result (SI Appendix, Figs.~\ref{fig:structure-esmfold-af2}--\ref{fig:esmfold-af2-concordance}). For SH3, the only family where ESMFold had indicated a pLDDT deficit, AlphaFold2 showed no pLDDT difference ($p{=}0.99$) and significantly higher TM-scores for SA generation ($0.736$ vs.\ $0.721$, $p < 0.001$). For the highest-confidence representative from each family, the AlphaFold2 prediction aligned to the experimentally determined reference with sub-angstrom root-mean-square deviation (Fig.~\ref{fig:structure-gallery}). The concordance between two independent structure predictors across all eight families strengthens the computational evidence that SA-generated sequences are predicted to adopt the target-family fold.

As a complementary, sequence-level assessment of biological plausibility, we scored all SA-generated and stored sequences using ESM2-650M~\cite{linEvolutionaryScaleModeling2023}, a protein language model pretrained on UniRef50 sequences that is entirely independent of SA (SI Appendix, Table~\ref{tab:esm2}). Overall, the model scored SA-generated sequences comparably to natural family members, with mean pseudo-perplexity $5.71$ versus $5.58$ for stored sequences. Lower pseudo-perplexity indicates a sequence the model finds more typical of natural proteins. In five of eight families, SA scored as well as the stored sequences or better: indistinguishable in four (RRM, zf-C2H2, Pkinase, Defensin\_beta; Wilcoxon $p > 0.29$) and significantly \emph{better} in Kunitz ($4.47$ vs.\ $4.98$, $p < 0.001$, $d{=}{-}0.74$). In the other three families (SH3, WW, PDZ), SA scored modestly higher ($d{=}0.60$--$1.03$), though all generated sequences remained within the range of natural proteins. Beyond this language-model scoring, we cross-referenced SA-generated substitutions against published deep mutational scanning (DMS) datasets to ask whether the Boltzmann sampling procedure preferentially generates individually tolerated mutations (SI Appendix, Tables~\ref{tab:esm2} and \ref{tab:dms}). For the PDZ domain (DLG4/PSD-95 PDZ3~\cite{mclaughlinSpatialArchitecture2012}), $91.7\%$ of $4{,}313$ matched substitutions were classified as experimentally tolerated, and for the SH3 domain (GRB2 SH3~\cite{faureDoubleDMS2022}), $90.4\%$ of $2{,}396$ matched substitutions were tolerated. Because SA preferentially substitutes at variable, more tolerant positions, comparing these rates against the overall dataset tolerance rate ($77.5\%$ and $67.5\%$) conflates position selection with residue choice. We therefore used a position-matched null that asks, for each position SA substituted, how often a randomly chosen tested substitution at that same position is tolerated. Against this stricter null, SA-generated substitutions remained significantly enriched in both families, indicating that SA selects function-compatible residues beyond simply targeting tolerant positions. The enrichment held for PDZ ($91.7\%$ vs.\ $84.3\%$, $1.09\times$, $p < 10^{-77}$) and SH3 ($90.4\%$ vs.\ $74.9\%$, $1.21\times$, $p < 10^{-121}$), and a conservation-matched null gave the same result (SI Appendix, Table~\ref{tab:dms-null}). The continuous DMS fitness scores of SA-generated substitutions were likewise significantly higher than those of random single mutations (Wilcoxon $p < 10^{-20}$; SI Appendix, Tables~\ref{tab:esm2} and \ref{tab:dms}). Taken together, these results indicate that the Boltzmann sampling procedure preferentially generates substitutions that are individually compatible with function in DMS assays, choosing both tolerant positions and tolerated residues, without any explicit fitness objective. However, they do not by themselves establish that every full generated sequence is functional, and the DMS evidence is limited to these two domains.

Finally, the beta defensin family (PF00711) provided an opportunity to test whether SA generation preserved not only sequence-level statistics but also the biophysical properties that underpin antimicrobial function~\cite{lazzaroAntimicrobialPeptides2020} (SI Appendix, Fig.~\ref{fig:defensin-biophysics}). Sequences generated by stochastic attention preserved the six-cysteine disulfide scaffold ($6.01 \pm 0.08$ mean cysteines, compared to $6.02 \pm 0.15$ for natural sequences). They maintained a cationic charge profile ($+6.1 \pm 1.6$), with the narrowest charge distribution of any method. They also occupied the same region of charge-hydrophobicity space as natural defensins. By a minimal antimicrobial plausibility filter (net charge $\geq +2$, hydrophobic fraction in $[0.3, 0.7]$, $\geq 4$ cysteines), $51\%$ of SA-generated sequences passed, compared to $42\%$ of stored sequences, $25\%$ of HMM-emitted sequences, and $16\%$ of MSA Transformer sequences. This implicit enforcement of biophysical constraints arose because the energy landscape concentrated probability near the stored patterns, which already satisfied these constraints by virtue of being functional proteins. No explicit objective for charge, hydrophobicity, or disulfide topology was required.

\section{Discussion}

Across eight Pfam families spanning a ten-fold range of sizes ($K{=}37$--$420$) and a seven-fold range of sequence lengths ($L{=}23$--$262$), SA generated novel sequences that preserved family-level amino acid statistics and were predicted by independent structure predictors to adopt the target-family fold. It did so using only the seed alignment, with no training, external data, or GPU resources. None of the learned baselines tested here achieved this combination. Profile HMMs and the MSA Transformer generated sequences with nearest-neighbor identity below the empirical within-family nearest-neighbor envelope. In five families (RRM, SH3, Kunitz, PDZ, and Pkinase), this identity fell below the minimum observed among natural sequences (SI Appendix, Table~\ref{tab:nn-band}). Their high novelty was therefore accompanied by sequence-identity drift rather than family-consistent exploration. EvoDiff required millions of pretraining sequences and hours of compute, and degraded on the longest family. Among the tested baselines, no method simultaneously stayed within the family identity range, maintained low composition divergence, and produced genuine novelty. That a training-free method matched systems built on orders of magnitude more data indicated that the statistical structure of protein families can support generation without iterative parameter optimization, backpropagation, or external pretraining, provided the generative framework respects the geometry of the family's sequence space. Beyond matching their output, SA avoids the pretraining corpora, GPU resources, and large parameter counts that place deep generative methods out of reach for the small, single-laboratory families that make up most of Pfam, requiring only a seed alignment and a standard workstation. The consensus-with-noise and column-permuted alignment controls supported the conclusion that this was not only a consequence of marginal statistics: SA produced a different output profile from isotropic perturbation of the consensus, and exploited inter-position correlations that column-independent models could not capture.

Sequences generated by stochastic attention were predicted to adopt the canonical family fold. In six of eight families, their TM-scores to the reference structure modestly exceeded those of the stored natural sequences under ESMFold (AlphaFold2 reproduced the pattern across all eight families, the two predictors' TM-scores correlating at $r{=}0.999$; SI Appendix, Fig.~\ref{fig:esmfold-af2-concordance}). The pLDDT scores were indistinguishable from stored sequences in six of eight families, supporting the conclusion that the sampler did not sacrifice predicted folding confidence for diversity, and the single exception (SH3) was resolved by AlphaFold2 cross-validation. Two explanations for the elevated TM-scores merit consideration. First, predictor bias: structure predictors trained on evolutionary data may assign higher confidence to consensus-proximal sequences. Second, genuine structural convergence: the Boltzmann ensemble averages over lineage-specific deviations, potentially producing sequences that encode the shared fold more cleanly than individual natural sequences. A direct test of the first explanation is whether predicted structural quality tracks consensus proximity within each family. The correlation was weak and inconsistent (pooled Spearman $\rho{=}0.17$ for TM-score and $0.23$ for pLDDT, each accounting for under $4\%$ of the variance, and negative in three families; SI Appendix, Section~\ref{si:consensus-reg}). This argues against predictor bias as the sole driver, though it does not exclude an absolute preference for family-typical sequences shared by both the sampler and the predictor. We therefore interpret the elevated TM-scores cautiously and do not read them as evidence that SA-generated sequences fold better than natural ones. The two explanations are not mutually exclusive. The decisive computational control, folding the consensus-with-noise ensemble through the same structure predictors, was not performed here and remains for future work. Definitive attribution requires experimental structure determination. The DMS validation provided partial evidence for the latter. If elevated TM-scores reflected only predictor bias, one would not necessarily expect SA substitutions to be enriched for experimentally tolerated mutations. Yet the position-matched enrichment of $1.09$--$1.21\times$ in two independent DMS datasets suggested that the generated substitutions are biased toward locally function-compatible changes. However, all computational validators used here (structure predictors, ESM2-650M, and DMS-derived fitness scores) were trained on evolutionary data, so high scores for family-like sequences are expected. Experimental characterization would provide definitive confirmation.

The WW domain scaling study showed that generation quality was stable down to $K{=}20$ sequences, with KL divergence below $0.02$ and novelty above $0.47$, and replication across three additional families supported the same trend. This robustness arose because the energy landscape is defined entirely by the stored patterns, with no parameters that could overfit to a small training set. The entropy inflection criterion adapted the operating temperature to the memory matrix at each scale, and $\beta^*$ could be predicted from PCA dimensionality alone ($R^2{=}0.97$), enabling fully automatic operation from a seed alignment. The $\sqrt{d}$ scaling of $\beta^*$ is consistent with concentration of measure on the unit sphere: random-probe similarities have variance $1/d$, so the inverse temperature required to resolve $\mathcal{O}(1/\sqrt{d})$ similarity differences grows as $\sqrt{d}$. The reduced coefficient ($0.28$ vs.\ ${\sim}1.6$ for random patterns) reflected self-similarity anchoring in the stored-pattern probes. Each stored pattern has unit inner product with itself but lower similarity to other patterns. The softmax therefore concentrated on the self-pattern at lower $\beta$ than a mean-field model predicts (SI Appendix, Section~\ref{si:beta-star-theory}). By contrast, learned generative models face a parameter-to-data ratio problem when $K < 100$: VAEs and diffusion models have thousands to millions of parameters that cannot be constrained by a few dozen sequences. This regime encompasses most Pfam families.

The connection between attention and Boltzmann sampling extends beyond proteins: any set of examples representable as vectors defines an energy landscape via the modern Hopfield energy, and Langevin dynamics on this landscape provides a generative model with no training. Two conditions determine whether the approach produces useful output. First, the stored examples must be structured enough that the energy landscape has well-defined basins rather than a featureless plateau. Protein families satisfy this because evolutionary constraint creates correlated patterns of conservation and variation. Second, the examples must be diverse enough that the sampler does not trivially retrieve: if all stored patterns are nearly identical, the distribution collapses and generation reduces to consensus with noise. The eight families spanned $22$--$37\%$ mean pairwise identity, diverse enough to avoid consensus collapse yet, through their correlated conservation and variation, structured enough to form well-defined basins. Classical Hopfield networks were limited to storing $\mathcal{O}(d)$ patterns before catastrophic interference~\cite{hopfieldNeuralNetworksPhysical1982,amitStoringInfiniteNumbers1985}. The modern continuous Hopfield energy lifts this limit, storing a number of well-separated patterns that grows exponentially with $d$~\cite{ramsauerHopfieldNetworksAll2021}. This is an advantage for scarce-data generation: with $K \ll d$ stored patterns the family lies below capacity, so the energy landscape is well-separated and the sampler can explore between memories without interference. The same approach may apply to other biological sequence families where data is scarce but examples are structured. Examples include antibody repertoires, where clonotypes have only a handful of somatic variants; RNA families with few seed sequences; and small-molecule libraries with tens of analogs per series. The Hopfield energy also admits a conditioning mechanism: adjusting pattern multiplicities via a scalar bias on the attention logits shifts generation toward a user-specified functional subset, enabling property-directed generation (e.g., toward binding activity) without retraining~\cite{varnerConditioningProtein2026}. The analytic score function makes the sampler amenable to the convergence theory of Langevin dynamics~\cite{durmusNonasymptoticAnalysisUnadjusted2017,dalalyanTheoreticalGuaranteesApproximate2017}. In the strongly regularized, log-concave regime ($\beta\sigma_{\max}^2 < 2$, with $\sigma_{\max}$ the largest singular value of the memory matrix), these results yield formal guarantees on sample quality that are absent from most neural generative models. The generation temperatures used here lie outside that regime ($\beta\sigma_{\max}^2 \gg 2$; SI Appendix, Section~\ref{si:convergence}), where the energy is non-convex and convergence is supported by the empirical mixing diagnostics and initialization-insensitivity analysis rather than by these bounds.

Several open questions remain for SA, each pointing to a direction for future development. The most significant concerns the fixed alignment representation: SA requires a pre-computed MSA and a fixed alignment length $L$, and cannot generate insertions, deletions, or variable-length sequences. This constraint is inherited from the one-hot-plus-PCA encoding, which maps each sequence to a vector of fixed dimension $20L$. Any change in $L$ would alter the representation space itself. In practice, this means the method is best suited to compact, well-aligned domains rather than multi-domain proteins or families with frequent large insertions. Extending stochastic attention to variable-length generation would likely require replacing PCA with a length-agnostic embedding while preserving the closed-form score function, an open direction for future work. The PCA projection itself discards information that may be relevant for capturing higher-order correlations, though a sensitivity analysis confirmed robustness across the 75\%--99\% retained variance range (SI Appendix, Table~\ref{tab:pca-sensitivity}). Relatedly, SA preserved pairwise residue covariation in seven of eight families but largely lost it in Pkinase ($r{=}0.05$), the longest and most data-starved family, where generation was consensus-proximal (SI Appendix, Table~\ref{tab:mi}). Covariation fidelity therefore degrades when the data-to-length ratio is low, and our covariation claims are restricted to the compact families. Further, the eight families studied here are all well-structured, globular domains with abundant structural and functional annotation. This choice was necessary for validation but introduced a selection bias toward the class of proteins that structure predictors handle best. Performance on intrinsically disordered regions, transmembrane domains, multi-domain proteins, or families with low sequence identity remains untested, and the method's reliance on PCA may be limiting for disordered families where sequence variation is less structured. Finally, our multi-chain protocol does not guarantee coverage of all modes in the Boltzmann distribution, particularly for large or diverse families, and wet-lab characterization of selected generated sequences would establish the method's utility for protein engineering applications.

\section{Materials and Methods}

\subsection{From Hopfield Energy to Score Function}
Let $\mathbf{X} = [\mathbf{m}_1, \dots, \mathbf{m}_K] \in \mathbb{R}^{d \times K}$ be the memory matrix whose columns are vectorized protein sequences from a family alignment of $K$ members, each represented in $d$ dimensions (after encoding and, optionally, PCA projection). The modern Hopfield energy~\cite{ramsauerHopfieldNetworksAll2021} is given by:
\begin{equation}\label{eq:hopfield-energy}
    E(\boldsymbol{\xi}) = \tfrac{1}{2}\|\boldsymbol{\xi}\|_2^2 - \tfrac{1}{\beta}\log\sum_{k=1}^{K} \exp\!\bigl(\beta\,\mathbf{m}_k^\top \boldsymbol{\xi}\bigr),
\end{equation}
where $\beta > 0$ is an inverse temperature. The gradient is $\nabla E(\boldsymbol{\xi}) = \boldsymbol{\xi} - \mathbf{T}(\boldsymbol{\xi})$, where
\begin{equation}\label{eq:retrieval-map}
    \mathbf{T}(\boldsymbol{\xi}) := \mathbf{X}\,\operatorname{softmax}\!\bigl(\beta\,\mathbf{X}^\top\boldsymbol{\xi}\bigr)
\end{equation}
is the softmax attention retrieval map~\cite{ramsauerHopfieldNetworksAll2021}. The Boltzmann distribution~\cite{ackeleyLearningAlgorithmBoltzmann1985} $p_\beta(\boldsymbol{\xi}) \propto \exp(-\beta\,E(\boldsymbol{\xi}))$ has score function given by:
\begin{equation}\label{eq:score}
    \nabla \log p_\beta(\boldsymbol{\xi}) = -\beta\,\nabla E(\boldsymbol{\xi}) = \beta\bigl(\mathbf{T}(\boldsymbol{\xi}) - \boldsymbol{\xi}\bigr).
\end{equation}
This score is \emph{exact} and computed by a single attention operation; no score network or training is required.

Applying the unadjusted Langevin algorithm (ULA)~\cite{robertsTweedieExponentialConvergence1996} to the potential $U = \beta E$ with step size $\tilde{\alpha} = \alpha/\beta$ yields the \emph{stochastic attention} update:
\begin{equation}\label{eq:sa-update}
    \boxed{\;
    \boldsymbol{\xi}_{t+1} = (1 - \alpha)\,\boldsymbol{\xi}_t + \alpha\,\mathbf{X}\,\operatorname{softmax}\!\bigl(\beta\,\mathbf{X}^\top\boldsymbol{\xi}_t\bigr) + \sqrt{\tfrac{2\alpha}{\beta}}\;\boldsymbol{\epsilon}_t, \quad \boldsymbol{\epsilon}_t \sim \mathcal{N}(\mathbf{0}, \mathbf{I})
    \;}
\end{equation}
Each step performs three operations: (i) contraction toward the origin, (ii) a softmax-weighted pull toward stored memories, and (iii) isotropic Gaussian noise scaled by $\sqrt{2\alpha/\beta}$. The per-step cost is $\mathcal{O}(dK)$: two matrix-vector products and one softmax. As $\beta \to \infty$ the noise vanishes and the update reduces to deterministic retrieval; as $\beta \to 0$ sampling becomes noise-dominated.

\subsection{Critical Temperature from PCA Dimensionality}\label{sec:beta-star}
The attention entropy $H(\beta) = -\sum_k p_k \log p_k$, where $p_k = \operatorname{softmax}(\beta\,\mathbf{X}^\top\boldsymbol{\xi})_k$, quantifies selectivity: $H = \log K$ when attention is uniform ($\beta \to 0$) and $H \to 0$ when attention concentrates on a single memory ($\beta \to \infty$). The entropy derivative satisfies~\cite{alswaidanVarnerStochasticAttention2026}:
\begin{equation}\label{eq:entropy-deriv}
    \frac{dH}{d\beta} = -\beta\,\mathrm{Var}_{\mathbf{p}}(e),
\end{equation}
where $e_k = \mathbf{m}_k^\top\boldsymbol{\xi}$ are the query-memory similarities. The maximum rate of entropy loss occurs at the inflection point $\beta^*$ satisfying $d^2H/d\beta^2 = 0$, which marks the onset of pattern-specific structure in the energy landscape. The form of $\beta^*$ is constrained by the geometry of the stored patterns. Because the memory columns lie on the unit sphere $\mathbb{S}^{d-1}$, concentration of measure dictates that the similarity variance is $\mathrm{Var}(e_k) = 1/d$ exactly for any query drawn uniformly from $\mathbb{S}^{d-1}$ (Appendix~\ref{si:beta-star-theory}). The characteristic similarity scale is therefore $\sigma = 1/\sqrt{d}$, and since $\beta$ enters the softmax as $\beta e_k$, the inflection must occur when $\beta\sigma = \mathcal{O}(1)$, giving $\beta^* \sim \sqrt{d}$. A separate calculation shows that the softmax entropy over $K$ i.i.d.\ Gaussian scores undergoes its inflection at a dimensionless temperature $\tau^* \approx 1.6$ that is nearly independent of $K$ for $K \geq 30$ (Appendix~\ref{si:beta-star-theory}), providing a universal baseline for the transition.

Empirically, we find a simple linear relationship across eight Pfam families and WW-domain subsamples spanning $K = 20$--$420$ and $d = 18$--$186$ (Results):
\begin{equation}\label{eq:beta-prediction}
    \beta^* \approx 1.52 + 0.28\sqrt{d},
\end{equation}
which captures 97\% of the variance ($R^2 = 0.97$, bootstrap median over $10{,}000$ resamples; see Appendix~\ref{si:beta-star-theory} for the full OLS fit and uncertainty analysis). The Gaussian mean-field prediction $\beta^* = \tau^*\sqrt{d} \approx 1.6\sqrt{d}$ has the correct $\sqrt{d}$ scaling but overshoots the empirical values by a factor of $\approx 4$: the fitted coefficient of $\sqrt{d}$ is only $0.28$, well below the $\tau^* \approx 1.6$ predicted for random patterns. This reduction arises because $\beta^*$ is computed at stored patterns $\mathbf{m}_k$, where the self-similarity $\mathbf{m}_k^\top\mathbf{m}_k = 1$ is an outlier relative to the $\mathcal{O}(1/\sqrt{d})$ cross-similarities, so the softmax concentrates on the self-pattern at lower $\beta$ than a random query would require; the structured correlation spectrum of real protein families further reduces the coefficient (Appendix~\ref{si:beta-star-theory}). From a practical standpoint, Eq.~\eqref{eq:beta-prediction} enables fully automatic operation: given a seed alignment, $d$ is determined by PCA at 95\% variance, and $\beta^*$ follows without a temperature sweep.

\subsection{Regularity and Convergence}\label{sec:convergence}
The modern Hopfield energy has Lipschitz-continuous gradient with constant $L = 1 + \beta\sigma_{\max}^2/2$, where $\sigma_{\max} = \|\mathbf{X}\|_{\mathrm{op}}$, and satisfies the dissipativity condition $\langle \nabla E(\boldsymbol{\xi}), \boldsymbol{\xi} \rangle \geq \|\boldsymbol{\xi}\|_2^2/2 - M^2/2$ with $M = \max_k \|\mathbf{m}_k\|_2$~\cite{alswaidanVarnerStochasticAttention2026}. When $\beta\sigma_{\max}^2 < 2$, the energy is strictly convex and ULA iterates converge to $p_\beta$ in Wasserstein-2 distance at a geometric rate~\cite{dalalyanTheoreticalGuaranteesApproximate2017,durmusNonasymptoticAnalysisUnadjusted2017}.

In the protein setting, operating temperatures of interest satisfy $\beta\sigma_{\max}^2 \gg 2$, placing the sampler in a non-convex regime where the energy landscape has multiple metastable basins concentrated near the stored patterns. Following standard practice for multimodal Langevin sampling~\cite{bovier2004metastability}, we adopt a multi-chain protocol: $N_c$ independent chains are initialized near distinct stored patterns, exploiting fast intra-basin mixing while achieving inter-basin coverage through diverse initialization (empirically, random initialization on $\mathbb{S}^{d-1}$ produces identical output after burn-in; Appendix~\ref{si:random-init}). The Metropolis-adjusted variant (MALA) provides an acceptance-rate diagnostic: acceptance rates near $1.0$ confirm that ULA discretization bias is negligible for the chosen step size $\alpha$.


We applied stochastic attention to protein sequence generation across eight Pfam families and conducted a systematic scaling study to characterize performance as a function of family size. All experiments used the unadjusted Langevin algorithm (ULA) with step size $\alpha=0.01$. For each family and condition, we ran 30 independent chains of $T=5{,}000$ iterations, initialized near randomly chosen stored patterns with Gaussian perturbation ($\sigma_{\mathrm{init}}=0.01$). After discarding $2{,}000$ burn-in iterations and thinning every 100 steps, we retained 5 evenly spaced samples per chain for a total of $S=150$ generated sequences per condition. The burn-in budget provides a $3.2$--$6.6\times$ margin over the empirical convergence point, and the thinning interval matches the integrated autocorrelation time ($\tau_{\mathrm{int}} \approx 76$--$122$ iterations), yielding approximately independent samples (Appendix~\ref{si:sampling-diagnostics}). Metropolis-adjusted Langevin algorithm (MALA) chains were run in parallel with identical parameters to provide acceptance-rate diagnostics; acceptance rates of $99.6$--$99.8\%$ confirmed that ULA discretization bias is negligible at this step size. Throughout, we used PCA dimensionality reduction retaining 95\% of the variance in the one-hot encoded alignment, followed by unit-norm projection to form the memory matrix $\mathbf{X}$. A sensitivity analysis confirmed that generation quality is robust across the 75\%--99\% range, with a practical floor near 70\% retained variance (Appendix~\ref{si:pca-sensitivity}). The complete pipeline, from seed alignment to decoded amino acid sequences, is summarized in Algorithm~\ref{alg:sa-pipeline} (SI Appendix).

The inverse temperature was set automatically for each family via the entropy inflection criterion described above. We identified $\beta^*$ from the attention entropy curve and operated in two regimes, both above $\beta^*$. Below $\beta^*$ the energy landscape is essentially flat and Langevin dynamics degenerates into a random walk on the sphere, producing sequences with no family-specific structure. Above $\beta^*$, metastable basins form around stored patterns and the sampler begins to produce protein-like sequences. We therefore defined a \emph{generation} regime at $\beta_{\mathrm{gen}} = \max(\lceil 2\beta^* \rceil, 5)$, just above the transition, where the basins are shallow enough that the sampler can hop between them, producing novel recombinations of family features. At much higher inverse temperature, we defined a \emph{retrieval} regime at $\beta_{\mathrm{ret}} \approx 20\beta^*$, where the basins are deep and each chain remains trapped near its initialization pattern, producing near-copies of stored sequences. Three simple baselines were evaluated under identical conditions: bootstrap replay (uniform resampling of stored patterns), Gaussian perturbation (stored pattern plus isotropic noise matched to the SA noise scale), and random convex combination (Dirichlet-weighted average of all stored patterns).

We evaluated generated sequences using five metrics. \emph{Novelty} measured departure from stored patterns as $1 - \max_k \cos(\hat{\boldsymbol{\xi}}, \mathbf{m}_k)$ in PCA space. \emph{Diversity} quantified the spread of the generated ensemble as the mean pairwise cosine distance among all $\binom{S}{2}$ pairs of generated samples; a value near $1.0$ indicates that generated sequences are uniformly distributed rather than collapsed to a single mode. \emph{Sequence identity} was the maximum fraction of matching residues between a generated sequence and any stored sequence, computed after argmax decoding from PCA space back to amino acid sequences. \emph{Kullback--Leibler (KL) divergence} of amino acid composition quantified how well generated sequences preserved the family's residue frequency distribution. \emph{Valid residue fraction} confirmed that all decoded positions were standard amino acids.

\paragraph{Protein families and scaling study.}
We selected eight families spanning a range of sizes, sequence lengths, and conservation levels (Table~\ref{tab:families}): RRM (PF00076, $K{=}68$, $L{=}71$), SH3 (PF00018, $K{=}55$, $L{=}48$), WW (PF00397, $K{=}420$, $L{=}31$), Kunitz (PF00014, $K{=}99$, $L{=}53$), zf-C2H2 (PF00096, $K{=}151$, $L{=}23$), PDZ (PF00595, $K{=}44$, $L{=}83$), Pkinase (PF00069, $K{=}37$, $L{=}262$), and Defensin\_beta (PF00711, $K{=}45$, $L{=}36$). Seed alignments were downloaded from InterPro~\cite{mistryPfamProteinFamilies2021} and cleaned by removing columns with $>50\%$ gaps and sequences with $>30\%$ gaps. The resulting PCA dimensions ranged from $d{=}34$ (Pkinase) to $d{=}186$ (WW). To characterize how generation quality degrades with decreasing family size, we also conducted a scaling study in which we fixed the WW domain (the largest family, $K{=}420$) and subsampled to $K \in \{20, 50, 100, 200, 400\}$. At each value of $K$ we drew five independent random subsets, rebuilt the memory matrix from scratch (re-computing PCA and $\beta^*$), ran stochastic attention and all baselines, and recorded metrics; this design isolates the effect of family size from family identity, and the five repeats provide error estimates.

\paragraph{Learned baselines.}
To contextualize SA against methods that require pretraining on large external corpora, we compared four established generative approaches for protein sequences. (i)~\emph{Profile HMM emit}: we built a profile HMM from each family's seed alignment using HMMER3~\cite{hmmer3} (\texttt{hmmbuild --amino}) and emitted $S{=}150$ sequences with \texttt{hmmemit -N 150 --seed 42}. Emitted sequences were filtered to standard amino acids and truncated or padded to the alignment length~$L$. (ii)~\emph{EvoDiff}: we used the MSA-conditioned order-agnostic discrete diffusion model (MSA\_OA\_DM\_RANDSUB)~\cite{alamdariProteinGeneration2023}, which was pretrained on UniRef50 MSAs. Generation used the default denoising schedule with MaxHamming subsampling of the conditioning MSA (depth~$\leq 64$). (iii)~\emph{MSA Transformer}: we used the 100M-parameter MSA Transformer~\cite{raoMSATransformer2021} (esm\_msa1b\_t12\_100M\_UR50S) for generation via iterative Gibbs-like masked language model sampling: a fully masked query row was appended to the family MSA (depth~$\leq 128$), and 50 rounds of masking 15\% of positions and resampling from the softmax were performed. (iv)~\emph{Potts model}: we fit a Potts model by pseudo-likelihood maximization (plmDCA)~\cite{ekeberg2013improved} and generated sequences by Gibbs sampling, implemented from scratch using only NumPy and SciPy with no external DCA packages. For each baseline, the same 150-sequence budget and the same evaluation metrics (KL divergence, novelty, sequence identity) were applied. Because EvoDiff and the MSA Transformer emit variable-length, unaligned sequences, each baseline sequence was conformed to the alignment length $L$ before metric computation by truncating longer sequences to $L$ and padding shorter ones with alanine; this length normalization applies only to the variable-length baselines, not to SA, whose decoded sequences are already length $L$. All statistical comparisons used Wilcoxon rank-sum tests with Benjamini-Hochberg false discovery rate correction at $q{=}0.05$, with Cohen's $d$ reported as the effect size. EvoDiff was run on CPU; for the Pkinase family ($L{=}262$), a single sequence required ${\sim}7$ minutes, yielding ${\sim}14$ hours for 150 sequences, though GPU inference would reduce this (compared to seconds for SA).

\begin{figure}[p]
\centering
\includegraphics[width=\textwidth]{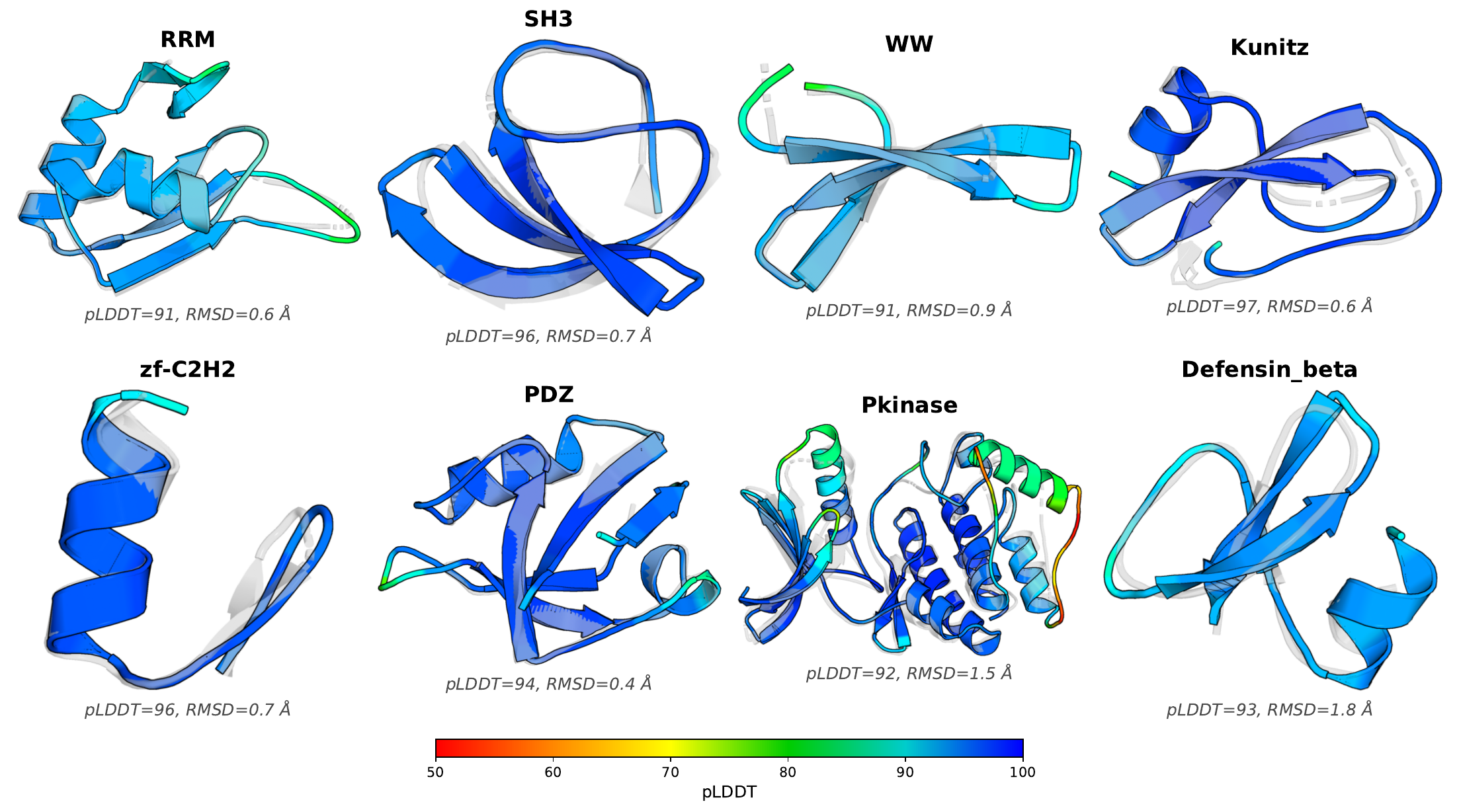}
\caption{AlphaFold2 predicted structures for representative SA-generated sequences (colored by per-residue pLDDT confidence) superimposed on the experimentally determined reference structure for each family (gray, semi-transparent). For visualization, the SA-generated sequence with the highest mean pLDDT in each family was selected and aligned to the reference using PyMOL; aggregate structural statistics for the generated ensembles are reported in Fig.~\ref{fig:structure-validation} and SI Appendix, Figs.~\ref{fig:structure-esmfold-af2}--\ref{fig:esmfold-af2-concordance}. The close predicted structural agreement (sub-angstrom root-mean-square deviation (RMSD)) and uniformly high pLDDT scores are consistent with SA-generated sequences encoding three-dimensional folds that recapitulate the canonical architecture of their target family, despite $40$--$65\%$ sequence novelty.}
\label{fig:structure-gallery}
\end{figure}

\paragraph{Structure validation.}
To assess whether generated sequences adopt structures consistent with their target family, we predicted the three-dimensional structure of 50 sequences per method per family using the ESMFold API~\cite{linEvolutionaryScaleModeling2023}. For each predicted structure, we extracted the mean predicted local distance difference test (pLDDT) score from the B-factor column as a measure of folding confidence (pLDDT~$> 70$ indicates a confident prediction). We then compared each predicted structure to a representative experimentally determined structure for the family using TM-align~\cite{zhangTMalign2005} (TM-score~$> 0.5$ indicates the same fold). Reference structures were obtained from the Research Collaboratory for Structural Bioinformatics Protein Data Bank (RCSB PDB): 1FXL (RRM), 1SHG (SH3), 1PIN (WW), 1BPI (Kunitz), 1ZAA (zf-C2H2), 1BE9 (PDZ), 1ATP (Pkinase), and 1E4S (Defensin\_beta). As a positive control, we also predicted structures for 50 stored (natural) sequences from each family, providing an upper bound on expected pLDDT and TM-score. As an independent cross-validation, we repeated the structure prediction for all eight families using AlphaFold2~\cite{jumperHighlyAccurateProtein2021} via ColabFold~\cite{mirdita2022colabfold}, predicting 50 SA-generated and 50 stored sequences per family with the \texttt{alphafold2\_ptm} model (single model, 3 recycles, no relaxation). For each family, we selected the SA-generated structure with the highest mean pLDDT and superimposed it on the experimental reference using PyMOL to produce the structural gallery in Fig.~\ref{fig:structure-gallery}.

\paragraph{Antimicrobial peptide biophysical analysis.}
The beta defensin family (PF00711) provided an opportunity to assess whether generated sequences preserved biophysical properties associated with antimicrobial function. For each generated and stored sequence, we computed the net charge at pH~7 (summing $+1$ for Arg and Lys, $-1$ for Asp and Glu), the hydrophobic residue fraction (Ala, Val, Ile, Leu, Met, Phe, Trp, Pro), the mean Kyte--Doolittle hydrophobicity, and the cysteine count. We defined an antimicrobial plausibility filter requiring net charge $\geq +2$, hydrophobic ratio in $[0.3, 0.7]$, and $\geq 4$ cysteines (reflecting the three disulfide bonds characteristic of the beta-defensin fold), and computed the pass rate for each method.

\paragraph{Critical temperature prediction.}
We pooled $n{=}33$ data points (the 8 Pfam families and 25 WW scaling replicates, 5 family sizes $\times$ 5 repeats) and fit ordinary least-squares regression models predicting $\beta^*$ from MSA-derived features. We considered four candidate features: $\sqrt{d}$ (PCA dimensionality at 95\% retained variance), mean per-column Shannon entropy $\bar{H}_{\mathrm{col}}$, $\log K_{\mathrm{eff}}$ (effective number of sequences at 80\% identity), and spectral concentration $\lambda_1 / \mathrm{tr}(\mathbf{C})$ (leading eigenvalue of the sequence covariance, normalized by trace). We evaluated nested models by adding features in order of marginal $R^2$ improvement. Uncertainty on regression coefficients was quantified via nonparametric bootstrap: we resampled the 33 observations with replacement $10{,}000$ times, refitting the model on each resample, and reported the standard deviation of the bootstrap distribution as the coefficient SE. Model comparison used root-mean-square error (RMSE) and $R^2$, with the naive random-pattern prediction $\beta^* = \sqrt{d}$ as a reference baseline.

\section*{Acknowledgments}
The idea for SA arose while developing a lecture on energy-based methods for eCornell, an opportunity without which this work would not have begun. I am deeply grateful to eCornell for making it possible. We thank the Pfam/InterPro consortium for maintaining publicly available seed alignments. Structure predictions were performed using ESMFold and AlphaFold2 (via ColabFold). Computations were performed on resources at Cornell University. This research received no external funding.

\paragraph{Author contributions.} J.D.V.\ designed research, performed research, analyzed data, and wrote the paper.

\paragraph{Competing interests.} The author declares no competing interest.

\paragraph{Data availability.} All code, data, seed alignments, generated sequences, baseline outputs, structure metrics and predictions, deep mutational scanning mappings, and analysis scripts needed to regenerate every table and figure are openly available. The exact version used for this study is archived at Zenodo (DOI: \href{https://doi.org/10.5281/zenodo.20835924}{10.5281/zenodo.20835924}), with ongoing development at \url{https://github.com/varnerlab/SA-Protein-Modeling-Study}. Pfam seed alignments were obtained from InterPro~\cite{mistryPfamProteinFamilies2021}. No new experimental data were generated for this study; all validation used publicly available computational tools (ESMFold, AlphaFold2 via ColabFold, ESM2-650M, the MSA Transformer, EvoDiff, HMMER, and TM-align) and published deep mutational scanning datasets.

\bibliographystyle{unsrtnat}
\bibliography{References_v1}

\clearpage

\clearpage  

\begin{table}[p]
\centering
\caption{Properties of the eight Pfam families used in this study. $K$: number of seed sequences after cleaning. $L$: alignment length (columns retained). $d$: PCA dimension (95\% variance). $\bar{H}_{\mathrm{col}}$: mean per-column Shannon entropy (nats). $K_{\mathrm{eff}}$: effective number of sequences at 80\% identity threshold. $\beta^*$: empirical entropy inflection point. $\beta^*/\!\sqrt{d}$: ratio to the random-pattern prediction.}
\label{tab:families}
\small
\begin{tabular}{@{}lrrrrrrrr@{}}
\toprule
Family & Pfam ID & $K$ & $L$ & $d$ & $\bar{H}_{\mathrm{col}}$ & $K_{\mathrm{eff}}$ & $\beta^*$ & $\beta^*/\!\sqrt{d}$ \\
\midrule
RRM     & PF00076 &  68 &  71 &  59 & 1.92 &  68 & 3.85 & 0.50 \\
SH3     & PF00018 &  55 &  48 &  46 & 1.68 &  55 & 3.85 & 0.57 \\
WW      & PF00397 & 420 &  31 & 186 & 1.74 & 419 & 5.45 & 0.40 \\
Kunitz  & PF00014 &  99 &  53 &  80 & 1.61 &  99 & 3.85 & 0.43 \\
zf-C2H2 & PF00096 & 151 &  23 & 106 & 1.91 & 151 & 4.58 & 0.44 \\
PDZ     & PF00595 &  44 &  83 &  37 & 1.88 &  43 & 3.23 & 0.53 \\
Pkinase & PF00069 &  37 & 262 &  34 & 1.72 &  37 & 3.23 & 0.55 \\
Defensin\_beta & PF00711 &  45 &  36 &  37 & 1.65 &  45 & 3.23 & 0.53 \\
\bottomrule
\end{tabular}
\end{table}

\clearpage
\setcounter{figure}{0}
\setcounter{table}{0}
\setcounter{equation}{0}
\renewcommand{\thefigure}{S\arabic{figure}}
\renewcommand{\thetable}{S\arabic{table}}
\renewcommand{\theequation}{S\arabic{equation}}
\setcounter{section}{0}
\renewcommand{\thesection}{S\arabic{section}}

\section*{SI Appendix}

\section{Derivation of the Stochastic Attention Model}\label{si:derivation}
Sampling from a broad class of generative models can be reduced to a single task: drawing a state vector $\boldsymbol{\xi}\in\mathbb{R}^d$ from a target density
$p$ written in Gibbs (Boltzmann) form as:
\[
  p(\boldsymbol{\xi}) \;\propto\; \exp\!\bigl(-U(\boldsymbol{\xi})\bigr),
\]
where the potential $U(\boldsymbol{\xi})\in\mathbb{R}$ is the negative log-density of configuration $\boldsymbol{\xi}$, so that low-potential states carry high probability.
The unadjusted Langevin algorithm (ULA)~\cite{robertsTweedieExponentialConvergence1996} samples from this density by descending the
potential gradient while adding Gaussian noise, with step size $\tilde\alpha>0$:
\[
  \boldsymbol{\xi}_{t+1} \;=\;
  \boldsymbol{\xi}_t
  \;\underbrace{-\,\tilde\alpha\,\nabla U(\boldsymbol{\xi}_t)}_{\textstyle\text{score step}}
  \;+\; \sqrt{2\tilde\alpha}\;\boldsymbol{\epsilon}_t,
  \qquad \boldsymbol{\epsilon}_t\sim\mathcal{N}(\mathbf{0},\mathbf{I}_d).
\]

The ULA requires the score function $\nabla\log p = -\nabla U$. In
diffusion and score-based models this score is not available in closed form, so a neural
network must learn it from a large set of samples. We represent the protein family as $K$ stored patterns
$\mathbf{m}_k\in\mathbb{R}^d$, the columns of a unit-normalized memory matrix
$\mathbf{X} = [\mathbf{m}_1,\dots,\mathbf{m}_K]\in\mathbb{R}^{d\times K}$.
The modern Hopfield log-sum-exp (LSE) energy~\cite{ramsauerHopfieldNetworksAll2021} at inverse temperature $\beta>0$ defines the
landscape:
\[
  E(\boldsymbol{\xi})
  \;=\;
  \tfrac{1}{2}\,\|\boldsymbol{\xi}\|_2^2
  \;-\;
  \tfrac{1}{\beta}\,\log\!\left(\sum_{k=1}^{K}\exp\!\bigl(\beta\,\mathbf{m}_k^{\top}\boldsymbol{\xi}\bigr)\right),
\]
whose Boltzmann distribution $p_\beta(\boldsymbol{\xi})\propto\exp(-\beta E(\boldsymbol{\xi}))$ is the target density. For this target the score is available in closed form: the
gradient of $E$ is precisely the attention residual, the current state minus the attention output, so:
\[
  \underbrace{\nabla\log p_\beta(\boldsymbol{\xi})}_{\text{score}}
  \;=\;
  -\beta\,\nabla E(\boldsymbol{\xi})
  \;=\;
  -\beta\Bigl\{\boldsymbol{\xi} \;-\;
  \underbrace{\mathbf{X}\,\operatorname{softmax}\!\bigl(\beta\,\mathbf{X}^{\top}\boldsymbol{\xi}\bigr)}_{\textstyle\mathbf{T}(\boldsymbol{\xi})}\Bigr\}.
\]
The softmax converts the pattern similarities $\beta\,\mathbf{X}^{\top}\boldsymbol{\xi}$ into
nonnegative weights that sum to one, so $\mathbf{T}(\boldsymbol{\xi})$ is a weighted average of
the stored patterns and a single attention evaluation yields the exact score, with no network to train. With this closed-form score, the ULA update uses
potential $U=\beta E$ and step size $\tilde\alpha=\alpha/\beta$, where the mixing coefficient $\alpha\in(0,1)$ controls
the step length; the resulting score step
$-\tilde\alpha\nabla U = \alpha\bigl(\mathbf{T}(\boldsymbol{\xi})-\boldsymbol{\xi}\bigr)$
gives the update:
\[
  \boxed{%
  \boldsymbol{\xi}_{t+1}
  \;=\;
  \underbrace{(1-\alpha)\,\boldsymbol{\xi}_t}_{\text{contraction}}
  \;+\;
  \underbrace{\alpha\,\mathbf{X}\,\operatorname{softmax}\!\bigl(\beta\,\mathbf{X}^{\top}\boldsymbol{\xi}_t\bigr)}_{\text{attention pull to memories}}
  \;+\;
  \underbrace{\sqrt{\tfrac{2\alpha}{\beta}}\;\boldsymbol{\epsilon}_t}_{\text{Gaussian noise}}
  }
\]

\section{Amino Acid Composition Profiles}\label{si:aa-freq}

\begin{figure}[ht]
\centering
\includegraphics[width=0.9\textwidth]{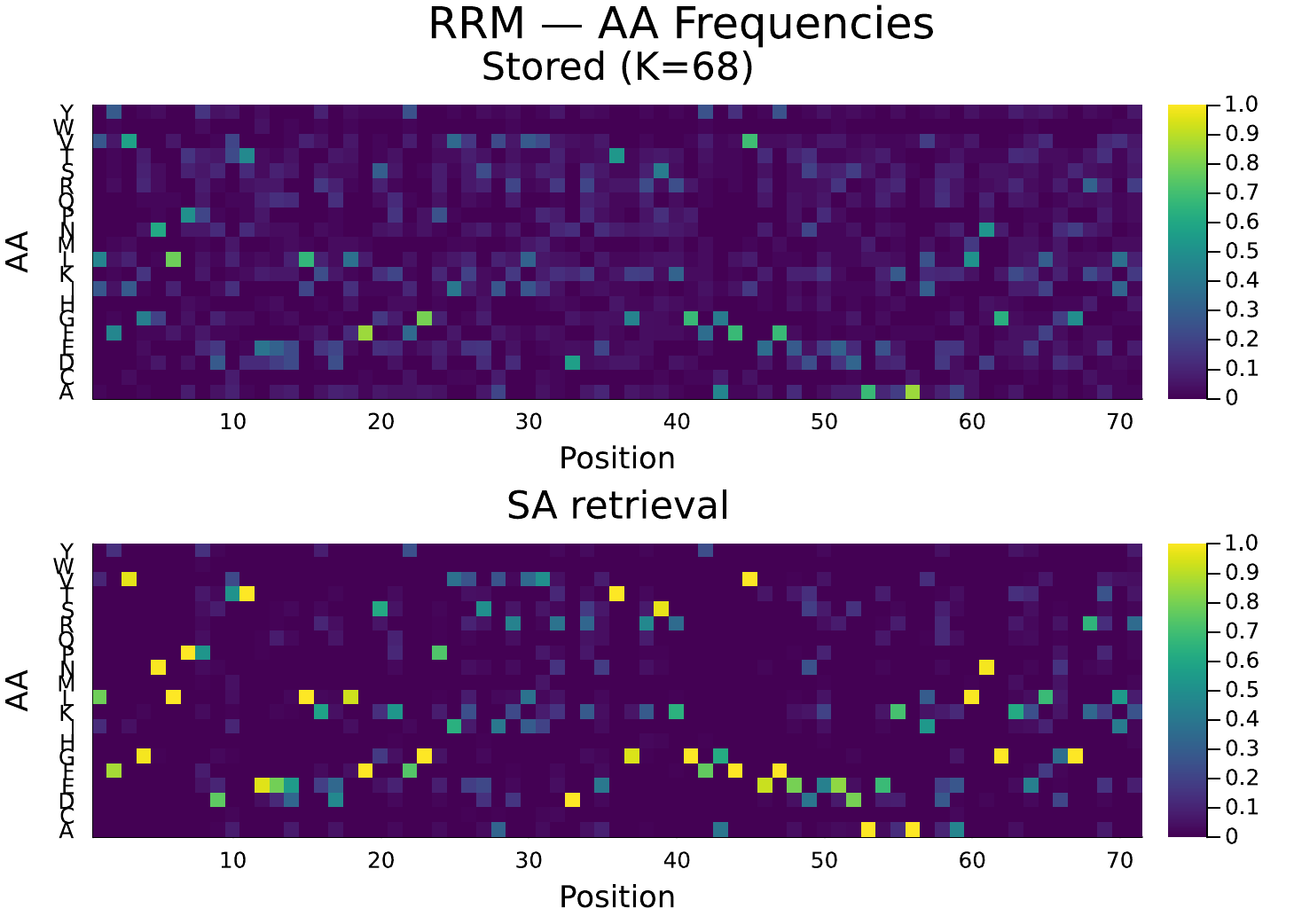}
\caption{Per-position amino acid frequency profiles for the RRM family (PF00076). \textbf{Top:} stored seed sequences ($K{=}68$). \textbf{Bottom:} SA-generated sequences ($S{=}150$, retrieval regime). The generated ensemble reproduced the position-specific conservation pattern of the family, including strongly conserved positions and variable regions.}
\label{fig:aa-freq}
\end{figure}

We compared the per-position amino acid frequency profiles of the stored seed sequences and stochastic attention (SA)-generated sequences for the RNA recognition motif (RRM) family (PF00076, Fig.~\ref{fig:aa-freq}). The generated ensemble closely reproduced the conservation pattern of the family: strongly conserved positions (e.g., the aromatic residues characteristic of the ribonucleoprotein motifs RNP1 and RNP2) retained high frequency in the generated output, while variable positions displayed a similar breadth of amino acid usage. This agreement is consistent with the low Kullback--Leibler (KL) divergences reported in Table~\ref{tab:results} and indicates that the Boltzmann sampling procedure captures not only global amino acid composition but also the position-specific constraints encoded in the seed alignment.

\section{Generation Metrics}\label{si:generation-metrics}

Table~\ref{tab:results} reports the full generation metrics for all eight Pfam families across five sampling conditions: SA generation ($\beta \approx 2\beta^*$), SA retrieval ($\beta \approx 20\beta^*$), bootstrap resampling, Gaussian perturbation, and convex combination. For each condition, we report the KL divergence of amino acid composition, principal component analysis (PCA)-space cosine novelty, and nearest sequence identity to a stored pattern. Standard errors (SE) are computed over 30 independent Langevin chains of 5 samples each; KL standard errors are obtained via 1{,}000-fold bootstrap resampling. Bold entries indicate the best-performing method for each family and metric. The generation regime provides the best balance of compositional fidelity, novelty, and diversity across all eight families, while the convex combination baseline increases KL divergence in most families due to interpolation artifacts in PCA space.

\begin{table}[ht]
\centering
\caption{Generation metrics across eight Pfam families (mean $\pm$ SE). The label SA gen denotes the generation regime ($\beta \approx 2\beta^*$), and SA ret denotes the retrieval regime ($\beta \approx 20\beta^*$). Baseline abbreviations are BS, bootstrap; GP, Gaussian perturbation; and CC, convex combination. The KL column reports amino acid composition KL divergence ($\downarrow$ lower is better; SE via 1{,}000-fold bootstrap). Novelty is $1 - \max_k\cos(\hat{\boldsymbol{\xi}}, \mathbf{m}_k)$ in PCA space ($\uparrow$ higher is better). The SeqID column reports nearest sequence identity to a stored pattern after decoding ($\downarrow$ lower is better). Bold indicates the best method per family and metric. The KL SE here is the $1{,}000$-fold bootstrap SE; because KL divergence is non-negative and near zero for several families, this bootstrap distribution is right-skewed, so the symmetric SE can exceed the point estimate (e.g., zf-C2H2, where a percentile interval would be more appropriate). The per-family comparison tables (Tables~\ref{tab:baseline-metrics} and~\ref{tab:potts}) instead report the SE across $30$ sub-blocks of five sequences; the SA-generation KL point estimates are identical across tables, and only the SE estimator differs.}
\label{tab:results}
\small
\begin{tabular}{@{}llccc@{}}
\toprule
Family & Method & KL$_{\mathrm{AA}}$ $(\downarrow)$ & Novelty $(\uparrow)$ & SeqID $(\downarrow)$ \\
\midrule
\multirow{5}{*}{RRM}
 & SA gen  & $\mathbf{0.060 \pm 0.005}$ & $\mathbf{0.513 \pm 0.011}$ & $\mathbf{0.538 \pm 0.006}$ \\
 & SA ret  & $0.107 \pm 0.010$ & $0.334 \pm 0.020$ & $0.616 \pm 0.012$ \\
 & BS      & $0.133 \pm 0.006$ & $0.000 \pm 0.000$ & $0.645 \pm 0.004$ \\
 & GP      & $0.132 \pm 0.005$ & $0.008 \pm 0.000$ & $0.633 \pm 0.006$ \\
 & CC      & $2.091 \pm 0.000$ & $0.510 \pm 0.009$ & $0.562 \pm 0.000$ \\
\midrule
\multirow{5}{*}{SH3}
 & SA gen  & $0.036 \pm 0.004$ & $\mathbf{0.471 \pm 0.012}$ & $\mathbf{0.593 \pm 0.006}$ \\
 & SA ret  & $0.035 \pm 0.004$ & $0.213 \pm 0.012$ & $0.722 \pm 0.011$ \\
 & BS      & $\mathbf{0.030 \pm 0.003}$ & $0.000 \pm 0.000$ & $0.772 \pm 0.006$ \\
 & GP      & $\mathbf{0.030 \pm 0.003}$ & $0.006 \pm 0.000$ & $0.752 \pm 0.006$ \\
 & CC      & $0.798 \pm 0.012$ & $0.474 \pm 0.008$ & $0.601 \pm 0.002$ \\
\midrule
\multirow{5}{*}{WW}
 & SA gen  & $\mathbf{0.008 \pm 0.002}$ & $\mathbf{0.653 \pm 0.006}$ & $\mathbf{0.562 \pm 0.005}$ \\
 & SA ret  & $0.046 \pm 0.013$ & $0.354 \pm 0.016$ & $0.763 \pm 0.011$ \\
 & BS      & $0.024 \pm 0.003$ & $0.000 \pm 0.000$ & $0.823 \pm 0.005$ \\
 & GP      & $0.029 \pm 0.004$ & $0.017 \pm 0.000$ & $0.809 \pm 0.006$ \\
 & CC      & $1.375 \pm 0.092$ & $0.637 \pm 0.004$ & $0.766 \pm 0.001$ \\
\midrule
\multirow{5}{*}{Kunitz}
 & SA gen  & $\mathbf{0.013 \pm 0.002}$ & $\mathbf{0.557 \pm 0.009}$ & $\mathbf{0.607 \pm 0.004}$ \\
 & SA ret  & $0.037 \pm 0.003$ & $0.257 \pm 0.016$ & $0.760 \pm 0.012$ \\
 & BS      & $0.023 \pm 0.002$ & $0.000 \pm 0.000$ & $0.834 \pm 0.004$ \\
 & GP      & $0.024 \pm 0.002$ & $0.010 \pm 0.000$ & $0.844 \pm 0.004$ \\
 & CC      & $0.721 \pm 0.013$ & $0.546 \pm 0.007$ & $0.590 \pm 0.001$ \\
\midrule
\multirow{5}{*}{zf-C2H2}
 & SA gen  & $0.013 \pm 0.028$ & $\mathbf{0.596 \pm 0.010}$ & $\mathbf{0.557 \pm 0.007}$ \\
 & SA ret  & $0.021 \pm 0.010$ & $0.167 \pm 0.014$ & $0.848 \pm 0.013$ \\
 & BS      & $\mathbf{0.007 \pm 0.002}$ & $0.000 \pm 0.000$ & $0.938 \pm 0.004$ \\
 & GP      & $\mathbf{0.007 \pm 0.002}$ & $0.011 \pm 0.000$ & $0.940 \pm 0.004$ \\
 & CC      & $2.346 \pm 0.060$ & $0.580 \pm 0.006$ & $0.600 \pm 0.001$ \\
\midrule
\multirow{5}{*}{PDZ}
 & SA gen  & $\mathbf{0.038 \pm 0.010}$ & $\mathbf{0.418 \pm 0.009}$ & $\mathbf{0.543 \pm 0.005}$ \\
 & SA ret  & $0.059 \pm 0.016$ & $0.236 \pm 0.017$ & $0.626 \pm 0.014$ \\
 & BS      & $0.041 \pm 0.002$ & $0.000 \pm 0.000$ & $0.642 \pm 0.009$ \\
 & GP      & $0.045 \pm 0.003$ & $0.006 \pm 0.000$ & $0.648 \pm 0.007$ \\
 & CC      & $0.339 \pm 0.001$ & $0.449 \pm 0.009$ & $0.549 \pm 0.001$ \\
\midrule
\multirow{5}{*}{Pkinase}
 & SA gen  & $\mathbf{0.035 \pm 0.001}$ & $\mathbf{0.478 \pm 0.011}$ & $0.515 \pm 0.004$ \\
 & SA ret  & $0.050 \pm 0.001$ & $0.316 \pm 0.013$ & $0.531 \pm 0.006$ \\
 & BS      & $0.054 \pm 0.001$ & $0.000 \pm 0.000$ & $0.522 \pm 0.003$ \\
 & GP      & $0.051 \pm 0.001$ & $0.005 \pm 0.000$ & $0.533 \pm 0.003$ \\
 & CC      & $0.062 \pm 0.001$ & $0.446 \pm 0.009$ & $\mathbf{0.470 \pm 0.000}$ \\
\midrule
\multirow{5}{*}{Defensin\_beta}
 & SA gen  & $\mathbf{0.022 \pm 0.004}$ & $\mathbf{0.402 \pm 0.013}$ & $\mathbf{0.655 \pm 0.007}$ \\
 & SA ret  & $0.025 \pm 0.008$ & $0.166 \pm 0.015$ & $0.790 \pm 0.015$ \\
 & BS      & $0.020 \pm 0.003$ & $0.000 \pm 0.000$ & $0.836 \pm 0.006$ \\
 & GP      & $0.023 \pm 0.003$ & $0.006 \pm 0.000$ & $0.835 \pm 0.006$ \\
 & CC      & $0.708 \pm 0.001$ & $0.442 \pm 0.009$ & $0.716 \pm 0.001$ \\
\bottomrule
\end{tabular}
\end{table}

Table~\ref{tab:baseline-metrics} reports the per-family generation metrics for the three learned baselines, profile HMM emit, EvoDiff, and the MSA Transformer, that stochastic attention is compared against in the main-text head-to-head analysis. For each family and method, the amino acid composition KL divergence, the PCA-space cosine novelty, and the nearest sequence identity to a stored pattern were computed under the same 150-sequence budget and evaluation pipeline applied to SA generation, and are reported as the mean and standard error over 30 sub-blocks of the generated set, following the convention of Table~\ref{tab:potts}. The learned baselines attained low KL divergence in most families, but their nearest sequence identities fell below the empirical within-family nearest-neighbor identity envelope (Table~\ref{tab:nn-band}) in several families. For example, profile HMM emit reached $11\%$ and the MSA Transformer $9\%$ in Pkinase, below the minimum nearest-neighbor identity of any natural sequence in that family ($28\%$), indicating compositional drift away from the stored patterns rather than family-consistent generation. EvoDiff showed elevated composition divergence on the longest family (Pkinase, KL${=}0.173$, more than an order of magnitude above its values in the other seven families). This largely reflects EvoDiff generating sequences short of the alignment length (${\approx}199$ residues vs.\ $L{=}262$); the length-normalization step pads these with alanine, inflating the apparent alanine frequency, rather than reflecting an intrinsic compositional failure of the model. Together with the SA generation metrics in Table~\ref{tab:results} and the Potts comparison in Table~\ref{tab:potts}, these values constitute the complete per-family record underlying the baseline comparison.

\begin{table}[ht]
\centering
\caption{Generation metrics for the three learned baselines (profile HMM emit, EvoDiff, and the MSA Transformer) across all eight Pfam families, computed identically to the SA generation analysis (150 sequences per family). The KL column reports amino acid composition KL divergence ($\downarrow$ lower is better). Novelty: PCA-space cosine novelty ($\uparrow$ higher is better). SeqID: nearest sequence identity to a stored pattern. Values are mean $\pm$ SE over 30 sub-blocks of the generated set. These are the per-family values shown in the main-text baseline comparison.}
\label{tab:baseline-metrics}
\small
\begin{tabular}{@{}llccc@{}}
\toprule
Family & Method & KL$_{\mathrm{AA}}$ $(\downarrow)$ & Novelty $(\uparrow)$ & SeqID \\
\midrule
\multirow{3}{*}{RRM}
 & HMM emit        & $0.009 \pm 0.005$ & $0.647 \pm 0.005$ & $0.209 \pm 0.005$ \\
 & EvoDiff         & $0.005 \pm 0.006$ & $0.593 \pm 0.008$ & $0.386 \pm 0.005$ \\
 & MSA Transformer & $0.022 \pm 0.003$ & $0.572 \pm 0.008$ & $0.266 \pm 0.007$ \\
\midrule
\multirow{3}{*}{SH3}
 & HMM emit        & $0.009 \pm 0.017$ & $0.614 \pm 0.004$ & $0.290 \pm 0.006$ \\
 & EvoDiff         & $0.003 \pm 0.021$ & $0.434 \pm 0.016$ & $0.509 \pm 0.011$ \\
 & MSA Transformer & $0.011 \pm 0.016$ & $0.544 \pm 0.011$ & $0.418 \pm 0.008$ \\
\midrule
\multirow{3}{*}{WW}
 & HMM emit        & $0.007 \pm 0.020$ & $0.706 \pm 0.003$ & $0.389 \pm 0.009$ \\
 & EvoDiff         & $0.009 \pm 0.011$ & $0.642 \pm 0.006$ & $0.509 \pm 0.004$ \\
 & MSA Transformer & $0.043 \pm 0.011$ & $0.670 \pm 0.006$ & $0.417 \pm 0.008$ \\
\midrule
\multirow{3}{*}{Kunitz}
 & HMM emit        & $0.030 \pm 0.011$ & $0.665 \pm 0.004$ & $0.229 \pm 0.006$ \\
 & EvoDiff         & $0.011 \pm 0.012$ & $0.621 \pm 0.008$ & $0.434 \pm 0.007$ \\
 & MSA Transformer & $0.024 \pm 0.018$ & $0.598 \pm 0.009$ & $0.338 \pm 0.013$ \\
\midrule
\multirow{3}{*}{zf-C2H2}
 & HMM emit        & $0.006 \pm 0.038$ & $0.670 \pm 0.004$ & $0.415 \pm 0.008$ \\
 & EvoDiff         & $0.003 \pm 0.053$ & $0.646 \pm 0.007$ & $0.469 \pm 0.005$ \\
 & MSA Transformer & $0.014 \pm 0.044$ & $0.619 \pm 0.007$ & $0.477 \pm 0.007$ \\
\midrule
\multirow{3}{*}{PDZ}
 & HMM emit        & $0.012 \pm 0.005$ & $0.591 \pm 0.004$ & $0.143 \pm 0.004$ \\
 & EvoDiff         & $0.005 \pm 0.007$ & $0.344 \pm 0.020$ & $0.436 \pm 0.016$ \\
 & MSA Transformer & $0.032 \pm 0.004$ & $0.367 \pm 0.016$ & $0.298 \pm 0.015$ \\
\midrule
\multirow{3}{*}{Pkinase}
 & HMM emit        & $0.019 \pm 0.001$ & $0.561 \pm 0.006$ & $0.112 \pm 0.002$ \\
 & EvoDiff         & $0.173 \pm 0.025$ & $0.554 \pm 0.008$ & $0.147 \pm 0.005$ \\
 & MSA Transformer & $0.052 \pm 0.004$ & $0.535 \pm 0.006$ & $0.092 \pm 0.002$ \\
\midrule
\multirow{3}{*}{Defensin\_beta}
 & HMM emit        & $0.031 \pm 0.012$ & $0.574 \pm 0.007$ & $0.362 \pm 0.007$ \\
 & EvoDiff         & $0.009 \pm 0.035$ & $0.389 \pm 0.012$ & $0.570 \pm 0.009$ \\
 & MSA Transformer & $0.038 \pm 0.052$ & $0.316 \pm 0.017$ & $0.590 \pm 0.015$ \\
\bottomrule
\end{tabular}
\end{table}

\section{Scaling Study and Multi-Family Generalization}\label{si:scaling}

\begin{figure}[ht]
\centering
\includegraphics[width=\textwidth]{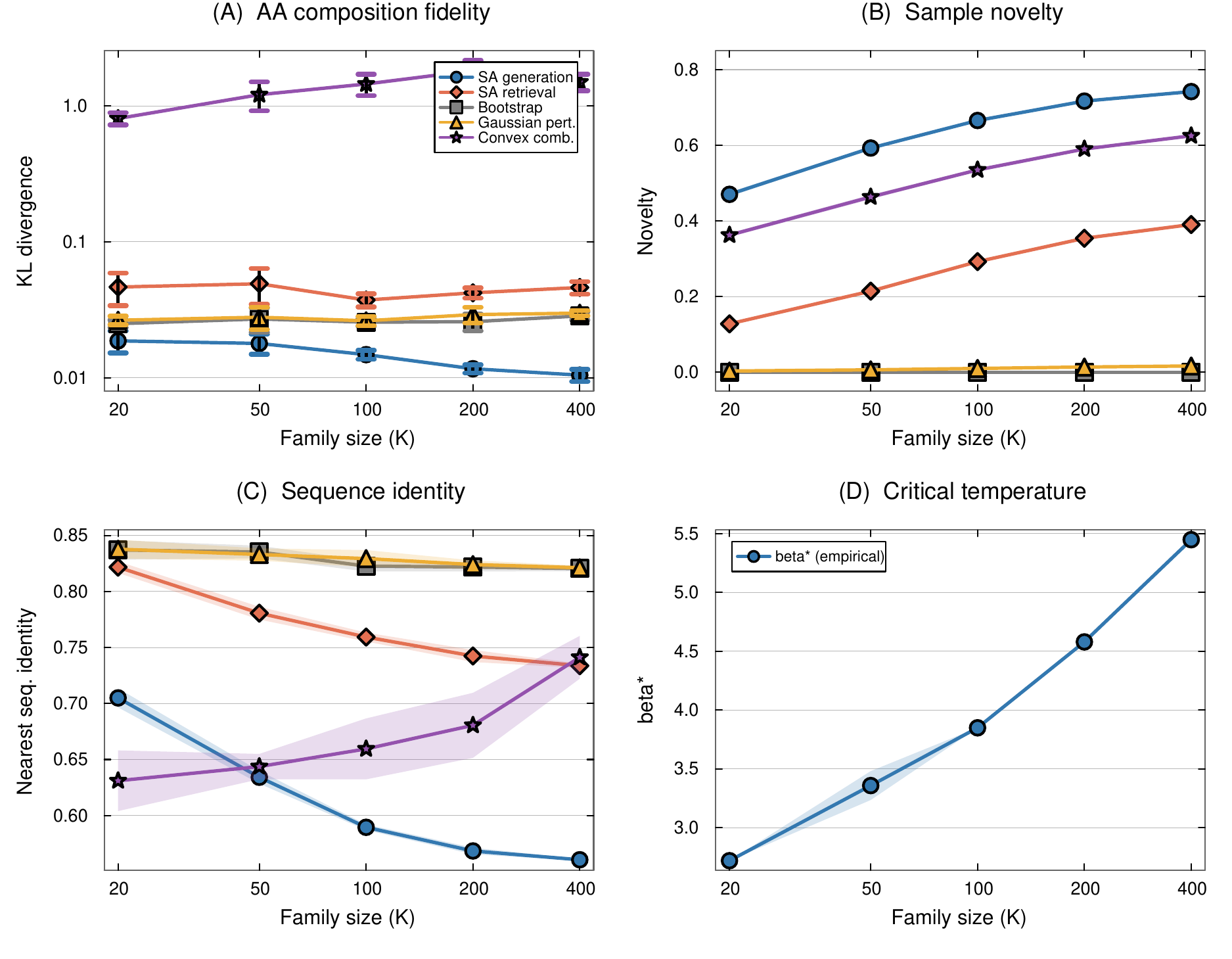}
\caption{Scaling study: WW domain subsampled at $K \in \{20, 50, 100, 200, 400\}$ with five independent replicates. \textbf{(A)}~KL divergence (log--log scale). Stochastic attention generation maintains low KL across the entire range; convex combination increases KL. \textbf{(B)}~Novelty increases with $K$ for SA methods; baselines remain near zero. \textbf{(C)}~Nearest sequence identity decreases with $K$, indicating the sampler explores further from stored patterns as more patterns shape the energy landscape. \textbf{(D)}~The empirical critical temperature $\beta^*$ increases with family size. Shaded regions and error bars show $\pm 1$ SE.}
\label{fig:scaling}
\end{figure}

To characterize how generation quality degrades with decreasing family size, we subsampled the WW domain ($K{=}420$) at $K \in \{20, 50, 100, 200, 400\}$ with five independent replicates per condition (Fig.~\ref{fig:scaling}). In the generation regime, stochastic attention maintained low KL divergence across the entire range ($0.019 \pm 0.008$ at $K{=}20$, improving to $0.010 \pm 0.002$ at $K{=}400$), while the convex combination baseline exceeded $\mathrm{KL}{=}0.8$ at every family size. Novelty increased with $K$ (from $0.47$ to $0.74$), reflecting the richer landscape of a larger memory set, while nearest sequence identity decreased correspondingly (from $0.71$ to $0.56$). The empirical critical temperature $\beta^*$ increased from $2.7$ at $K{=}20$ to $5.5$ at $K{=}400$, tracking the growth in PCA dimensionality. These results establish that SA generation is robust down to alignments as small as 20 sequences, though with reduced novelty reflecting the more constrained energy landscape.

The WW-domain-only scaling study showed robustness down to $K{=}20$, but a single family cannot establish generality. To test whether this robustness holds across families with different sequence lengths, conservation levels, and PCA dimensionalities, we subsampled three additional families to $K{=}20$ with five independent replicates each: SH3 ($K_{\mathrm{full}}{=}55$, $L{=}48$), Kunitz ($K_{\mathrm{full}}{=}99$, $L{=}53$), and zf-C2H2 ($K_{\mathrm{full}}{=}151$, $L{=}23$). All stochastic attention parameters were identical to the main experiment; PCA and $\beta^*$ were recomputed from scratch for each subsample. Table~\ref{tab:multifamily-k20} reports the results alongside the original WW subsamples at $K{=}20$. Stochastic attention maintained low KL divergence ($0.013$--$0.026$), novelty ($0.46$--$0.47$), and moderate sequence identity ($0.67$--$0.73$) across all four families, with narrow standard deviations across the five replicates. The consistency of these metrics across families that differ in length by more than a factor of two (zf-C2H2: $L{=}23$ vs.\ Kunitz: $L{=}53$) indicates that the entropy inflection criterion adapts the operating temperature to each family's geometry, and that the robustness observed in the WW domain scaling study is not family-specific.

\begin{table}[ht]
\centering
\small
\caption{SA generation quality at $K{=}20$ across four Pfam families (five independent subsamples each). Values are mean $\pm$ SD over the five replicates.}
\label{tab:multifamily-k20}
\begin{tabular}{lcccccc}
\toprule
Family & $K_{\mathrm{full}}$ & $L$ & $d_{\mathrm{PCA}}$ & KL & Novelty & Seq.\ ID \\
\midrule
SH3     & 55  & 48 & 17--18 & $0.026 \pm 0.003$ & $0.46 \pm 0.01$ & $0.67 \pm 0.01$ \\
Kunitz  & 99  & 53 & 17--18 & $0.020 \pm 0.004$ & $0.47 \pm 0.01$ & $0.72 \pm 0.01$ \\
zf-C2H2 & 151 & 23 & 18     & $0.013 \pm 0.004$ & $0.47 \pm 0.01$ & $0.73 \pm 0.00$ \\
WW      & 420 & 31 & 17--18 & $0.019 \pm 0.008$ & $0.47 \pm 0.01$ & $0.71 \pm 0.02$ \\
\bottomrule
\end{tabular}
\end{table}

\section{AlphaFold2 Cross-Validation}\label{si:af2}

\begin{figure}[ht]
\centering
\includegraphics[width=\textwidth]{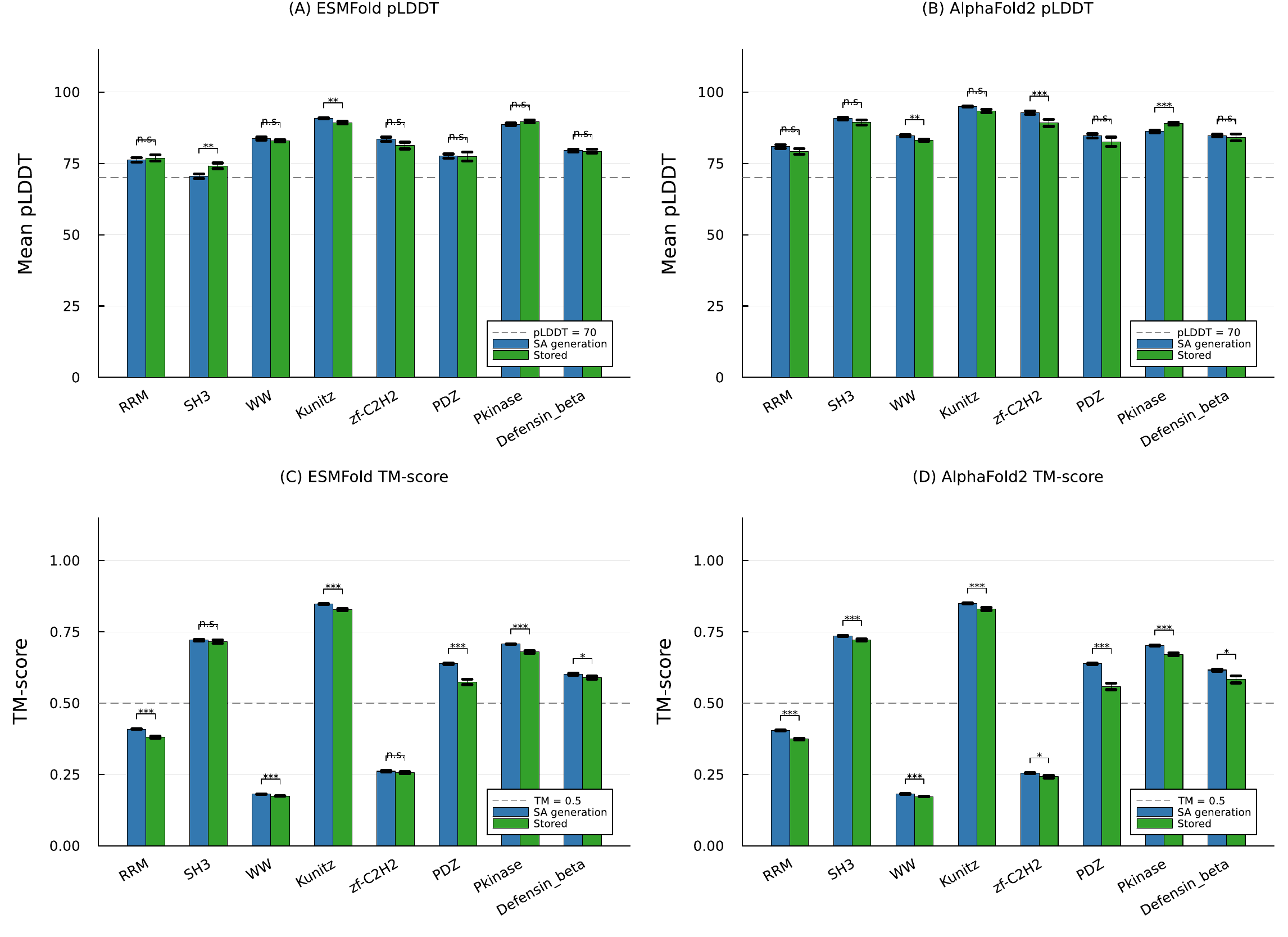}
\caption{Structure validation results from ESMFold \textbf{(A, C)} and AlphaFold2 \textbf{(B, D)} across all eight Pfam families. \textbf{(A, B)}~Mean pLDDT for SA-generated (blue) and stored (green) sequences; dashed line at pLDDT${=}70$. \textbf{(C, D)}~Mean TM-score to the family reference structure; dashed line at TM${=}0.5$. Significance brackets show Wilcoxon rank-sum tests comparing SA generation to stored sequences. The two predictors produce similar qualitative patterns across all families.}
\label{fig:structure-esmfold-af2}
\end{figure}

\begin{figure}[ht]
\centering
\includegraphics[width=0.7\textwidth]{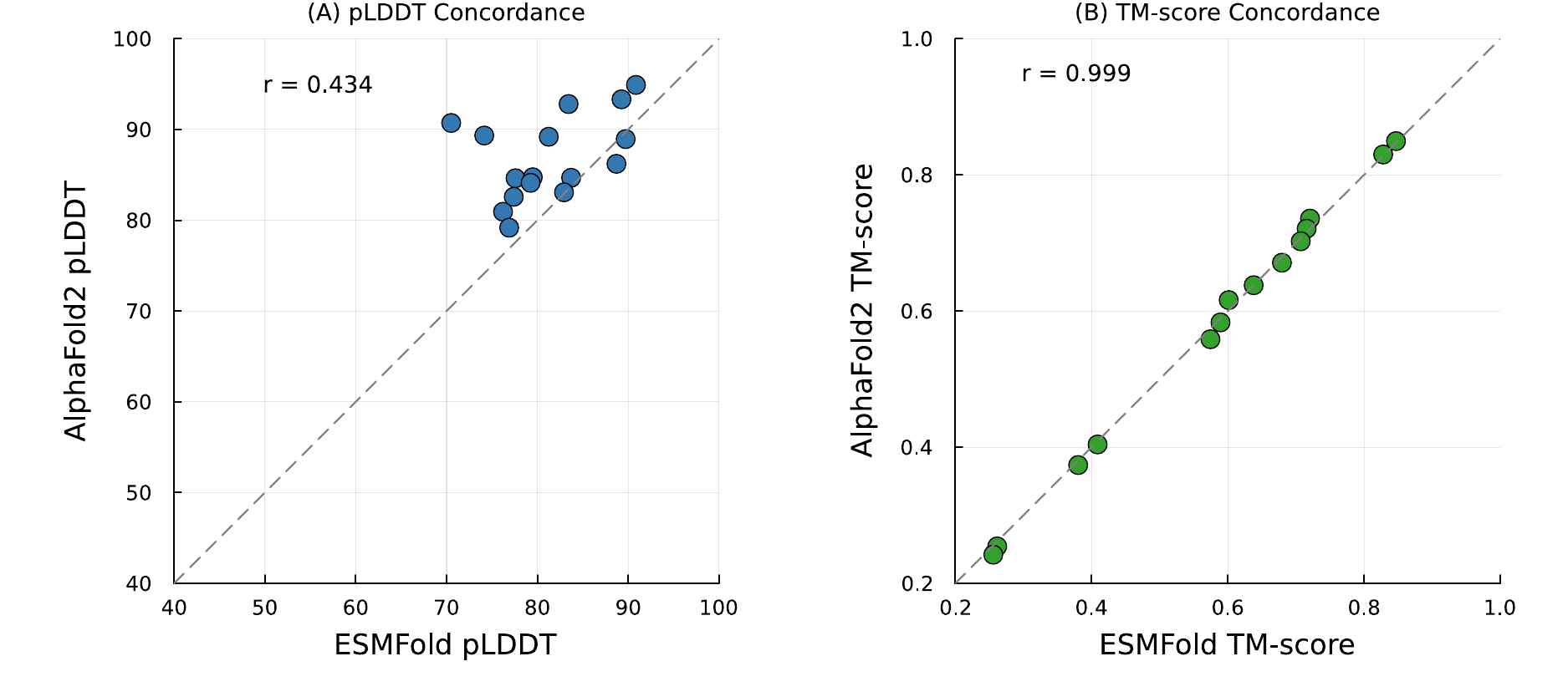}
\caption{Concordance between ESMFold and AlphaFold2 structure predictions for SA-generated sequences across all eight families. Each point is one family-condition pair. \textbf{(A)}~pLDDT concordance ($r{=}0.434$); AlphaFold2 pLDDT values are systematically higher, reflecting differences in confidence calibration. \textbf{(B)}~TM-score concordance ($r{=}0.999$); points lie near the identity line, indicating strong agreement in condition ranking.}
\label{fig:esmfold-af2-concordance}
\end{figure}

Figure~\ref{fig:structure-esmfold-af2} shows predicted local distance difference test (pLDDT) and template modeling (TM)-score results from both ESMFold and AlphaFold2 side by side across all eight families. The two predictors produce similar qualitative patterns: sequences generated by stochastic attention achieve pLDDT above the 70 threshold in every family under both predictors, and the family-to-family ordering of TM-scores is preserved. Template modeling scores were significantly higher for SA generation than for stored sequences in all eight families ($p < 0.05$; Cohen's $d{=}0.47$--$1.73$). Figure~\ref{fig:esmfold-af2-concordance} quantifies the agreement directly: TM-scores are concordant across predictors (Pearson $r{=}0.999$), supporting the conclusion that the structural ranking of conditions is not specific to ESMFold's architecture or training data. The pLDDT concordance is more moderate ($r{=}0.434$), reflecting the fact that each predictor has its own confidence calibration scale; AlphaFold2 pLDDT values are systematically higher than ESMFold values for the same sequences, consistent with known differences in the two models' confidence scoring. The TM-score concordance supports the structural conclusions drawn from ESMFold in the main text.

\subsection*{Reference structures}
Table~\ref{tab:pdb-refs} lists the experimentally determined structures used as TM-align references for each family. References are canonical, high-resolution structures of the cognate domain.

\begin{table}[ht]
\centering
\caption{Experimentally determined reference structures used for TM-align scoring.}
\label{tab:pdb-refs}
\small
\begin{tabular}{@{}llll@{}}
\toprule
Family & Pfam ID & PDB (chain) & Reference protein \\
\midrule
RRM            & PF00076 & 1FXL (A) & Sex-lethal RRM \\
SH3            & PF00018 & 1SHG (A) & $\alpha$-spectrin SH3 domain \\
WW             & PF00397 & 1PIN (A) & Pin1 WW domain \\
Kunitz         & PF00014 & 1BPI (A) & BPTI (Kunitz domain) \\
zf-C2H2        & PF00096 & 1ZAA (C) & Zif268 zinc fingers \\
PDZ            & PF00595 & 1BE9 (A) & PSD-95 PDZ3 \\
Pkinase        & PF00069 & 1ATP (E) & PKA catalytic subunit \\
Defensin\_beta & PF00711 & 1E4S (A) & Beta-defensin \\
\bottomrule
\end{tabular}
\end{table}

\section{Consensus Proximity and Predicted Structural Quality}\label{si:consensus-reg}

Sequences generated by stochastic attention achieved template modeling (TM)-scores that, in several families, exceeded those of the stored natural sequences. Because all structure predictors are trained on evolutionary data, one explanation is predictor bias: a predictor might assign higher confidence to sequences that lie closer to the family consensus, in which case the elevated TM-scores would be an artifact of consensus proximity rather than evidence of structural quality. To test this directly, we asked whether predicted structural quality tracks consensus proximity \emph{within} each family. For every SA-generated sequence whose structure was predicted ($n{=}50$ per family), we computed its identity to the family consensus (the most frequent residue per column among stored sequences) along with its mean predicted local distance difference test (pLDDT) score and reference-normalized TM-score. We then computed the within-family Spearman rank correlation between consensus identity and each structural metric (Table~\ref{tab:consensus-reg}).

The association is weak and inconsistent in sign. Across families, the Spearman correlation between consensus identity and TM-score ranged from $-0.13$ to $0.53$ (negative in SH3, WW, and Pkinase), and between consensus identity and pLDDT from $-0.10$ to $0.43$. Pooling the within-family-standardized values gave $\rho{=}0.17$ for TM-score ($p{=}5\times10^{-4}$, $n{=}400$) and $\rho{=}0.23$ for pLDDT ($p{=}2\times10^{-6}$); each accounts for under $4\%$ and $6\%$ of the variance, respectively. Consensus distance is therefore not the dominant determinant of predicted structural quality, which argues against the interpretation that the elevated TM-scores are primarily an artifact of consensus proximity. This analysis does not exclude an \emph{absolute} preference, shared by both the sampler and the predictor, for family-typical sequences; a complementary control that folds the consensus-with-noise ensemble through the same predictors would further isolate that effect, and definitive resolution requires experimental structure determination. We also note that SA generation is most consensus-proximal in the Pkinase family (mean identity to consensus $0.73$, the highest of any family), which accounts for both its near-natural nearest-neighbor identity and its low mutual-information correlation ($r{=}0.05$; Table~\ref{tab:mi}). Near-consensus sequences carry little pairwise covariation.

\begin{table}[ht]
\centering
\caption{Within-family Spearman correlation between a generated sequence's identity to the family consensus and its predicted structural quality (ESMFold pLDDT and reference-normalized TM-score), with the mean identity to consensus. $n{=}50$ generated sequences per family.}
\label{tab:consensus-reg}
\small
\begin{tabular}{@{}lccc@{}}
\toprule
Family & $\rho$(consID, pLDDT) & $\rho$(consID, TM) & mean consID \\
\midrule
RRM            & $0.21$  & $0.36$  & $0.60$ \\
SH3            & $0.30$  & $-0.13$ & $0.58$ \\
WW             & $0.29$  & $-0.07$ & $0.49$ \\
Kunitz         & $0.17$  & $0.28$  & $0.56$ \\
zf-C2H2        & $0.34$  & $0.18$  & $0.44$ \\
PDZ            & $0.43$  & $0.53$  & $0.56$ \\
Pkinase        & $-0.10$ & $-0.09$ & $0.73$ \\
Defensin\_beta & $0.28$  & $0.36$  & $0.58$ \\
\midrule
Pooled (standardized) & $0.23$ & $0.17$ & n/a \\
\bottomrule
\end{tabular}
\end{table}

\section{Sequence-Level Analysis}\label{si:sequence-analysis}

\begin{figure}[ht]
\centering
\includegraphics[width=\textwidth]{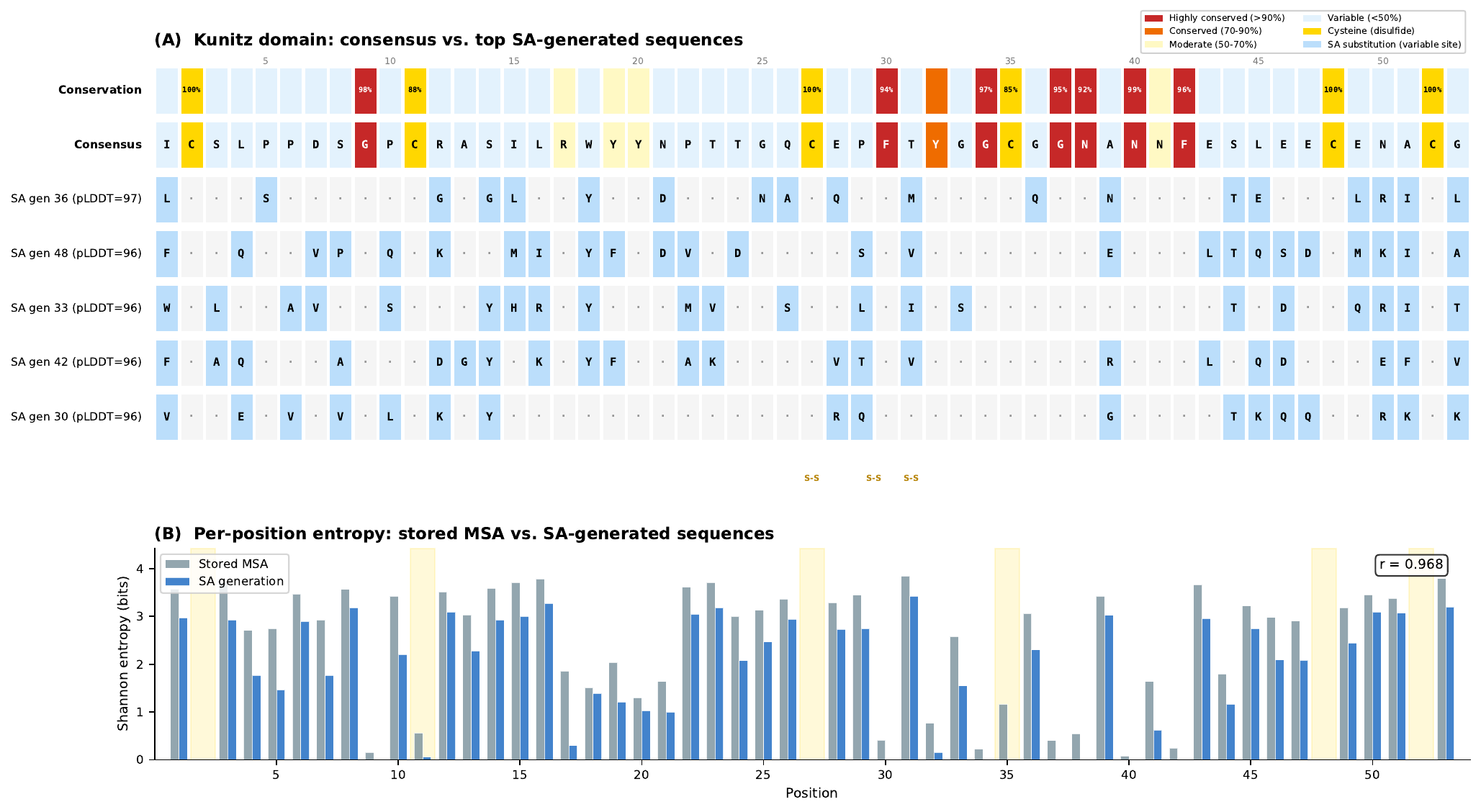}
\caption{Sequence-level analysis of SA-generated Kunitz domain sequences. \textbf{(A)}~Alignment colored by conservation level. \textbf{(B)}~Per-position Shannon entropy correlation between the SA-generated ensemble and the stored seed alignment ($r{=}0.968$).}
\label{fig:sequence-analysis}
\end{figure}

We aligned the five highest-confidence SA-generated Kunitz domain sequences against the family consensus to examine how the method handles the interplay between conservation and variation at the single-residue level (Fig.~\ref{fig:sequence-analysis}). All five sequences preserved every highly conserved position (11 of 11 positions with $>90\%$ conservation in the seed alignment) and all six cysteines that form the three disulfide bonds essential to the Kunitz fold, while introducing 17--25 substitutions concentrated at variable positions. Per-position Shannon entropy correlated between the SA-generated ensemble and the stored multiple sequence alignment (MSA; $r{=}0.968$), indicating that the method distinguishes conserved positions from positions that tolerate variation. The concentration of substitutions at high-entropy positions is consistent with the Boltzmann sampling mechanism: the softmax attention weights distribute probability mass across stored patterns in proportion to their similarity to the current sample, so strongly conserved positions (where all stored patterns agree) are decoded to the consensus residue, while variable positions (where stored patterns diverge) sample from the local distribution of amino acids.

\section{Pairwise Mutual Information Analysis}\label{si:mi}

To assess whether SA-generated sequences preserve pairwise residue covariation, we computed mutual information (MI) matrices for the stored seed alignment and for sequences generated by each method. For each family, we constructed the $L \times L$ MI matrix, where position pair $(i, j)$ has MI defined as $\mathrm{MI}(i,j) = \sum_{a,b} p(a,b) \log\frac{p(a,b)}{p_i(a)\, p_j(b)}$, with an additive pseudocount of 1.0 to regularize sparse counts. We then extracted the upper-triangle elements ($L(L{-}1)/2$ pairs) and computed Pearson and Spearman correlations between the stored MI vector and the MI vector of each generated ensemble. Table~\ref{tab:mi} reports the results across all eight families.

\begin{figure}[ht]
\centering
\includegraphics[width=\textwidth]{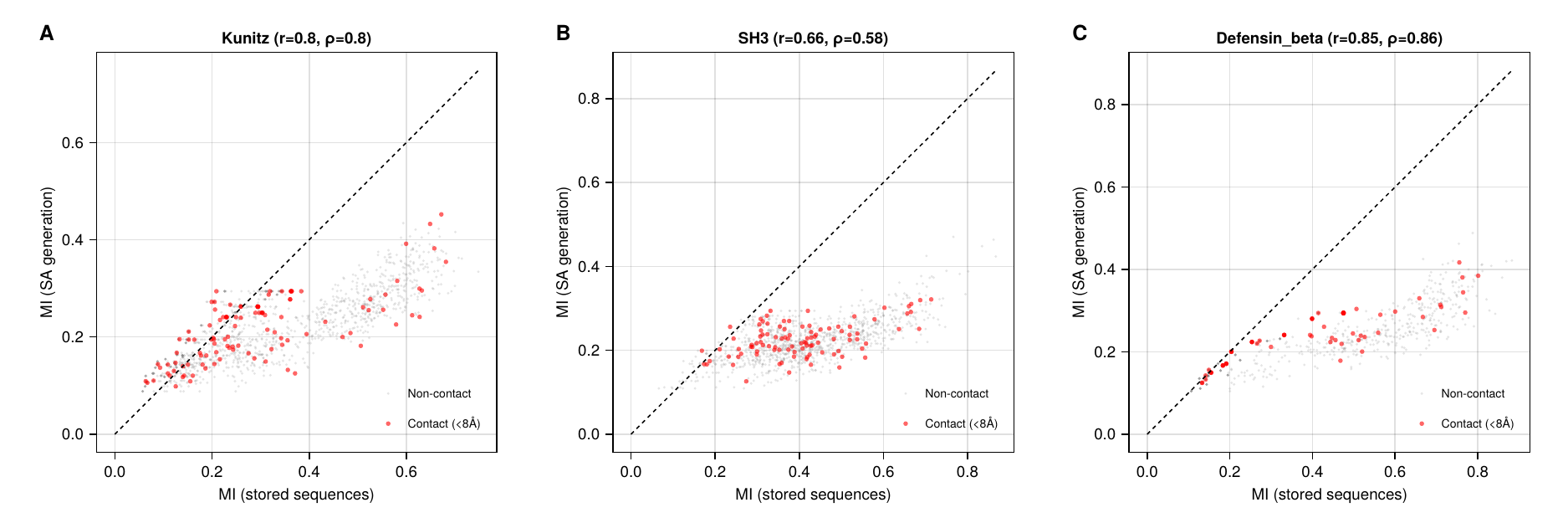}
\caption{Pairwise MI scatter plots for representative families. Each point represents one position pair $(i, j)$; the $x$-axis is MI from the stored seed alignment and the $y$-axis is MI from SA-generated sequences. Red points indicate structurally contacting residue pairs (C$\alpha$--C$\alpha$ distance ${<}8$~\AA, sequence separation ${\geq}5$ residues). Pearson $r$ and Spearman $\rho$ are shown in each panel.}
\label{fig:mi-scatter}
\end{figure}

\begin{figure}[ht]
\centering
\includegraphics[width=\textwidth]{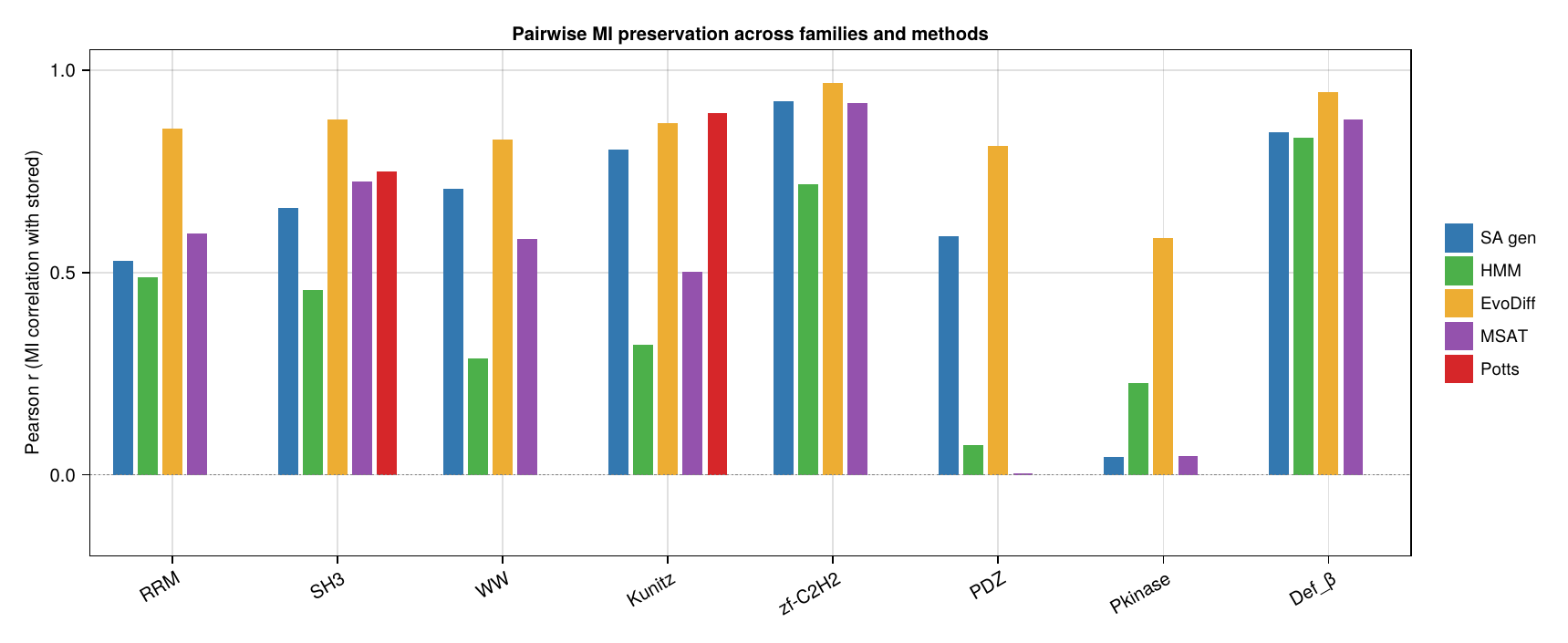}
\caption{Pairwise MI preservation across all eight Pfam families. Each bar shows the Pearson correlation between the upper-triangle MI vector of the stored seed alignment and the MI vector of sequences generated by the indicated method. The general ordering HMM $<$ SA gen $<$ Potts $\leq$ EvoDiff reflects the information hierarchy of each model in several families, though Defensin\_beta is an exception where the Potts model ($r{=}0.23$) falls below both SA and HMM. On Pkinase, the shallowest family ($K{=}37$, $L{=}262$), SA generation is consensus-proximal and so carries little covariation ($r{=}0.05$), with HMM and MSAT similarly low; Potts ($r{=}0.95$) and EvoDiff recover the covariation through explicitly fitted couplings and large-scale pretraining.}
\label{fig:mi-barplot}
\end{figure}

SA-generated sequences preserved much of the pairwise covariation structure of the stored alignment, with Pearson $r{=}0.53$--$0.92$ in seven of eight families (mean $r{=}0.72$ excluding Pkinase). In those seven families, SA outperformed profile hidden Markov model (HMM) sequences, which model each position independently (mean HMM $r{=}0.42$). The Pkinase family ($K{=}37$, $L{=}262$) was the lone exception for SA ($r{=}0.05$). Its alignment is shallow, but limited estimation is not the explanation, because the Potts model recovers the covariation on the identical alignment ($r{=}0.95$). The outlier instead reflects SA generation being most consensus-proximal in this family (mean identity to the consensus $0.73$, the highest of any family); near-consensus sequences vary little from one another and so carry little pairwise covariation. Methods that explicitly model pairwise couplings (Potts/plmDCA: $r{=}0.65$--$0.96$ in seven of eight families, with Defensin\_beta an outlier at $r{=}0.23$) or leverage pretraining on millions of MSAs (EvoDiff: $r{=}0.59$--$0.97$) generally achieved higher MI correlations, as expected given their direct access to pairwise or higher-order statistics. The ordering HMM $<$ SA $<$ Potts $\leq$ EvoDiff held in three of eight families. This ordering is consistent with the information each method encodes: position-independent frequencies for HMMs, the implicit all-order couplings of the Hopfield energy for SA, explicitly fitted pairwise couplings for Potts, and deep coevolutionary priors from large-scale pretraining for EvoDiff. The Defensin\_beta exception (Potts $r{=}0.23$, below both SA and HMM) reflects the same compositional limitation observed in the generation metrics (Table~\ref{tab:potts}). Figure~\ref{fig:mi-barplot} shows this ordering across all eight families simultaneously, and Fig.~\ref{fig:mi-scatter} shows MI scatter plots for representative families, with structural contact pairs highlighted to assess whether SA preserves covariation at structurally relevant positions.

\begin{table}[ht]
\centering
\caption{Pairwise mutual information (MI) preservation: Pearson correlation $r$ between upper-triangle MI vectors of stored sequences and sequences generated by each method. Higher values indicate better preservation of pairwise residue covariation. $^*$MSAT: MSA Transformer.}
\label{tab:mi}
\small
\begin{tabular}{@{}lcccccc@{}}
\toprule
Family & SA gen & SA ret & Potts & HMM & EvoDiff & MSAT$^*$ \\
\midrule
RRM            & $0.53$ & $0.34$ & $0.65$  & $0.49$ & $0.86$ & $0.60$ \\
SH3            & $0.66$ & $0.64$ & $0.75$  & $0.46$ & $0.88$ & $0.73$ \\
WW             & $0.71$ & $0.49$ & $0.88$  & $0.29$ & $0.83$ & $0.58$ \\
Kunitz         & $0.80$ & $0.69$ & $0.89$  & $0.32$ & $0.87$ & $0.50$ \\
zf-C2H2        & $0.92$ & $0.87$ & $0.96$  & $0.72$ & $0.97$ & $0.92$ \\
PDZ            & $0.59$ & $0.44$ & $0.94$  & $0.07$ & $0.81$ & $0.01$ \\
Pkinase        & $0.05$ & $-0.09$ & $0.95$ & $0.23$ & $0.59$ & $0.05$ \\
Defensin\_beta & $0.85$ & $0.75$ & $0.23$  & $0.83$ & $0.95$ & $0.88$ \\
\bottomrule
\end{tabular}
\end{table}

\section{Biophysical Properties of Beta Defensin Sequences}\label{si:defensin-biophysics}

\begin{figure}[ht]
\centering
\includegraphics[width=0.9\textwidth]{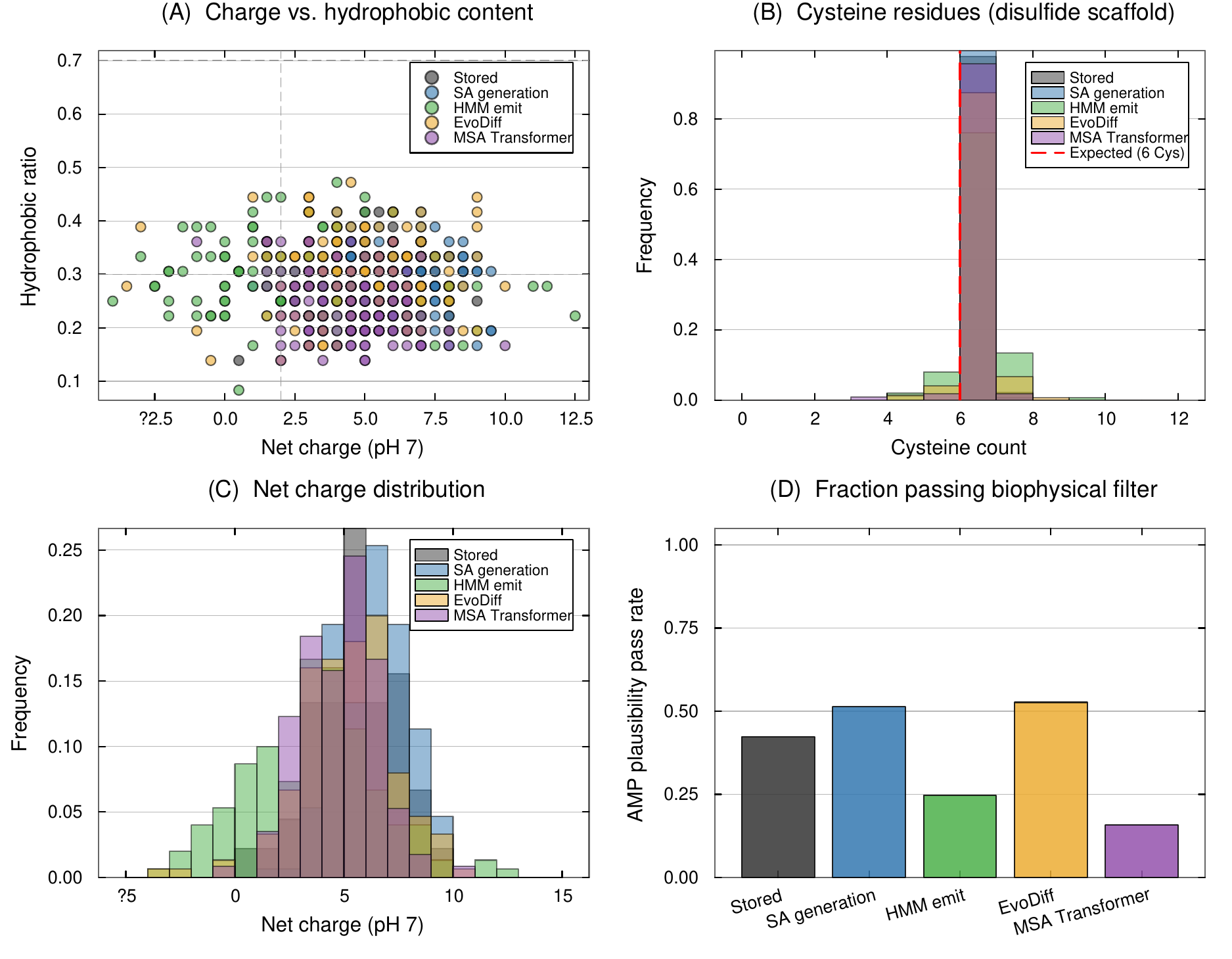}
\caption{Biophysical properties of beta defensin sequences (PF00711). \textbf{(A)}~Net charge at pH~7 vs.\ hydrophobic residue fraction. Sequences generated by stochastic attention (blue) overlapped the stored natural sequences (gray), while HMM emit (green) and MSA Transformer (purple) drifted to lower charge. Dashed lines indicate the antimicrobial plausibility thresholds (charge $\geq +2$, hydrophobic fraction $0.3$--$0.7$). \textbf{(B)}~Cysteine count distribution. Stochastic attention generation preserved the six-cysteine disulfide scaffold (red dashed line) with the same fidelity as the stored alignment. \textbf{(C)}~Net charge distribution. Stochastic attention generation produced a charge profile closely matching stored sequences. \textbf{(D)}~Fraction of sequences passing a minimal biophysical plausibility filter (charge $\geq +2$, hydrophobic fraction $\in [0.3, 0.7]$, $\geq 4$ cysteines).}
\label{fig:defensin-biophysics}
\end{figure}

Figure~\ref{fig:defensin-biophysics} shows the biophysical properties of SA-generated beta defensin sequences (PF00711) compared to stored natural sequences and baseline methods. Sequences generated by stochastic attention preserved the six-cysteine disulfide scaffold, maintained a cationic charge profile, and occupied the same region of charge-hydrophobicity space as natural defensins (Fig.~\ref{fig:defensin-biophysics}A--C). The HMM emit and MSA Transformer baselines, by contrast, drifted toward lower net charge, away from the cationic regime that natural defensins occupy. Consistent with this preserved biophysical profile, a larger fraction of SA-generated sequences than of HMM- or MSA-Transformer-generated sequences passed a minimal biophysical plausibility filter (net charge $\geq +2$, hydrophobic fraction in $[0.3, 0.7]$, and at least four cysteines; Fig.~\ref{fig:defensin-biophysics}D).

\section{Independent Fitness Validation via ESM2}\label{si:esm2}

As an independent assessment of whether SA-generated sequences resemble natural proteins, we scored all generated and stored sequences using ESM2-650M~\cite{linEvolutionaryScaleModeling2023}, a protein language model pretrained on millions of UniRef50 sequences that is entirely independent of the SA framework. For each sequence, we computed the pseudo-perplexity via masked marginal scoring: each position was masked in turn, the model predicted a distribution over amino acids, and the log-probability of the true residue was recorded. The pseudo-perplexity (lower is more protein-like) is $\exp(-\frac{1}{L}\sum_{\ell=1}^{L} \log P(a_\ell \mid \mathbf{a}_{\setminus \ell}))$, where $a_\ell$ is the amino acid at position $\ell$ and $\mathbf{a}_{\setminus \ell}$ denotes all other positions. Wilcoxon rank-sum tests with Cohen's $d$ effect sizes compared SA-generated sequences ($n{=}150$) to stored natural sequences ($n{=}37$--$420$) for each family (Table~\ref{tab:esm2}).

Sequences generated by stochastic attention were statistically indistinguishable from natural family members in four of eight families (Wilcoxon $p > 0.05$; RRM, zf-C2H2, Pkinase, Defensin\_beta: $p > 0.29$). In the Kunitz domain, SA generation achieved significantly \emph{lower} (better) pseudo-perplexity than stored sequences ($4.47$ vs.\ $4.98$, $p < 0.001$, $d{=}{-}0.74$). Three families showed modestly higher pseudo-perplexity for SA generation (SH3: $d{=}0.60$; WW: $d{=}1.03$; PDZ: $d{=}0.60$), though all SA-generated sequences remained within the range of natural protein sequences. The overall mean pseudo-perplexity across all families was $5.71$ for SA-generated sequences versus $5.58$ for stored sequences, indicating that an independent protein language model trained on billions of sequences scores SA-generated sequences comparably to natural family members.

\begin{table}[ht]
\centering
\caption{ESM2-650M pseudo-perplexity validation. Lower pseudo-perplexity indicates a more protein-like sequence. $p$: Wilcoxon rank-sum test (two-sided). $d$: Cohen's effect size (negative values indicate SA is better). Significance: $^{***}p < 0.001$, $^{**}p < 0.01$, n.s.\ = not significant ($p > 0.05$).}
\label{tab:esm2}
\small
\begin{tabular}{@{}lccccl@{}}
\toprule
Family & SA gen ppl & Stored ppl & $p$ & $d$ & Sig. \\
\midrule
RRM            & $4.87 \pm 0.68$ & $5.20 \pm 1.92$ & $0.30$    & $-0.29$ & n.s. \\
SH3            & $5.24 \pm 0.82$ & $4.62 \pm 1.48$ & $<0.001$  & $0.60$  & $^{***}$ \\
WW             & $6.78 \pm 1.77$ & $5.04 \pm 1.36$ & $<0.001$  & $1.03$  & $^{***}$ \\
Kunitz         & $4.47 \pm 0.57$ & $4.98 \pm 0.94$ & $<0.001$  & $-0.74$ & $^{***}$ \\
zf-C2H2        & $7.26 \pm 1.66$ & $8.10 \pm 3.47$ & $0.41$    & $-0.37$ & n.s. \\
PDZ            & $6.30 \pm 1.21$ & $5.31 \pm 2.61$ & $0.002$   & $0.60$  & $^{**}$ \\
Pkinase        & $4.44 \pm 0.32$ & $4.32 \pm 0.80$ & $0.39$    & $0.27$  & n.s. \\
Defensin\_beta & $6.35 \pm 1.27$ & $6.82 \pm 2.09$ & $0.32$    & $-0.31$ & n.s. \\
\bottomrule
\end{tabular}
\end{table}

\section{Deep Mutational Scanning Validation}\label{si:dms}

To assess whether SA-generated substitutions correspond to experimentally tolerated mutations, we cross-referenced the amino acid changes introduced by SA generation against published deep mutational scanning (DMS) datasets for two families: the PDZ domain (DLG4/PSD-95 PDZ3; McLaughlin et al.~\cite{mclaughlinSpatialArchitecture2012}) and the SH3 domain (GRB2 SH3; Faure et al.~\cite{faureDoubleDMS2022}). For each family, we identified the DMS reference protein in the Pfam seed alignment, mapped alignment positions to DMS coordinates, and classified each SA-generated substitution relative to the family consensus as experimentally tolerated ($\mathrm{DMS\_score\_bin}{=}1$) or deleterious ($\mathrm{DMS\_score\_bin}{=}0$) based on the published fitness data (Table~\ref{tab:dms}).

Substitutions generated by stochastic attention were enriched for experimentally tolerated mutations in both families. In the PDZ domain, $91.7\%$ of the $4{,}313$ matched substitutions were classified as tolerated, compared to an overall tolerance rate of $77.5\%$ in the DMS dataset ($1.18\times$ enrichment, binomial $p < 10^{-134}$). The continuous DMS fitness scores of SA-generated substitutions were also significantly higher than those of random single mutations (Wilcoxon rank-sum $p < 10^{-20}$, Cohen's $d{=}0.42$). In the SH3 domain, the enrichment was larger: $90.4\%$ of $2{,}396$ matched substitutions were tolerated versus $67.5\%$ overall ($1.34\times$ enrichment, binomial $p < 10^{-155}$), with a larger effect size on continuous fitness scores ($d{=}0.70$, $p < 10^{-54}$). These results indicate that the Boltzmann sampling procedure preferentially generates substitutions at positions and to residues that are individually tolerated in single-mutant assays, without any explicit fitness objective or training signal.

\begin{table}[ht]
\centering
\caption{Deep mutational scanning validation. Substitutions generated by stochastic attention relative to the family consensus were matched to published single-mutant DMS fitness data. The tolerance rate is the fraction of matched substitutions classified as tolerated. The DMS overall column gives the baseline tolerance rate across all single mutations in the dataset. Enrichment is the ratio of the SA tolerance rate to the DMS overall rate.}
\label{tab:dms}
\small
\begin{tabular}{@{}lcccccc@{}}
\toprule
Family & DMS source & Matched & SA tol.\ rate & DMS overall & Enrichment & $p$ \\
\midrule
PDZ & McLaughlin et al.\ 2012 & 4{,}313 & 91.7\% & 77.5\% & $1.18\times$ & $<10^{-134}$ \\
SH3 & Faure et al.\ 2022      & 2{,}396 & 90.4\% & 67.5\% & $1.34\times$ & $<10^{-155}$ \\
\bottomrule
\end{tabular}
\end{table}

A potential confound in the comparison above is that SA preferentially substitutes at variable positions, which tend to be more mutation-tolerant; the enrichment relative to the overall dataset rate could then reflect \emph{where} SA mutates rather than \emph{which} residue it selects. To separate these effects, we recomputed the enrichment against a position-matched null. For each matched substitution, the null tolerance probability is the fraction of all tested mutant amino acids at that same DMS position that are classified as tolerated. Summing these per-substitution probabilities gives the number of tolerated substitutions expected if SA had chosen a random tested residue at each position it mutated, holding the position distribution fixed. The position-matched null rate is indeed higher than the overall rate in both families ($0.843$ for PDZ and $0.749$ for SH3, versus $0.775$ and $0.675$ overall), confirming that SA does target tolerant positions. Even against this stricter null, however, SA-generated substitutions remained significantly enriched for tolerated mutations: $91.7\%$ versus $84.3\%$ for PDZ ($1.09\times$; Poisson-binomial $z{=}18.6$, $p < 10^{-77}$; $p < 5\times10^{-5}$ by a $20{,}000$-draw permutation test) and $90.4\%$ versus $74.9\%$ for SH3 ($1.21\times$; $z{=}23.4$, $p < 10^{-121}$). A coarser conservation-matched null, which draws comparison substitutions from positions in the same alignment-conservation quintile, gave essentially identical null rates ($0.841$ and $0.744$) and the same conclusion (Table~\ref{tab:dms-null}). The residue choices made by SA are therefore enriched for experimentally tolerated mutations beyond what its preference for tolerant positions alone would produce, although the enrichment is more modest than the comparison against the overall dataset rate suggests.

\begin{table}[ht]
\centering
\caption{Position-matched and conservation-matched DMS nulls. The position-matched null fixes the set of positions SA substituted and asks how often a randomly chosen \emph{tested} substitution at each such position is tolerated; the conservation-matched null draws comparison substitutions from positions in the same alignment-conservation quintile. Enrichment is the ratio of the observed SA tolerance rate to the null rate. The position-matched $p$ is from a Poisson-binomial $z$-test (one-sided, SA $>$ null) and is corroborated by a $20{,}000$-draw permutation test ($p < 5\times10^{-5}$ in both families).}
\label{tab:dms-null}
\small
\begin{tabular}{@{}lccccc@{}}
\toprule
Family & SA tol.\ rate & Overall null & Position-matched null & Conservation-matched null & Pos.-matched $p$ \\
\midrule
PDZ & $0.917$ ($n{=}4{,}313$) & $0.775$ ($1.18\times$) & $0.843$ ($1.09\times$) & $0.841$ ($1.09\times$) & $<10^{-77}$ \\
SH3 & $0.904$ ($n{=}2{,}396$) & $0.675$ ($1.34\times$) & $0.749$ ($1.21\times$) & $0.744$ ($1.22\times$) & $<10^{-121}$ \\
\bottomrule
\end{tabular}
\end{table}

\section{Potts Model (plmDCA) Baseline}\label{si:potts}

To assess whether pairwise coevolutionary models provide a stronger baseline than profile HMMs, we fitted a Potts model via pseudo-likelihood maximization (plmDCA)~\cite{ekeberg2013improved} to all eight Pfam seed alignments and generated 150 sequences per family by Gibbs sampling (500 sweeps per chain, initialized from random stored sequences). The Potts model captures pairwise couplings $J_{ij}(a,b)$ between all position pairs in addition to single-site fields $h_i(a)$, giving it more parameters than a profile HMM. The number of coupling parameters scales as $L(L{-}1)q^2/2$, which yields data-to-parameter ratios ranging from $2.5 \times 10^{-6}$ (Pkinase) to $2.0 \times 10^{-3}$ (WW). Despite this high parameter count, L2 regularization ($\lambda_J{=}\lambda_h{=}0.01$) yielded well-behaved generation in most families (Table~\ref{tab:potts}).

Across the eight families, the Potts model achieved comparable or better amino acid composition fidelity than SA in six of eight families, consistent with its explicit fitting of positional statistics. However, SA produced higher novelty in five of eight families, indicating that the Hopfield energy landscape supports broader exploration of sequence space than Gibbs sampling from a fitted Potts distribution. Both methods occupy an intermediate regime between profile HMMs (which produced lower family identity) and bootstrap resampling (which has zero novelty). Two families highlighted limitations for the Potts model. In Pkinase ($K{=}37$, $L{=}262$), the high data-to-parameter ratio led to low novelty ($0.22$ vs.\ $0.48$ for SA) despite good compositional fidelity, suggesting the Gibbs sampler remained near stored patterns. In Defensin\_beta ($K{=}45$, $L{=}36$), the Potts model produced the highest KL divergence of any method (KL$=$0.112), possibly reflecting sensitivity to the cysteine-rich, low-entropy character of this family. The main distinction between the two approaches is that the Potts model required fitting hundreds of thousands to millions of parameters via L-BFGS-B optimization, followed by Gibbs sampling (total wall time: $0.5$--$52$ minutes per family), while SA required no parameter fitting and generated sequences in seconds. The broadly similar output quality across the remaining six families supports the conclusion that some of the correlational structure captured by pairwise couplings is also present in the Hopfield energy landscape, which implicitly encodes higher-order correlations through the softmax attention operation.

\begin{table}[ht]
\centering
\caption{Potts model (plmDCA) baseline comparison across all eight Pfam families. Metrics are computed identically to the main baseline analysis (30 chains $\times$ 5 samples). Standard errors are computed over $30$ sub-blocks of five sequences (as in Table~\ref{tab:baseline-metrics}); consequently the KL SE for SA generation differs from the $1{,}000$-fold bootstrap KL SE reported in Table~\ref{tab:results}, although the SA-generation point estimates are identical between the two tables.}
\label{tab:potts}
\small
\begin{tabular}{@{}llccc@{}}
\toprule
Family & Method & KL$_{\mathrm{AA}}$ $(\downarrow)$ & Novelty $(\uparrow)$ & SeqID \\
\midrule
\multirow{2}{*}{RRM}
 & SA generation    & $0.060 \pm 0.036$ & $0.513 \pm 0.011$ & $0.538 \pm 0.006$ \\
 & Potts (plmDCA)   & $0.018 \pm 0.015$ & $0.562 \pm 0.008$ & $0.396 \pm 0.011$ \\
\midrule
\multirow{2}{*}{SH3}
 & SA generation    & $0.036 \pm 0.026$ & $0.471 \pm 0.012$ & $0.593 \pm 0.007$ \\
 & Potts (plmDCA)   & $0.025 \pm 0.019$ & $0.374 \pm 0.011$ & $0.539 \pm 0.009$ \\
\midrule
\multirow{2}{*}{WW}
 & SA generation    & $0.008 \pm 0.012$ & $0.653 \pm 0.006$ & $0.562 \pm 0.005$ \\
 & Potts (plmDCA)   & $0.011 \pm 0.030$ & $0.657 \pm 0.007$ & $0.436 \pm 0.011$ \\
\midrule
\multirow{2}{*}{Kunitz}
 & SA generation    & $0.013 \pm 0.022$ & $0.557 \pm 0.009$ & $0.607 \pm 0.004$ \\
 & Potts (plmDCA)   & $0.010 \pm 0.017$ & $0.540 \pm 0.010$ & $0.548 \pm 0.010$ \\
\midrule
\multirow{2}{*}{zf-C2H2}
 & SA generation    & $0.013 \pm 0.058$ & $0.596 \pm 0.010$ & $0.557 \pm 0.007$ \\
 & Potts (plmDCA)   & $0.005 \pm 0.043$ & $0.597 \pm 0.009$ & $0.512 \pm 0.008$ \\
\midrule
\multirow{2}{*}{PDZ}
 & SA generation    & $0.038 \pm 0.017$ & $0.418 \pm 0.009$ & $0.543 \pm 0.005$ \\
 & Potts (plmDCA)   & $0.012 \pm 0.007$ & $0.359 \pm 0.022$ & $0.302 \pm 0.019$ \\
\midrule
\multirow{2}{*}{Pkinase}
 & SA generation    & $0.035 \pm 0.001$ & $0.478 \pm 0.011$ & $0.515 \pm 0.004$ \\
 & Potts (plmDCA)   & $0.015 \pm 0.001$ & $0.222 \pm 0.018$ & $0.347 \pm 0.022$ \\
\midrule
\multirow{2}{*}{Defensin\_beta}
 & SA generation    & $0.022 \pm 0.058$ & $0.402 \pm 0.013$ & $0.655 \pm 0.007$ \\
 & Potts (plmDCA)   & $0.112 \pm 0.089$ & $0.215 \pm 0.011$ & $0.753 \pm 0.013$ \\
\bottomrule
\end{tabular}
\end{table}

\section{Sampling Diagnostics}\label{si:sampling-diagnostics}

To characterize the mixing properties of the Langevin sampler and justify the burn-in and thinning parameters used throughout the paper, we computed autocorrelation functions and effective sample sizes (ESS) from the Hopfield energy trace $E(\boldsymbol{\xi}_t)$ along each chain. For each family, we ran 10 independent unadjusted Langevin algorithm (ULA) and Metropolis-adjusted Langevin algorithm (MALA) chains of $T{=}5{,}000$ steps at the generation temperature $\beta_{\mathrm{gen}} \approx 2\beta^*$, recording the full trajectory. We computed the normalized autocorrelation function of the post-burn-in energy trace (iterations $2{,}001$--$5{,}000$), estimated the integrated autocorrelation time $\tau_{\mathrm{int}}$ using the initial positive sequence estimator with a cutoff at $\rho(\tau) < 0.05$, and derived the effective sample size as $\mathrm{ESS} = n/\tau_{\mathrm{int}}$ where $n{=}3{,}000$ is the number of post-burn-in samples.

Table~\ref{tab:sampling-diag} reports the results across all eight families. The integrated autocorrelation time ranged from $76$ to $122$ iterations for ULA and from $82$ to $131$ for MALA, yielding $32$--$42$ effective samples per chain (ULA) and $27$--$39$ per chain (MALA). The thinning interval of 100 is thus well matched to the autocorrelation time ($\mathrm{thin}/\tau_{\mathrm{int}} = 0.8$--$1.3$), yielding retained samples that are approximately independent. Across 30 chains, the total effective sample size is approximately $30 \times 35 \approx 1{,}050$ independent draws from the Boltzmann distribution, from which we retain 150 decoded sequences ($5$ per chain). The burn-in of $2{,}000$ iterations provides a $3.2$--$6.6\times$ margin over the empirical convergence point ($304$--$617$ iterations), supporting the use of this burn-in before sampling begins. The MALA acceptance rates ($99.6$--$99.8\%$) show that the Metropolis correction rejects fewer than $0.4\%$ of proposed moves at step size $\alpha{=}0.01$, so the unadjusted and adjusted updates rarely differ in this regime. A high acceptance rate bounds rather than directly quantifies the stationary-distribution bias of ULA; we therefore report it only as evidence that the two samplers take nearly identical steps at this step size, and adopt ULA for its computational efficiency.

\begin{table}[ht]
\centering
\caption{Sampling diagnostics for ULA and MALA chains across eight Pfam families. The quantity $\tau_{\mathrm{int}}$ is the integrated autocorrelation time of the Hopfield energy (mean over 10 chains). Effective sample size is reported per chain ($n{=}3{,}000$ post-burn-in samples); MALA AR is the Metropolis-adjusted acceptance rate; Conv.\ iter is the mean iteration at which burn-in energy converged (1\% criterion); and Margin is the ratio of burn-in budget (2{,}000) to convergence iteration.}
\label{tab:sampling-diag}
\small
\begin{tabular}{@{}lrrrrrrr@{}}
\toprule
Family & $\tau_{\mathrm{int}}^{\mathrm{ULA}}$ & ESS$^{\mathrm{ULA}}$ & $\tau_{\mathrm{int}}^{\mathrm{MALA}}$ & ESS$^{\mathrm{MALA}}$ & MALA AR & Conv.\ iter & Margin \\
\midrule
RRM            &  76 &  42 &  99 & 35 & 0.998 &  509 & $3.9\times$ \\
SH3            &  95 &  35 &  84 & 39 & 0.998 &  520 & $3.8\times$ \\
WW             &  86 &  37 &  96 & 34 & 0.996 &  369 & $5.4\times$ \\
Kunitz         & 100 &  35 & 109 & 33 & 0.997 &  500 & $4.0\times$ \\
zf-C2H2        & 105 &  33 &  82 & 39 & 0.997 &  447 & $4.5\times$ \\
PDZ            & 113 &  32 & 131 & 27 & 0.998 &  304 & $6.6\times$ \\
Pkinase        &  83 &  38 &  98 & 36 & 0.998 &  617 & $3.2\times$ \\
Defensin\_beta & 122 &  32 & 124 & 29 & 0.998 &  306 & $6.5\times$ \\
\bottomrule
\end{tabular}
\end{table}

\section{Initialization Sensitivity}\label{si:random-init}

The multi-chain protocol described in Materials and Methods initializes each Langevin chain near a randomly selected stored pattern ($\boldsymbol{\xi}_0 = \mathbf{m}_k + \sigma_{\mathrm{init}}\boldsymbol{\epsilon}$, $\sigma_{\mathrm{init}} = 0.01$). To test whether this choice affects the generated ensemble, we repeated the full SA generation pipeline with random initialization on the unit sphere: each chain was started from $\boldsymbol{\xi}_0 = \mathbf{z}/\|\mathbf{z}\|_2$ with $\mathbf{z} \sim \mathcal{N}(\mathbf{0}, \mathbf{I}_d)$, placing the initial point at a random location unrelated to any stored pattern. All other parameters were held fixed.

The results were consistent across initialization strategies: random sphere initialization produced output indistinguishable from stored-pattern initialization across all eight families (Table~\ref{tab:random-init}). The maximum absolute difference in KL divergence was $0.0003$ (RRM), and all other families showed differences below $10^{-4}$ in every metric. This insensitivity is expected from the theory: at $\beta_{\mathrm{gen}} \approx 2\beta^*$, the energy landscape has well-defined basins near stored patterns, and the dissipative drift term $-\alpha\nabla E(\boldsymbol{\xi})$ rapidly pulls any initial condition toward the nearest basin. The 2{,}000-step burn-in (at step size $\alpha{=}0.01$, corresponding to $20$ effective time units) is long relative to the observed convergence point. The result indicates that stored-pattern initialization is a computational convenience (reducing burn-in time) rather than a requirement, and that the generated ensemble is not an artifact of the initialization strategy.

\begin{table}[ht]
\centering
\caption{Initialization sensitivity: stored-pattern vs.\ random sphere initialization. All metrics are computed using the same procedure as in the main analysis (30 chains $\times$ 5 samples). The two strategies produce similar output across all eight families.}
\label{tab:random-init}
\small
\begin{tabular}{@{}llcccc@{}}
\toprule
Family & Init & KL$_{\mathrm{AA}}$ & Novelty & SeqID & MaxCos \\
\midrule
\multirow{2}{*}{RRM}
 & Stored & $0.060$ & $0.513$ & $0.538$ & $0.487$ \\
 & Random & $0.059$ & $0.511$ & $0.539$ & $0.489$ \\
\midrule
\multirow{2}{*}{SH3}
 & Stored & $0.036$ & $0.471$ & $0.593$ & $0.529$ \\
 & Random & $0.036$ & $0.471$ & $0.593$ & $0.529$ \\
\midrule
\multirow{2}{*}{WW}
 & Stored & $0.008$ & $0.653$ & $0.562$ & $0.347$ \\
 & Random & $0.008$ & $0.653$ & $0.562$ & $0.347$ \\
\midrule
\multirow{2}{*}{Kunitz}
 & Stored & $0.013$ & $0.557$ & $0.607$ & $0.443$ \\
 & Random & $0.013$ & $0.557$ & $0.607$ & $0.443$ \\
\midrule
\multirow{2}{*}{zf-C2H2}
 & Stored & $0.013$ & $0.596$ & $0.557$ & $0.404$ \\
 & Random & $0.013$ & $0.596$ & $0.557$ & $0.404$ \\
\midrule
\multirow{2}{*}{PDZ}
 & Stored & $0.038$ & $0.418$ & $0.543$ & $0.582$ \\
 & Random & $0.038$ & $0.418$ & $0.543$ & $0.582$ \\
\midrule
\multirow{2}{*}{Pkinase}
 & Stored & $0.035$ & $0.478$ & $0.515$ & $0.522$ \\
 & Random & $0.035$ & $0.478$ & $0.515$ & $0.522$ \\
\midrule
\multirow{2}{*}{Defensin\_beta}
 & Stored & $0.022$ & $0.402$ & $0.655$ & $0.598$ \\
 & Random & $0.022$ & $0.402$ & $0.655$ & $0.598$ \\
\bottomrule
\end{tabular}
\end{table}

\section{Near-Duplicate Analysis}\label{si:identity-distribution}

A key concern for any generative model operating on small datasets is whether the generated sequences are trivially close to stored patterns, effectively copying inputs with minor perturbations rather than producing novel outputs. To quantify the degree of near-duplication, we computed the full pairwise sequence identity between every SA-generated sequence ($S{=}150$) and every stored sequence ($K$) for each family, yielding a $150 \times K$ identity matrix. For each generated sequence, we recorded the maximum identity to any stored sequence (nearest neighbor) and the minimum number of amino acid substitutions.

Table~\ref{tab:identity-dist} reports the near-duplicate statistics across all eight families. Zero generated sequences across all eight families had $>95\%$ or $>90\%$ identity to any stored sequence, and zero sequences were within 2 substitutions of a stored pattern. Only in the Defensin\_beta family ($L{=}36$, the shortest alignment) did $6\%$ of generated sequences exceed $80\%$ identity to a stored sequence, which corresponds to at most 7 substitutions in a 36-residue sequence. The mean nearest-neighbor identity for generated sequences ($0.515$--$0.655$) was higher than the mean within-family pairwise identity ($0.217$--$0.368$), indicating that generated sequences were closer to individual stored patterns than stored patterns were to each other on average, but not trivially close. These results support the interpretation that SA generates sequences in intermediate regions of sequence space rather than copying stored patterns.

\begin{table}[ht]
\centering
\caption{Near-duplicate analysis: sequence identity between SA-generated and stored sequences. MaxID: mean maximum identity of each generated sequence to its nearest stored neighbor. \%$>$95, \%$>$90, \%$>$80: fraction of generated sequences exceeding the indicated identity threshold. \%$\leq$2sub: fraction within 2 substitutions of a stored sequence. Stored ID: mean pairwise identity among stored sequences.}
\label{tab:identity-dist}
\small
\begin{tabular}{@{}lcccccc@{}}
\toprule
Family & Mean MaxID & \%$>$95 & \%$>$90 & \%$>$80 & \%$\leq$2sub & Stored ID \\
\midrule
RRM            & $0.538$ & $0.0$ & $0.0$ & $0.0$ & $0.0$ & $0.230$ \\
SH3            & $0.593$ & $0.0$ & $0.0$ & $0.7$ & $0.0$ & $0.288$ \\
WW             & $0.562$ & $0.0$ & $0.0$ & $0.0$ & $0.0$ & $0.324$ \\
Kunitz         & $0.607$ & $0.0$ & $0.0$ & $0.0$ & $0.0$ & $0.368$ \\
zf-C2H2        & $0.557$ & $0.0$ & $0.0$ & $0.0$ & $0.0$ & $0.304$ \\
PDZ            & $0.543$ & $0.0$ & $0.0$ & $0.7$ & $0.0$ & $0.217$ \\
Pkinase        & $0.515$ & $0.0$ & $0.0$ & $0.0$ & $0.0$ & $0.274$ \\
Defensin\_beta & $0.655$ & $0.0$ & $0.0$ & $6.0$ & $0.0$ & $0.332$ \\
\bottomrule
\end{tabular}
\end{table}

The nearest sequence identity reported for generated sequences (``SeqID'') is a nearest-neighbor quantity: the identity of each generated sequence to its closest stored pattern. The appropriate natural reference is therefore the within-family nearest-neighbor identity, computed by leave-one-out as each stored sequence's identity to its closest neighbor in the same family, not the mean pairwise identity (``Stored ID'' above), which averages over all distant pairs and is consequently lower. Table~\ref{tab:nn-band} reports this nearest-neighbor envelope. Mean nearest-neighbor identity among natural sequences ranges from $0.38$ (Pkinase) to $0.60$ (WW), with $10$th--$90$th percentile envelopes spanning roughly $0.29$--$0.74$, above the $0.22$--$0.37$ mean pairwise identity. Generated SeqID ($0.51$--$0.66$) fell within each family's full nearest-neighbor range for seven of eight families; in Pkinase, generated SeqID ($0.515$) slightly exceeded the natural maximum ($0.454$), consistent with this family's consensus-proximal generation. By contrast, profile HMM emit and the MSA Transformer reached nearest-neighbor identities of $0.09$--$0.21$ in several families, below the minimum nearest-neighbor identity of any natural sequence in those families (e.g., $0.22$--$0.28$ for Pkinase, PDZ, and RRM), supporting the conclusion that their high novelty reflects drift outside the family rather than family-consistent exploration. All $150$ generated sequences were distinct in every family, with mean pairwise sequence identity among generated sequences of $0.35$--$0.65$ (mean pairwise Hamming distance $0.36$--$0.65$), a sequence-space diversity measure that does not rely on high-dimensional cosine distance.

\begin{table}[ht]
\centering
\caption{Empirical within-family nearest-neighbor identity envelope. For each family, $\bar{I}_{\mathrm{NN}}$ is the mean over stored sequences of each sequence's leave-one-out nearest-neighbor identity (identity to its closest neighbor in the same family), with standard deviation, $10$th/$90$th percentiles, and range. The ``SA SeqID'' column is the mean nearest-neighbor identity of generated sequences (from Table~\ref{tab:results}), and ``Gen.\ pair ID'' is the mean pairwise identity among the $150$ generated sequences. All $150$ generated sequences were distinct in every family.}
\label{tab:nn-band}
\small
\begin{tabular}{@{}lccccccc@{}}
\toprule
Family & $\bar{I}_{\mathrm{NN}}$ & SD & $p_{10}$ & $p_{90}$ & [min, max] & SA SeqID & Gen.\ pair ID \\
\midrule
RRM            & $0.397$ & $0.099$ & $0.292$ & $0.483$ & $[0.254, 0.803]$ & $0.538$ & $0.483$ \\
SH3            & $0.506$ & $0.099$ & $0.396$ & $0.625$ & $[0.292, 0.708]$ & $0.593$ & $0.476$ \\
WW             & $0.601$ & $0.074$ & $0.516$ & $0.677$ & $[0.387, 0.806]$ & $0.562$ & $0.376$ \\
Kunitz         & $0.568$ & $0.117$ & $0.449$ & $0.736$ & $[0.358, 0.792]$ & $0.607$ & $0.477$ \\
zf-C2H2        & $0.479$ & $0.048$ & $0.435$ & $0.522$ & $[0.304, 0.565]$ & $0.557$ & $0.347$ \\
PDZ            & $0.526$ & $0.179$ & $0.301$ & $0.711$ & $[0.217, 0.807]$ & $0.543$ & $0.445$ \\
Pkinase        & $0.381$ & $0.046$ & $0.325$ & $0.431$ & $[0.275, 0.454]$ & $0.515$ & $0.645$ \\
Defensin\_beta & $0.584$ & $0.122$ & $0.400$ & $0.722$ & $[0.333, 0.778]$ & $0.655$ & $0.470$ \\
\bottomrule
\end{tabular}
\end{table}

\section{Wall-Clock Timing Comparison}\label{si:timing}

Table~\ref{tab:timing} reports wall-clock generation times for 150 sequences across all eight families. Stochastic attention and profile HMM emit were timed on a single CPU core (Apple M2, 16~GB RAM) with all preprocessing included (PCA construction, phase transition detection, sampling, and decoding for SA; HMM building and sequence emission for HMM emit). EvoDiff and MSA Transformer times were measured on the same hardware (CPU only, no graphics processing unit), and Potts model (plmDCA) times include both parameter fitting and Gibbs sampling.

Stochastic attention generated 150 sequences in $0.2$--$4.6$~s across all eight families, with the variation driven primarily by the number of stored patterns $K$ (which determines the cost of the softmax attention at each Langevin step). The largest family (WW, $K{=}420$) required $4.6$~s, while all families with $K < 100$ completed in under $1$~s. Profile HMM emit was consistently the fastest method ($0.02$--$0.09$~s), as expected for a method that samples from a precomputed position-specific model with no iterative computation. EvoDiff required $2$--$14$ hours on CPU depending on sequence length (GPU inference would reduce these times, though EvoDiff's intended use case assumes GPU availability), and the MSA Transformer required $1$--$8$ hours for 50 rounds of iterative masked language model sampling. In these CPU-only runs, SA was $3$--$5$ orders of magnitude faster than EvoDiff and the MSA Transformer, and $2$--$3$ orders of magnitude faster than Potts model fitting, while requiring no GPU and no pretraining.

\begin{table}[ht]
\centering
\small
\caption{Wall-clock time for generating 150 sequences per family. Stochastic attention and HMM times are averaged over three runs on a single CPU core (Apple M2). EvoDiff and MSA Transformer times are from CPU-only runs. Potts times include pseudo-likelihood fitting and Gibbs sampling.}
\label{tab:timing}
\begin{tabular}{@{}lcrrrrrr@{}}
\toprule
Family & $L$ & $K$ & SA (s) & HMM (s) & EvoDiff (h) & MSAT (h) & Potts (min) \\
\midrule
RRM            & 71  & 68  & 0.5  & 0.05 & ${\sim}3$   & ${\sim}4$   & ${\sim}4$ \\
SH3            & 48  & 55  & 0.4  & 0.03 & ${\sim}2$   & ${\sim}2$   & ${\sim}2$ \\
WW             & 31  & 420 & 4.6  & 0.03 & ${\sim}2$   & ${\sim}3$   & ${\sim}1$ \\
Kunitz         & 53  & 99  & 0.7  & 0.03 & ${\sim}2$   & ${\sim}3$   & ${\sim}2$ \\
zf-C2H2        & 23  & 151 & 1.0  & 0.02 & ${\sim}2$   & ${\sim}1$   & ${\sim}1$ \\
PDZ            & 83  & 44  & 0.3  & 0.04 & ${\sim}4$   & ${\sim}4$   & ${\sim}5$ \\
Pkinase        & 262 & 37  & 0.2  & 0.09 & ${\sim}14$  & ${\sim}8$   & ${\sim}52$ \\
Defensin\_beta & 36  & 45  & 0.3  & 0.03 & ${\sim}2$   & ${\sim}1$   & ${\sim}1$ \\
\bottomrule
\end{tabular}
\end{table}

\section{Consensus-with-Noise Baseline}\label{si:consensus-baseline}

A natural concern is that SA may function as a ``consensus generator'' that merely perturbs the family consensus sequence. To test this directly, we implemented a control that adds calibrated noise to the consensus sequence in PCA space: for each family, we computed the position-wise consensus sequence (most frequent amino acid per column), one-hot encoded it, projected it into the same PCA space used by SA, and generated 150 sequences by adding isotropic Gaussian noise with standard deviation matched to the empirical spread of stored patterns in PCA space. We then decoded each noisy PCA vector via inverse PCA and argmax, and evaluated the resulting sequences using the same metrics as SA (Table~\ref{tab:consensus-baseline}).

The comparison revealed a characteristic signature that distinguishes SA from consensus perturbation. The consensus-with-noise baseline produced sequences with systematically higher novelty ($0.63$--$0.79$ vs.\ $0.40$--$0.65$ for SA) and lower sequence identity to stored patterns ($0.44$--$0.62$ vs.\ $0.52$--$0.66$), reflecting the fact that isotropic noise in PCA space scatters sequences away from all stored patterns rather than interpolating between them. The KL divergence pattern was mixed: the consensus baseline achieved lower KL in some families (e.g., RRM: $0.013$ vs.\ $0.060$) but higher KL in others (e.g., zf-C2H2: $0.095$ vs.\ $0.013$), showing that isotropic noise can either improve or degrade composition depending on whether the family's covariance aligns with the noise directions. The key distinction was that SA generation maintained higher sequence identity to stored patterns while achieving novelty, indicating that it explored the family's sequence space in a structured way that respects the correlational landscape, whereas consensus perturbation explored isotropically without regard to the family's covariance structure.

\begin{table}[ht]
\centering
\caption{Consensus-with-noise baseline vs.\ SA generation. The label CN denotes the consensus sequence with calibrated Gaussian noise in PCA space; SA denotes stochastic attention generation ($\beta \approx 2\beta^*$). Metrics are computed identically to the main analysis.}
\label{tab:consensus-baseline}
\small
\begin{tabular}{@{}llccc@{}}
\toprule
Family & Method & KL$_{\mathrm{AA}}$ & Novelty & SeqID \\
\midrule
\multirow{2}{*}{RRM}
 & CN & $0.013 \pm 0.002$ & $0.686 \pm 0.005$ & $0.487 \pm 0.005$ \\
 & SA gen & $0.060 \pm 0.005$ & $0.513 \pm 0.011$ & $0.538 \pm 0.006$ \\
\midrule
\multirow{2}{*}{SH3}
 & CN & $0.004 \pm 0.001$ & $0.658 \pm 0.005$ & $0.537 \pm 0.005$ \\
 & SA gen & $0.036 \pm 0.004$ & $0.471 \pm 0.012$ & $0.593 \pm 0.006$ \\
\midrule
\multirow{2}{*}{WW}
 & CN & $0.021 \pm 0.003$ & $0.787 \pm 0.002$ & $0.437 \pm 0.003$ \\
 & SA gen & $0.008 \pm 0.004$ & $0.653 \pm 0.006$ & $0.562 \pm 0.005$ \\
\midrule
\multirow{2}{*}{Kunitz}
 & CN & $0.004 \pm 0.001$ & $0.722 \pm 0.004$ & $0.551 \pm 0.003$ \\
 & SA gen & $0.013 \pm 0.002$ & $0.557 \pm 0.009$ & $0.607 \pm 0.004$ \\
\midrule
\multirow{2}{*}{zf-C2H2}
 & CN & $0.095 \pm 0.002$ & $0.742 \pm 0.003$ & $0.495 \pm 0.004$ \\
 & SA gen & $0.013 \pm 0.027$ & $0.596 \pm 0.010$ & $0.557 \pm 0.007$ \\
\midrule
\multirow{2}{*}{PDZ}
 & CN & $0.021 \pm 0.008$ & $0.633 \pm 0.005$ & $0.522 \pm 0.005$ \\
 & SA gen & $0.038 \pm 0.010$ & $0.418 \pm 0.009$ & $0.543 \pm 0.005$ \\
\midrule
\multirow{2}{*}{Pkinase}
 & CN & $0.015 \pm 0.001$ & $0.630 \pm 0.005$ & $0.528 \pm 0.004$ \\
 & SA gen & $0.035 \pm 0.001$ & $0.478 \pm 0.011$ & $0.515 \pm 0.004$ \\
\midrule
\multirow{2}{*}{Defensin\_beta}
 & CN & $0.013 \pm 0.002$ & $0.631 \pm 0.005$ & $0.620 \pm 0.006$ \\
 & SA gen & $0.022 \pm 0.003$ & $0.402 \pm 0.013$ & $0.655 \pm 0.007$ \\
\bottomrule
\end{tabular}
\end{table}

\section{Column-Permuted Alignment Control}\label{si:permuted-alignment}

To test directly whether SA captures information beyond marginal per-position amino acid frequencies, we constructed a control alignment for each family by independently shuffling each column. This procedure preserves the exact per-position amino acid frequency distribution while destroying all inter-position correlations (covariation, structural contacts, epistatic couplings). We then ran the full SA pipeline on the permuted alignment and compared the output against SA run on the real alignment (Table~\ref{tab:permuted-alignment}).

The results showed a consistent pattern across all eight families. Stochastic attention on the real alignment produced sequences with lower novelty but higher sequence identity to stored patterns compared to stochastic attention on the permuted alignment. This is expected: the real alignment contains inter-position correlations that constrain the energy landscape and keep generated sequences closer to family-typical regions, whereas the permuted alignment provides only position-independent frequency information, yielding an energy landscape that allows more isotropic exploration. The MI correlation differences between real and permuted SA were small but consistently favored the real alignment in seven of eight families (the exception being zf-C2H2, where the difference was $<0.01$), with the magnitude of the difference ranging from $0.01$ to $0.08$. The larger differences in novelty and sequence identity indicated that the inter-position correlations present in real alignments shaped the generated ensemble in ways that marginal frequencies alone cannot reproduce.

\begin{table}[ht]
\centering
\caption{Column-permuted alignment control. The label SA (real) denotes stochastic attention on the original alignment, and SA (perm) denotes stochastic attention on a column-permuted alignment that preserves per-position frequencies but destroys inter-position correlations; MI $r$ is the Pearson correlation between pairwise MI vectors of stored and generated sequences.}
\label{tab:permuted-alignment}
\small
\begin{tabular}{@{}llcccc@{}}
\toprule
Family & Method & KL$_{\mathrm{AA}}$ & Novelty & SeqID & MI $r$ \\
\midrule
\multirow{2}{*}{RRM}
 & SA (real) & $0.060 \pm 0.005$ & $0.513 \pm 0.011$ & $0.538 \pm 0.006$ & $0.53$ \\
 & SA (perm) & $0.050 \pm 0.017$ & $0.629 \pm 0.005$ & $0.460 \pm 0.003$ & $0.48$ \\
\midrule
\multirow{2}{*}{SH3}
 & SA (real) & $0.036 \pm 0.004$ & $0.471 \pm 0.012$ & $0.593 \pm 0.006$ & $0.67$ \\
 & SA (perm) & $0.032 \pm 0.004$ & $0.611 \pm 0.005$ & $0.500 \pm 0.003$ & $0.59$ \\
\midrule
\multirow{2}{*}{WW}
 & SA (real) & $0.008 \pm 0.002$ & $0.653 \pm 0.006$ & $0.562 \pm 0.005$ & $0.73$ \\
 & SA (perm) & $0.006 \pm 0.002$ & $0.688 \pm 0.004$ & $0.540 \pm 0.004$ & $0.73$ \\
\midrule
\multirow{2}{*}{Kunitz}
 & SA (real) & $0.013 \pm 0.002$ & $0.557 \pm 0.009$ & $0.607 \pm 0.004$ & $0.80$ \\
 & SA (perm) & $0.012 \pm 0.002$ & $0.660 \pm 0.007$ & $0.546 \pm 0.003$ & $0.78$ \\
\midrule
\multirow{2}{*}{zf-C2H2}
 & SA (real) & $0.013 \pm 0.027$ & $0.596 \pm 0.010$ & $0.557 \pm 0.007$ & $0.92$ \\
 & SA (perm) & $0.006 \pm 0.005$ & $0.654 \pm 0.006$ & $0.511 \pm 0.004$ & $0.93$ \\
\midrule
\multirow{2}{*}{PDZ}
 & SA (real) & $0.038 \pm 0.011$ & $0.418 \pm 0.009$ & $0.543 \pm 0.005$ & $0.58$ \\
 & SA (perm) & $0.028 \pm 0.005$ & $0.545 \pm 0.007$ & $0.433 \pm 0.003$ & $0.52$ \\
\midrule
\multirow{2}{*}{Pkinase}
 & SA (real) & $0.035 \pm 0.001$ & $0.478 \pm 0.011$ & $0.515 \pm 0.004$ & $0.06$ \\
 & SA (perm) & $0.035 \pm 0.001$ & $0.612 \pm 0.005$ & $0.449 \pm 0.001$ & $0.00$ \\
\midrule
\multirow{2}{*}{Defensin\_beta}
 & SA (real) & $0.022 \pm 0.003$ & $0.402 \pm 0.013$ & $0.655 \pm 0.007$ & $0.85$ \\
 & SA (perm) & $0.017 \pm 0.002$ & $0.542 \pm 0.008$ & $0.552 \pm 0.004$ & $0.82$ \\
\bottomrule
\end{tabular}
\end{table}

\section{PCA Variance Threshold Sensitivity}\label{si:pca-sensitivity}

To assess whether the default 95\% variance threshold for PCA dimensionality reduction is a fragile design choice, and to identify the minimum information content needed for reliable generation, we repeated the full SA generation pipeline at eleven retained-variance thresholds spanning 50\% to 99\%. For each family and threshold, we constructed the memory matrix, identified the critical temperature $\beta^*$ via the entropy inflection criterion, and generated 150 sequences (30 chains $\times$ 5 samples) at $\beta_{\mathrm{gen}} = \max(\lceil 2\beta^* \rceil, 5)$. Table~\ref{tab:pca-sensitivity} reports six representative thresholds for all eight families.

The results revealed two regimes separated by a transition near 70--75\% retained variance. Above this threshold, generation quality was robust: KL divergence remained below 0.09 in every condition, novelty broadly increased with the variance threshold (higher $d$ provides more dimensions for exploration), and nearest sequence identity was essentially constant within each family. The critical temperature $\beta^*$ was largely insensitive to the threshold: in six of eight families, $\beta^*$ was identical across all eleven thresholds, reflecting the discrete resolution of the entropy inflection search grid. Below 70\% retained variance, compositional fidelity began to degrade in the more diverse families. The clearest example was the WW domain ($K{=}420$), where KL divergence was $0.03$ at 80\% but rose to $0.15$ at 50\%, a five-fold increase indicating that discarded principal components encoded position-specific conservation patterns needed for faithful generation. By contrast, families with lower intrinsic diversity (PDZ, Pkinase, Defensin\_beta) were nearly flat across the entire range because their first few principal components already captured the relevant variation. The transition was thus family-dependent, governed by the alignment's spectral structure: families whose eigenvalue spectrum decays slowly required more principal components to preserve compositional fidelity, while compact families tolerated aggressive compression. These findings establish that the 95\% default lies within the robust regime and that practitioners can safely vary the threshold between 75\% and 99\% without substantially affecting output quality.

\begin{table}[p]
\centering
\caption{PCA variance threshold sensitivity study. For each family and retained-variance threshold, the table reports the PCA dimension $d$, empirical critical temperature $\beta^*$, generation temperature $\beta_{\mathrm{gen}}$, amino acid composition KL divergence, PCA-space cosine novelty, and nearest sequence identity. The horizontal rule within each family separates the degraded regime ($\leq 70\%$) from the robust regime ($\geq 80\%$).}
\label{tab:pca-sensitivity}
\scriptsize
\setlength{\tabcolsep}{4pt}
\renewcommand{\arraystretch}{0.85}
\begin{tabular}{@{}llrrrrcc@{}}
\toprule
Family & Threshold & $d$ & $\beta^*$ & $\beta_{\mathrm{gen}}$ & KL$_{\mathrm{AA}}$ & Novelty & SeqID \\
\midrule
\multirow{6}{*}{RRM}
 & 50\% &  21 & 3.85 &  8 & 0.186 & 0.392 & 0.539 \\
 & 60\% &  27 & 3.85 &  8 & 0.111 & 0.426 & 0.546 \\
 & 70\% &  34 & 3.85 &  8 & 0.135 & 0.463 & 0.539 \\
\cmidrule{2-8}
 & 80\% &  43 & 3.85 &  8 & 0.085 & 0.497 & 0.535 \\
 & 90\% &  53 & 3.85 &  8 & 0.067 & 0.508 & 0.540 \\
 & 95\% &  59 & 3.85 &  8 & 0.060 & 0.513 & 0.538 \\
\midrule
\multirow{6}{*}{SH3}
 & 50\% &  15 & 3.23 &  6 & 0.087 & 0.312 & 0.601 \\
 & 60\% &  19 & 3.23 &  6 & 0.072 & 0.372 & 0.593 \\
 & 70\% &  25 & 3.23 &  6 & 0.039 & 0.402 & 0.592 \\
\cmidrule{2-8}
 & 80\% &  32 & 3.23 &  6 & 0.038 & 0.447 & 0.587 \\
 & 90\% &  41 & 3.85 &  8 & 0.032 & 0.444 & 0.596 \\
 & 95\% &  46 & 3.85 &  8 & 0.036 & 0.471 & 0.593 \\
\midrule
\multirow{6}{*}{WW}
 & 50\% &  33 & 5.45 & 11 & 0.152 & 0.410 & 0.646 \\
 & 60\% &  47 & 5.45 & 11 & 0.133 & 0.476 & 0.631 \\
 & 70\% &  67 & 5.45 & 11 & 0.125 & 0.530 & 0.624 \\
\cmidrule{2-8}
 & 80\% &  95 & 5.45 & 11 & 0.028 & 0.588 & 0.600 \\
 & 90\% & 142 & 5.45 & 11 & 0.020 & 0.636 & 0.576 \\
 & 95\% & 186 & 5.45 & 11 & 0.008 & 0.653 & 0.562 \\
\midrule
\multirow{6}{*}{Kunitz}
 & 50\% &  23 & 3.85 &  8 & 0.053 & 0.387 & 0.638 \\
 & 60\% &  31 & 3.85 &  8 & 0.047 & 0.433 & 0.636 \\
 & 70\% &  40 & 3.85 &  8 & 0.031 & 0.469 & 0.628 \\
\cmidrule{2-8}
 & 80\% &  53 & 3.85 &  8 & 0.022 & 0.500 & 0.619 \\
 & 90\% &  69 & 3.85 &  8 & 0.023 & 0.535 & 0.616 \\
 & 95\% &  80 & 3.85 &  8 & 0.013 & 0.557 & 0.607 \\
\midrule
\multirow{6}{*}{zf-C2H2}
 & 50\% &  28 & 4.58 &  9 & 0.112 & 0.397 & 0.610 \\
 & 60\% &  38 & 4.58 &  9 & 0.037 & 0.458 & 0.599 \\
 & 70\% &  49 & 4.58 &  9 & 0.106 & 0.483 & 0.595 \\
\cmidrule{2-8}
 & 80\% &  65 & 4.58 &  9 & 0.017 & 0.528 & 0.583 \\
 & 90\% &  88 & 4.58 &  9 & 0.010 & 0.565 & 0.572 \\
 & 95\% & 106 & 4.58 &  9 & 0.013 & 0.596 & 0.557 \\
\midrule
\multirow{6}{*}{PDZ}
 & 50\% &  12 & 2.70 &  5 & 0.038 & 0.269 & 0.558 \\
 & 60\% &  16 & 3.23 &  6 & 0.035 & 0.305 & 0.564 \\
 & 70\% &  20 & 3.23 &  6 & 0.042 & 0.337 & 0.553 \\
\cmidrule{2-8}
 & 80\% &  25 & 3.23 &  6 & 0.035 & 0.359 & 0.549 \\
 & 90\% &  32 & 3.23 &  6 & 0.046 & 0.400 & 0.544 \\
 & 95\% &  37 & 3.23 &  6 & 0.038 & 0.418 & 0.543 \\
\midrule
\multirow{6}{*}{Pkinase}
 & 50\% &  15 & 3.23 &  6 & 0.041 & 0.356 & 0.501 \\
 & 60\% &  18 & 3.23 &  6 & 0.041 & 0.390 & 0.500 \\
 & 70\% &  22 & 3.23 &  6 & 0.038 & 0.426 & 0.506 \\
\cmidrule{2-8}
 & 80\% &  27 & 3.23 &  6 & 0.035 & 0.463 & 0.506 \\
 & 90\% &  31 & 3.23 &  6 & 0.035 & 0.480 & 0.509 \\
 & 95\% &  34 & 3.23 &  6 & 0.035 & 0.478 & 0.515 \\
\midrule
\multirow{6}{*}{Defensin\_beta}
 & 50\% &  11 & 2.70 &  5 & 0.036 & 0.223 & 0.679 \\
 & 60\% &  15 & 3.23 &  6 & 0.035 & 0.262 & 0.678 \\
 & 70\% &  19 & 3.23 &  6 & 0.030 & 0.307 & 0.668 \\
\cmidrule{2-8}
 & 80\% &  25 & 3.23 &  6 & 0.029 & 0.312 & 0.680 \\
 & 90\% &  32 & 3.23 &  6 & 0.020 & 0.381 & 0.657 \\
 & 95\% &  37 & 3.23 &  6 & 0.022 & 0.402 & 0.655 \\
\bottomrule
\end{tabular}
\end{table}

\section{Sequence Encoding, PCA Projection, and Decoding}\label{si:pca}

Stochastic attention operates on continuous vectors on the unit sphere $\mathbb{S}^{d-1}$, while protein sequences are discrete objects over a 20-letter amino acid alphabet. Bridging the two requires three transformations (one-hot encoding, PCA dimensionality reduction, and unit-norm projection), and the reverse path (inverse PCA followed by argmax decoding) recovers discrete sequences from continuous samples. Given a cleaned alignment of $K$ sequences, each of length $L$ amino acids over the standard 20-letter alphabet $\mathcal{A} = \{\text{A}, \text{R}, \text{N}, \ldots, \text{V}\}$, we encode each sequence as a binary vector in $\mathbb{R}^{20L}$ by concatenating per-position indicator vectors:
\begin{equation}
    \mathbf{x}_k = \bigl[\mathbf{e}_{a_1^{(k)}},\; \mathbf{e}_{a_2^{(k)}},\; \ldots,\; \mathbf{e}_{a_L^{(k)}}\bigr]^\top \in \{0,1\}^{20L},
\end{equation}
where $\mathbf{e}_a \in \{0,1\}^{20}$ is the standard basis vector for amino acid $a$. The full-dimensional representation is $d_{\mathrm{full}} = 20L$, which ranges from $460$ (zf-C2H2, $L{=}23$) to $5{,}240$ (Pkinase, $L{=}262$) across the eight families studied. This encoding is lossless and treats each position-amino acid pair as an independent coordinate, making no assumptions about residue similarity or physicochemical properties. Gap and non-standard-residue positions are encoded as the all-zero $20$-vector, so they contribute nothing to the inner products that drive retrieval. On decoding, a position is assigned a gap only if its reconstructed $20$-channel block is numerically zero (maximum entry below $10^{-10}$). Because inverse PCA produces dense reconstructions, every decoded position resolved to a standard amino acid; decoded sequences therefore contained no gaps, consistent with the $100\%$ valid-amino-acid rate. Sequence identity is computed over non-gap positions only.

The full one-hot space is high-dimensional relative to the number of stored patterns; the ratio $d_{\mathrm{full}}/K$ ranges from $1.5$ (WW) to $142$ (Pkinase). This creates two problems for SA. First, in the modern Hopfield energy, the similarity scores $e_k = \mathbf{m}_k^\top\boldsymbol{\xi}$ have variance $1/d$ for a random query on $\mathbb{S}^{d-1}$. When $d$ is large, scores concentrate tightly around zero and high $\beta$ is required to differentiate among stored patterns; this flattens the energy landscape and renders Langevin sampling inefficient. Second, the one-hot encoding introduces redundancy: at each position, only 1 of 20 coordinates is nonzero. Principal component analysis removes this redundancy by projecting onto the directions of actual variation. We compute the economy singular value decomposition of the centered data matrix $\tilde{\mathbf{X}} = \mathbf{U}\boldsymbol{\Sigma}\mathbf{V}^\top$ and select the smallest $p$ such that the retained variance fraction $\rho(p) = \sum_{j=1}^p \sigma_j^2 / \sum_j \sigma_j^2 \geq 0.95$. Among all rank-$p$ linear projections, PCA minimizes the mean squared reconstruction error, making it the optimal linear dimensionality reduction for preserving the structure of the data. Across the eight families, the retained PCA dimension $d$ ranged from $34$ (Pkinase) to $186$ (WW).

After PCA, we normalize each projected vector to unit $L_2$ norm: $\mathbf{m}_k = \mathbf{z}_k / \|\mathbf{z}_k\|_2 \in \mathbb{S}^{d-1}$, which ensures that the similarity scores lie in $[-1, 1]$ and that the concentration-of-measure results hold exactly for random unit-sphere probes. To decode a Langevin sample $\boldsymbol{\xi} \in \mathbb{R}^d$ back to an amino acid sequence, we apply inverse PCA ($\hat{\mathbf{x}} = \bar{\mathbf{x}} + \mathbf{W}_d\boldsymbol{\xi}$) followed by per-position argmax decoding: for each position $\ell$, the amino acid with the largest coordinate in the 20-dimensional sub-vector is selected. This deterministic decoding produced $100\%$ valid amino acids across all families and conditions. Several properties make PCA well suited for the modern Hopfield energy: linearity (both projection and inverse are matrix multiplications, keeping the score function in closed form), variance-optimality (it captures maximum family-level variation in the fewest dimensions), and parameter-free determinism (the projection is uniquely defined by the SVD and variance threshold, with no hyperparameters to tune). Algorithm~\ref{alg:sa-pipeline} summarizes the complete pipeline.

\begin{algorithm}[ht]
\caption{Stochastic Attention Protein Sequence Generation}\label{alg:sa-pipeline}
\begin{algorithmic}[1]
\REQUIRE Seed alignment $\mathbf{S} \in \mathcal{A}^{K \times L}$, variance threshold $\rho_{\min} = 0.95$, step size $\alpha = 0.01$, chains $N_c = 30$, iterations $T = 5000$, burn-in $T_b = 2000$, thinning $\Delta t = 100$, samples/chain $s = 5$
\ENSURE Generated sequences $\{\mathbf{g}_1, \ldots, \mathbf{g}_S\} \subset \mathcal{A}^L$, \; $S = N_c \cdot s$
\STATE \textbf{Encoding:}
\STATE One-hot encode: $\mathbf{X} \leftarrow [\mathrm{onehot}(\mathbf{S}_{1,:}), \ldots, \mathrm{onehot}(\mathbf{S}_{K,:})] \in \{0,1\}^{20L \times K}$
\STATE Center: $\bar{\mathbf{x}} \leftarrow \frac{1}{K}\sum_k \mathbf{X}_{:,k}$; \; $\tilde{\mathbf{X}} \leftarrow \mathbf{X} - \bar{\mathbf{x}}\mathbf{1}_K^\top$
\STATE SVD: $\tilde{\mathbf{X}} = \mathbf{U}\boldsymbol{\Sigma}\mathbf{V}^\top$
\STATE Select $d$: smallest $p$ such that $\sum_{j=1}^{p}\sigma_j^2 / \sum_j \sigma_j^2 \geq \rho_{\min}$; set $\mathbf{W}_d \leftarrow \mathbf{U}_{:,1:d}$
\STATE Project and normalize: $\mathbf{m}_k \leftarrow \mathbf{W}_d^\top(\mathbf{X}_{:,k} - \bar{\mathbf{x}}) / \|\mathbf{W}_d^\top(\mathbf{X}_{:,k} - \bar{\mathbf{x}})\|_2$ \; for $k = 1,\ldots,K$
\STATE Assemble memory: $\hat{\mathbf{X}} \leftarrow [\mathbf{m}_1, \ldots, \mathbf{m}_K] \in \mathbb{R}^{d \times K}$
\STATE \textbf{Temperature selection:}
\STATE Set $K_p \leftarrow \min(K,20)$ and evaluate 50 log-spaced $\beta$ values in $[0.1,500]$
\STATE For $i=1,\ldots,K_p$, compute $\mathbf{p}_i(\beta)=\operatorname{softmax}(\beta\,\hat{\mathbf{X}}^\top\mathbf{m}_i)$ and $H_i(\beta)=-\sum_{k=1}^{K} p_{ik}(\beta)\log p_{ik}(\beta)$
\STATE Set $\beta^*$ to the entropy inflection point of $\bar H(\beta)=K_p^{-1}\sum_i H_i(\beta)$ on this grid
\STATE Set $\beta_{\mathrm{gen}} \leftarrow \max\!\bigl(\lceil 2\beta^* \rceil,\; 5\bigr)$
\STATE \textbf{Multi-chain Langevin sampling:}
\STATE $\mathcal{G} \leftarrow \emptyset$
\FOR{$c = 1, \ldots, N_c$}
    \STATE Draw $k_c \sim \mathrm{Uniform}\{1,\ldots,K\}$
    \STATE $\boldsymbol{\xi}_0 \leftarrow \mathbf{m}_{k_c} + 0.01\,\boldsymbol{\eta}$, \; $\boldsymbol{\eta} \sim \mathcal{N}(\mathbf{0}, \mathbf{I}_d)$
    \FOR{$t = 0, \ldots, T-1$}
        \STATE $\mathbf{a}_t \leftarrow \operatorname{softmax}(\beta_{\mathrm{gen}}\,\hat{\mathbf{X}}^\top\boldsymbol{\xi}_t)$
        \STATE $\boldsymbol{\epsilon}_t \sim \mathcal{N}(\mathbf{0}, \mathbf{I}_d)$
        \STATE $\boldsymbol{\xi}_{t+1} \leftarrow (1-\alpha)\,\boldsymbol{\xi}_t + \alpha\,\hat{\mathbf{X}}\,\mathbf{a}_t + \sqrt{2\alpha/\beta_{\mathrm{gen}}}\;\boldsymbol{\epsilon}_t$
    \ENDFOR
    \STATE Collect $s$ evenly spaced samples from $\{\boldsymbol{\xi}_{T_b+1}, \boldsymbol{\xi}_{T_b+\Delta t+1}, \ldots, \boldsymbol{\xi}_T\}$ into $\mathcal{G}$
\ENDFOR
\STATE \textbf{Decoding:}
\FOR{each $\boldsymbol{\xi} \in \mathcal{G}$}
    \STATE Inverse PCA: $\hat{\mathbf{x}} \leftarrow \bar{\mathbf{x}} + \mathbf{W}_d\,\boldsymbol{\xi} \in \mathbb{R}^{20L}$
    \STATE Argmax decode: $g_\ell \leftarrow \arg\max_{a \in \mathcal{A}}\; \hat{x}_{20(\ell-1)+a}$ \; for $\ell = 1,\ldots,L$
\ENDFOR
\STATE \textbf{return} $\{\mathbf{g}_1, \ldots, \mathbf{g}_S\}$
\end{algorithmic}
\end{algorithm}

\section{Analysis of the Critical Temperature}\label{si:beta-star-theory}

\begin{figure}[ht]
\centering
\includegraphics[width=0.7\textwidth]{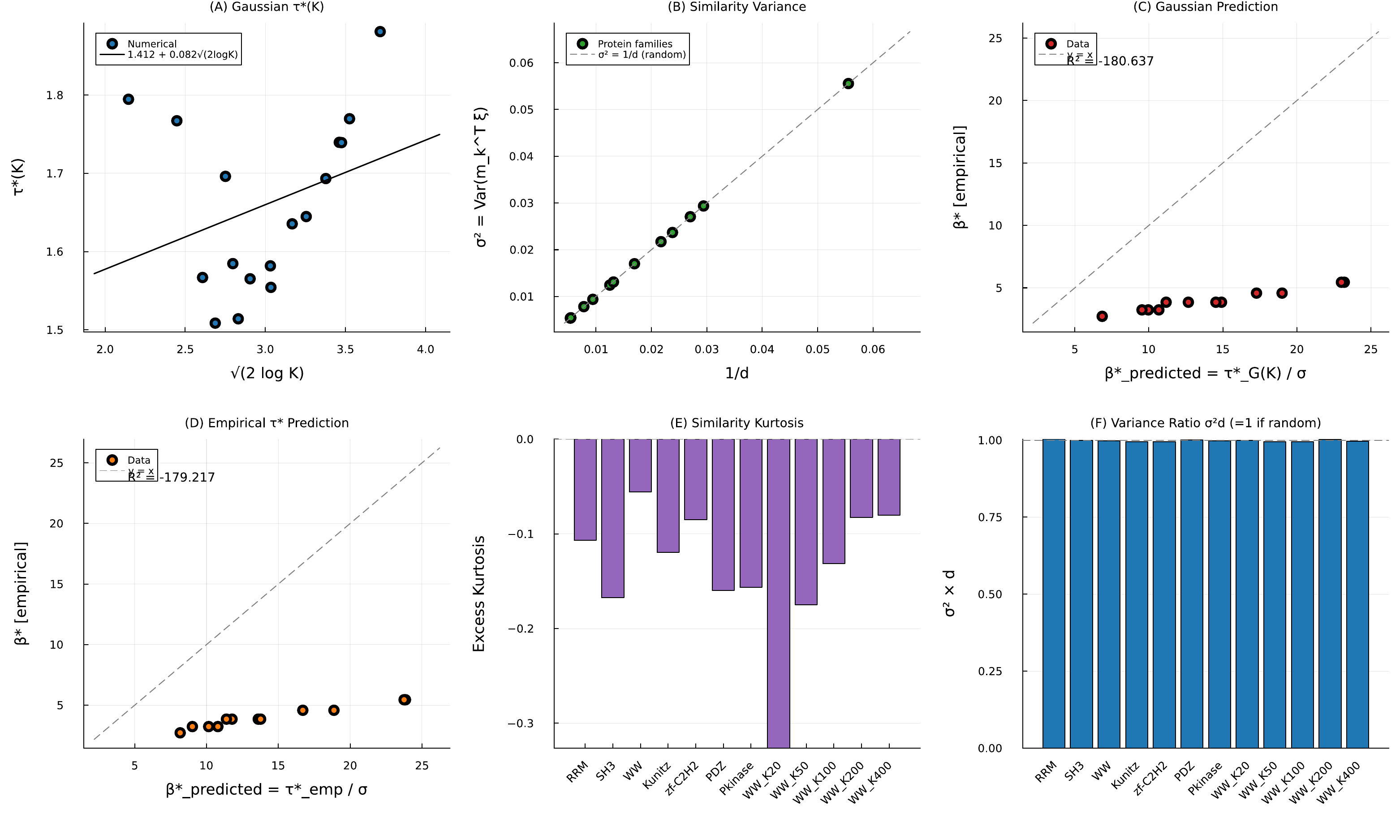}
\caption{\textbf{Analytical investigation of $\beta^*$.}
(A)~Dimensionless softmax entropy inflection $\tau^*(K)$ for Gaussian scores.
(B)~Similarity variance $\sigma^2$ vs.\ $1/d$ for random probes; all points lie on the identity line.
(C)~Gaussian prediction $\beta^*_{\mathrm{pred}} = \tau^*_G(K)/\sigma$ vs.\ empirical $\beta^*$; correct trend ($r = 0.98$) but overshoots by ${\sim}4\times$.
(D)~Same as (C) using the empirical $\tau^*$; overshoot persists.
(E)~Excess kurtosis of the similarity distribution.
(F)~The ratio $\sigma^2 d$ across all datasets, showing $\sigma^2 d \approx 1$.}
\label{fig:analytical-beta-star}
\end{figure}

We developed an analytical framework to understand the empirical relationship $\beta^* \approx 1.52 + 0.28\sqrt{d}$ reported in Materials and Methods, starting from exact results on similarity variance, building a Gaussian mean-field prediction, and then diagnosing why the mean-field coefficient overshoots and how stored-pattern self-similarity resolves the discrepancy. After one-hot encoding and PCA projection to $d$ dimensions, the stored patterns $\mathbf{m}_k$ are normalized to lie on $\mathbb{S}^{d-1}$. For a random query $\boldsymbol{\xi}$ drawn uniformly from $\mathbb{S}^{d-1}$, each similarity $e_k = \mathbf{m}_k^\top\boldsymbol{\xi}$ has mean zero and variance $1/d$, regardless of the structure of $\mathbf{m}_k$, a consequence of the rotational invariance of the uniform distribution on the sphere. We verified this numerically for all eight families using 1{,}000 random probes per family: the ratio $\sigma^2 \cdot d$ is $1.000 \pm 0.003$ across all families and WW subsamples (Fig.~\ref{fig:analytical-beta-star}B,F), and the excess kurtosis is mildly negative ($-0.06$ to $-0.33$), consistent with sub-Gaussian behavior on $\mathbb{S}^{d-1}$. The characteristic similarity scale is therefore $\sigma = 1/\sqrt{d}$, and since $\beta$ enters the softmax as $\beta e_k$, the entropy inflection must occur when $\beta\sigma = \mathcal{O}(1)$, giving $\beta^* \sim \sqrt{d}$.

These observations motivated a Gaussian mean-field model: if the $K$ similarity scores were i.i.d.\ draws from $\mathcal{N}(0, 1/d)$, the softmax entropy $H(\tau) = -\sum_k p_k \log p_k$ with $p_k \propto \exp(\tau z_k)$ undergoes an entropy inflection between the uniform-attention regime ($H = \log K$, $\tau = 0$) and the pattern-selective regime ($H \to 0$, $\tau \to \infty$). We defined the dimensionless entropy inflection $\tau^*$ using the same entropy-inflection criterion as the empirical $\beta^*$ search and computed it numerically by averaging over 500 Gaussian realizations for each $K \in \{10, 20, \ldots, 1000\}$. We found that $\tau^*$ is nearly constant at $\approx 1.6$ for $K \geq 30$, with only weak logarithmic growth (Fig.~\ref{fig:analytical-beta-star}A). Translating back to physical units, the Gaussian prediction $\beta^* = \tau^*\sqrt{d} \approx 1.6\sqrt{d}$ has the correct $\sqrt{d}$ scaling and achieves Pearson $r = 0.98$ with the empirical values, but systematically overpredicts $\beta^*$ by a factor of $\approx 4$.

\begin{figure}[ht]
\centering
\includegraphics[width=0.7\textwidth]{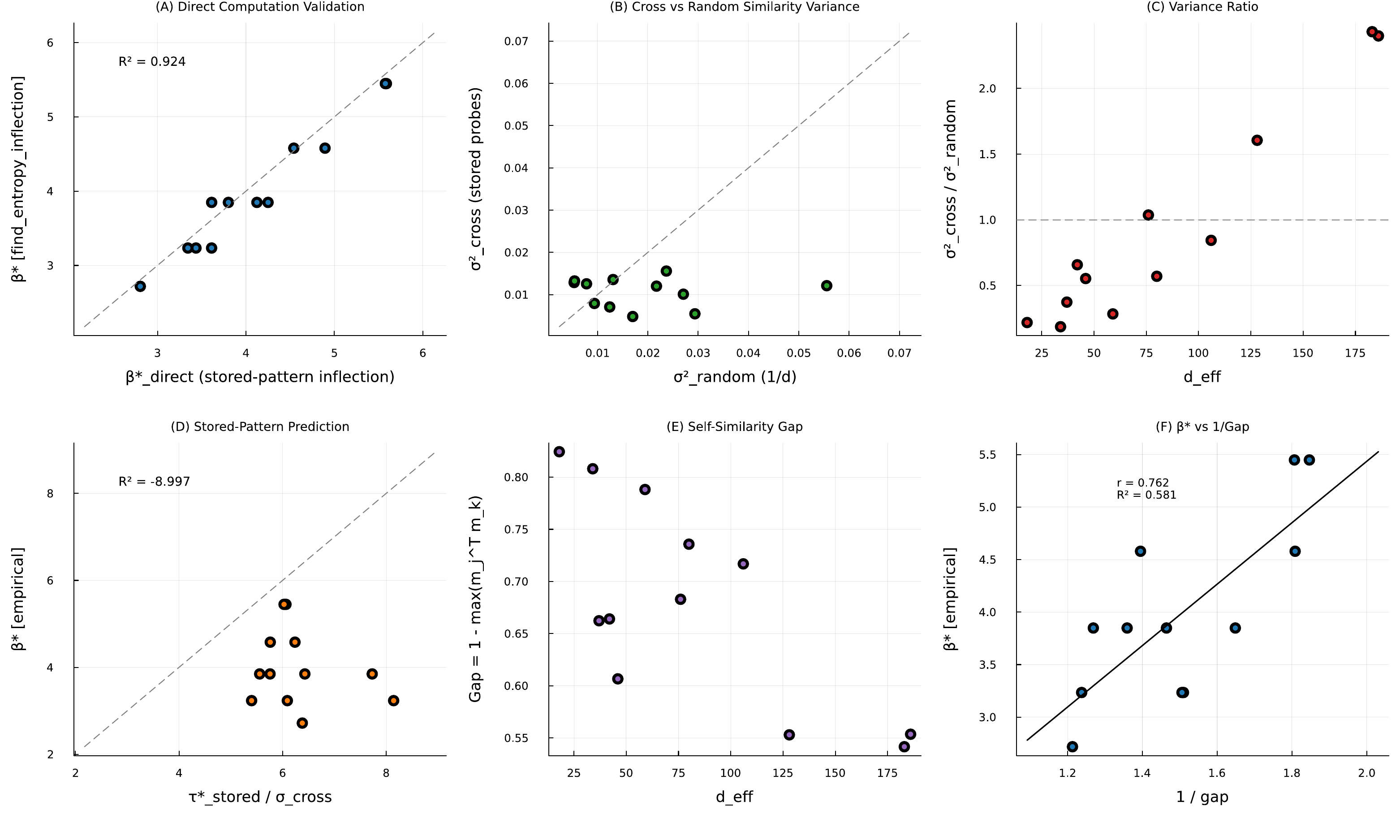}
\caption{\textbf{Stored-pattern probe analysis.}
(A)~$\beta^*$ from stored-pattern entropy inflection vs.\ the value from the entropy-inflection search.
(B)~Cross-similarity variance (stored probes) vs.\ random-probe variance.
(C)~Variance ratio vs.\ $d$.
(D)~The decomposition $\tau^*_{\mathrm{stored}}/\sigma_{\mathrm{cross}}$ fails as a predictor ($R^2 = -9.0$).
(E)~Self-similarity gap $\Delta$ vs.\ $d$.
(F)~$\beta^*$ vs.\ $1/\Delta$ ($r = 0.76$).}
\label{fig:stored-pattern-analysis}
\end{figure}

This overprediction arises because the empirical $\beta^*$ is computed from stored-pattern probes $\mathbf{m}_i$, not from random queries. When $\boldsymbol{\xi} = \mathbf{m}_i$, the self-similarity score $\mathbf{m}_i^\top\mathbf{m}_i = 1$ is an outlier relative to the cross-similarities $\mathbf{m}_j^\top\mathbf{m}_i$ for $j \neq i$, which have magnitudes $\mathcal{O}(1/\sqrt{d})$. The resulting gap $\Delta = 1 - \max_{j\neq i}\mathbf{m}_j^\top\mathbf{m}_i$ ranges from $0.55$ to $0.82$ across families (Fig.~\ref{fig:stored-pattern-analysis}E), meaning the softmax concentrates on the self-pattern at lower $\beta$ than would be needed for a random query. Substituting the cross-similarity standard deviation $\sigma_{\mathrm{cross}}$ for $1/\sqrt{d}$ fails decisively ($R^2 = -9.0$; Fig.~\ref{fig:stored-pattern-analysis}D), because the cross-similarity distribution is heavy-tailed (excess kurtosis $6$--$25$) and the self-similarity outlier violates the Gaussian assumption. The fitted model $\beta^* = 1.52 + 0.28\sqrt{d}$ nonetheless remains robust: the $\sqrt{d}$ scaling is set by concentration of measure, and the reduced coefficient $0.28$ (vs.\ $\tau^* \approx 1.6$) encapsulates the combined effect of self-similarity anchoring and the protein family correlation structure.

To quantify uncertainty in the fitted regression, we performed a nonparametric bootstrap analysis: resampling the 33 observations (8 families plus 25 WW domain scaling replicates) with replacement 10{,}000 times yielded bootstrap SE of $0.07$ for the intercept and $0.007$ for the slope, with 95\% confidence intervals of $[1.39, 1.68]$ and $[0.266, 0.294]$, respectively. The bootstrap $R^2$ distribution has a median of $0.969$ (95\% CI $[0.940, 0.987]$). Because 25 of the 33 data points come from WW domain subsamples, we also refit the model using only the eight Pfam families: the resulting fit, $\beta^* = 1.64 + 0.278\sqrt{d}$ ($R^2{=}0.95$), is nearly identical, indicating that the result is not driven by overrepresentation of a single family. Leave-one-family-out cross-validation achieved $R^2{=}0.96$ (root mean square error, RMSE${=}0.18$) on the full dataset and $R^2{=}0.93$ (RMSE${=}0.19$) on the 8-family subset, with no individual family acting as an outlier (Table~\ref{tab:beta-loocv}). Two cautions attend the pooled fit. First, $25$ of the $33$ observations are nested WW subsamples and are therefore not independent; the eight-family-only fit above is the more conservative estimate, and the leave-one-family-out results confirm out-of-sample generalization. Second, the empirical $\beta^*$ is identified on a fixed grid of $50$ logarithmically spaced values over $[0.1, 500]$ (multiplicative spacing $\approx 1.19$), and the eight family values land on consecutive grid points ($\beta^* \in \{3.23, 3.85, 4.58, 5.45\}$). As a result, $\beta^*$ is resolved only to within ${\approx}19\%$, and the reported $R^2$ is computed against these grid-resolved targets. A finer grid would refine the point estimates but is not expected to alter the $\sqrt{d}$ scaling, which follows from concentration of measure rather than from the grid.

\begin{figure}[ht]
\centering
\includegraphics[width=0.7\textwidth]{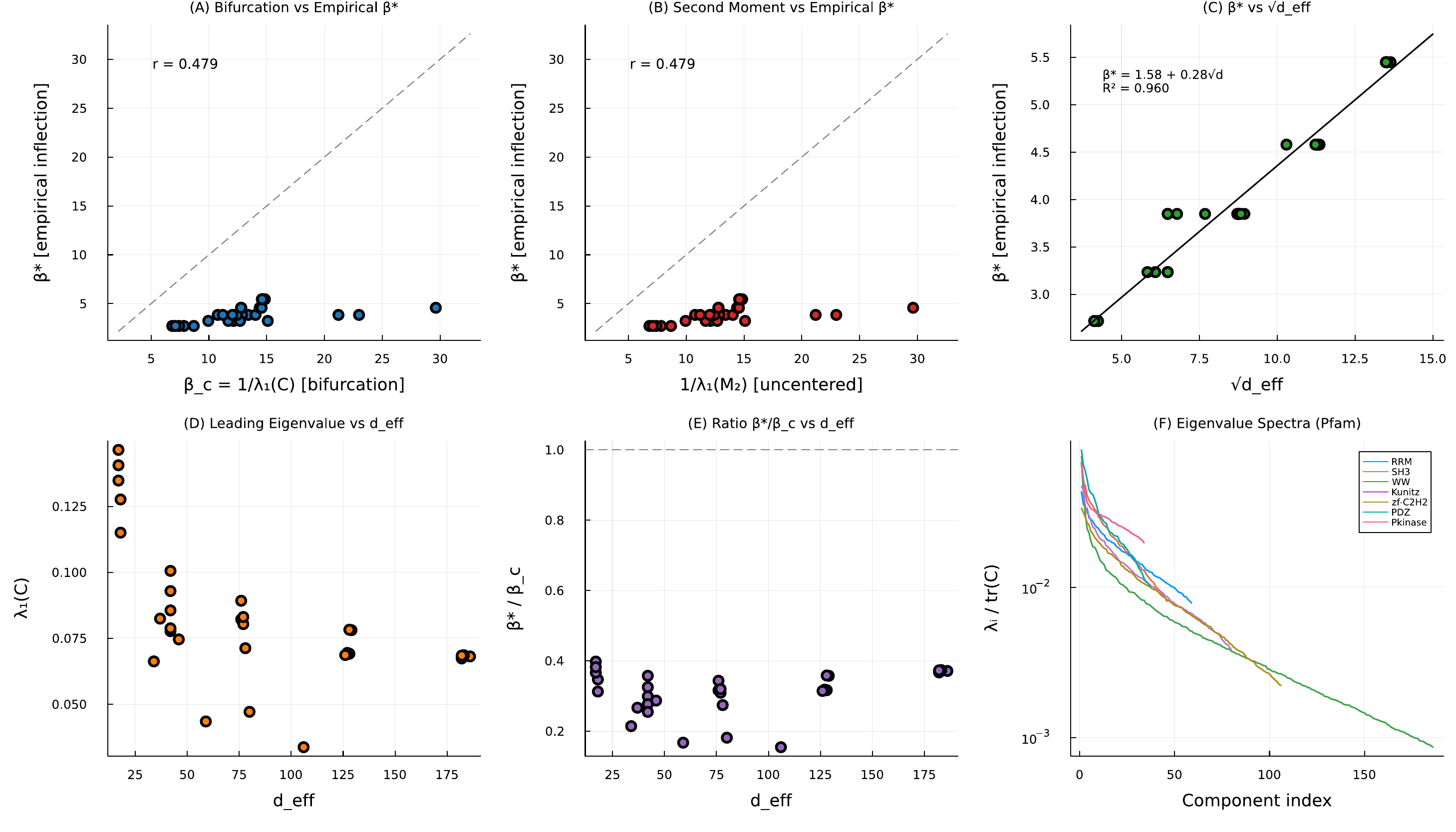}
\caption{\textbf{Bifurcation-based prediction of $\beta^*$.}
(A)~$\beta_c = 1/\lambda_1(\mathbf{C})$ vs.\ empirical $\beta^*$ ($r{=}0.479$); bifurcation overshoots by $3$--$5\times$.
(B)~Second-moment predictor performance is identical.
(C)~$\beta^*$ vs.\ $\sqrt{d}$ with fitted model ($R^2{=}0.960$).
(D)~Leading eigenvalue vs.\ $d$; no consistent relationship.
(E)~Ratio $\beta^*/\beta_c$ showing that the bifurcation estimate systematically overestimates.
(F)~Normalized eigenvalue spectra for all eight families.}
\label{fig:bifurcation-analysis}
\end{figure}

\begin{table}[ht]
\centering
\caption{Leave-one-family-out cross-validation of $\beta^*$ prediction. Scheme 1: full 33-point dataset. Scheme 2: 8-family-only LOO. Both schemes support out-of-sample generalization.}
\label{tab:beta-loocv}
\small
\begin{tabular}{@{}lccc@{}}
\toprule
Scheme & $n$ & LOOCV $R^2$ & LOOCV RMSE \\
\midrule
Full 33-pt LOFO & 33 & $0.959$ & $0.184$ \\
8-family LOO    &  8 & $0.928$ & $0.195$ \\
\bottomrule
\end{tabular}
\end{table}

\section{Regularity and Convergence Properties}\label{si:convergence}

The modern Hopfield energy has Lipschitz-continuous gradient with constant at most $L = 1 + \beta\sigma_{\max}^2/2$, where $\sigma_{\max} = \|\hat{\mathbf{X}}\|_{\mathrm{op}}$ for the normalized memory matrix, and satisfies the dissipativity condition $\langle \nabla E(\boldsymbol{\xi}), \boldsymbol{\xi} \rangle \geq \|\boldsymbol{\xi}\|_2^2/2 - M^2/2$ with $M = \max_k \|\mathbf{m}_k\|_2$. These regularity properties determine the convergence regime of the Langevin sampler. When $\beta\sigma_{\max}^2 < 2$, this sufficient bound implies convexity of the energy and ULA iterates converge to $p_\beta$ in Wasserstein-2 distance at a geometric rate~\cite{dalalyanTheoreticalGuaranteesApproximate2017,durmusNonasymptoticAnalysisUnadjusted2017}. In the protein setting, operating temperatures satisfy $\beta\sigma_{\max}^2 \gg 2$, placing the sampler outside this globally convex regime, where the energy landscape can have multiple metastable basins corresponding to stored patterns or clusters thereof. Despite this non-convexity, the multi-chain protocol with long burn-in and thinning yields well-mixed chains, and the initialization sensitivity analysis indicates that the sampled distribution is insensitive to starting conditions.

\section{Pfam Family Size Census}\label{si:pfam-census}

To verify the claims about Pfam family sizes made in the main text, we performed a census of all Pfam-A families using release 36.0, the final standalone Pfam release before its integration into InterPro~\cite{paysanlafosse2023interpro}. We downloaded both the seed alignment file (\texttt{Pfam-A.seed.gz}) and the full alignment file (\texttt{Pfam-A.full.gz}) from the European Bioinformatics Institute FTP archive and parsed each Stockholm-format entry to count the number of unique sequences per family. Seed alignments are small, manually curated sets of representative sequences that Pfam curators selected to capture the diversity of each family while maintaining alignment quality. These seed alignments serve as the input to HMMER, which builds a profile HMM and searches UniProtKB to produce the full alignment. The full alignment therefore contains every protein sequence that HMMER can confidently assign to the family, making it larger but also noisier: it includes automatically identified homologs, sequence fragments, and partial matches that were not subject to manual curation. Because our method and most alignment-based generative approaches operate on curated seed alignments rather than raw database hits, the seed alignment sizes are the operationally relevant quantities.

\begin{figure}[htbp]
\centering
\includegraphics[width=0.85\textwidth]{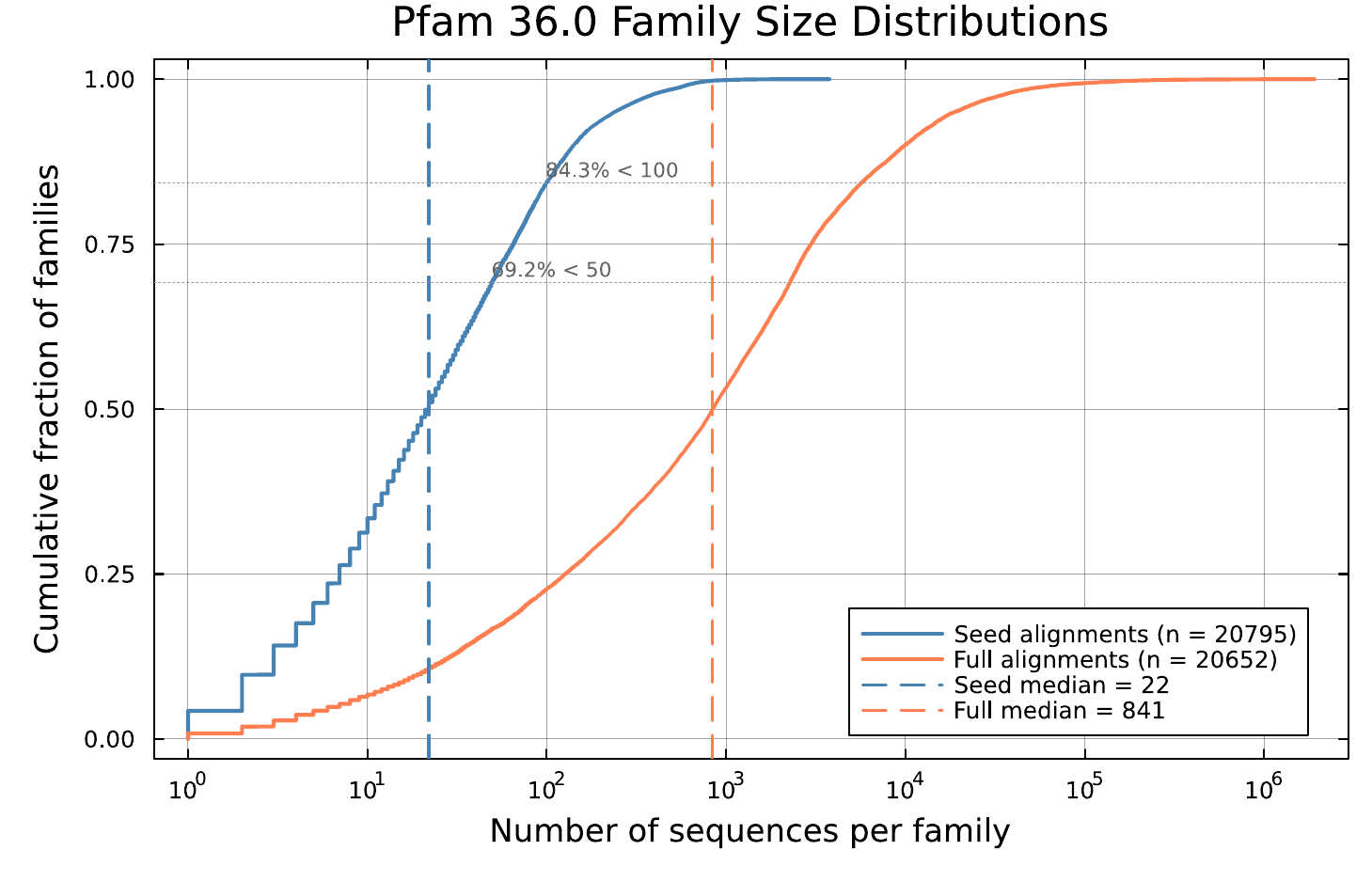}
\caption{Empirical cumulative distribution functions of family size for Pfam~36.0 seed alignments (blue) and full alignments (red). Dashed vertical lines mark medians. The majority of seed alignments contain fewer than 50 sequences.}
\label{fig:pfam-census}
\end{figure}

Table~\ref{tab:pfam-census} summarizes the resulting distributions across all 20{,}795 Pfam-A families. The median seed alignment contains only 22 sequences (interquartile range: 7--65), and 69.2\% of families have fewer than 50 seed sequences. Nearly half (47.5\%) have fewer than 20. The full alignments are larger, with a median of 841 sequences (interquartile range: 127--2{,}973), reflecting the breadth of automated homology detection across UniProtKB. Nevertheless, even at the full-alignment level, 22.7\% of families have fewer than 100 sequences, and the distribution remains heavily right-skewed: a small number of ubiquitous superfamilies (protein kinases, immunoglobulins, zinc fingers) contain hundreds of thousands of members, while the long tail of specialized families remains data-poor. Figure~\ref{fig:pfam-census} shows the empirical cumulative distribution functions for both seed and full alignment sizes on a logarithmic scale. These distributions confirm that the majority of protein families fall below the data requirements of current deep generative architectures, whether measured by curated seed alignments or by the broader full-alignment census, motivating training-free approaches such as stochastic attention.

\begin{table}[htbp]
\centering
\caption{Summary statistics for Pfam 36.0 family sizes. Values report the number of sequences per family across all Pfam-A entries.}
\label{tab:pfam-census}
\small
\begin{tabular}{lrr}
\toprule
Statistic & Seed alignments & Full alignments \\
\midrule
Total families      & 20{,}795           & 20{,}652           \\
Median              & 22                 & 841                \\
Mean                & 59.5               & 5{,}075.9          \\
25th percentile     & 7                  & 127                \\
75th percentile     & 65                 & 2{,}973            \\
\midrule
Families $< 20$     & 9{,}888 (47.5\%)   & 2{,}028 (9.8\%)    \\
Families $< 50$     & 14{,}386 (69.2\%)  & 3{,}438 (16.6\%)   \\
Families $< 100$    & 17{,}523 (84.3\%)  & 4{,}684 (22.7\%)   \\
\bottomrule
\end{tabular}
\end{table}

\section{Software Versions, Random Seeds, and Reproducibility}\label{si:reproducibility}

All experiments were run on an Apple M2 workstation (16~GB RAM). The stochastic-attention sampler, baseline evaluation, scaling study, sampling diagnostics, and all metric computations were implemented in Julia~1.12.5; the principal packages and their versions were Flux~0.16.9, NNlib~0.9.33, MultivariateStats~0.10.4 (principal component analysis), Distributions~0.25.123, StatsBase~0.34.10, Clustering~0.15.8, DataFrames~1.8.1, CSV~0.10.16, and Plots~1.41.6 with StatsPlots~0.15.8. The complete Julia environment is pinned by the \texttt{Project.toml} and \texttt{Manifest.toml} files in the repository. The deep-learning baselines and validators were run in Python~3.12 using the following published model checkpoints: ESM2-650M (\texttt{esm2\_t33\_650M\_UR50D}) for pseudo-perplexity scoring; the MSA Transformer (\texttt{esm\_msa1b\_t12\_100M\_UR50S}); EvoDiff (\texttt{MSA\_OA\_DM\_RANDSUB}); and ESMFold together with AlphaFold2 for structure prediction, the latter run through ColabFold with model type \texttt{alphafold2\_ptm}, three recycles, one model, and seed~0. Profile HMMs were built and sampled with HMMER~3.4, structural superpositions used TM-align~\cite{zhangTMalign2005}, and the Potts baseline used a custom pseudo-likelihood maximization (plmDCA) implementation included in the repository. The exact versions of the Python dependencies (PyTorch, \texttt{fair-esm}, EvoDiff, ColabFold, NumPy, SciPy, Biopython) and of the TM-align binary are recorded in the environment specification archived with the versioned software record described in the main-text Data, Materials, and Software Availability statement.

Random seeds were fixed for every stochastic step so that each experiment is reproducible. Table~\ref{tab:seeds} lists the seeds used; per-chain seeds use a fixed base offset plus the chain index $c$.

\begin{table}[ht]
\centering
\caption{Fixed random seeds by experiment.}
\label{tab:seeds}
\small
\begin{tabular}{@{}lll@{}}
\toprule
Experiment & Implementation & Seed(s) \\
\midrule
Main sequence generation             & Julia  & $42$ (setup); $12345{+}c$ per chain \\
HMM validation                       & Julia  & $42$; $12345{+}c$ per chain \\
Sampling diagnostics (ULA/MALA)      & Julia  & $42$; $12345{+}c$ per chain \\
Multifamily scaling                  & Julia  & per-replicate seed \\
Baseline evaluation and bootstrap    & Julia  & $12345$ \\
Permuted-alignment control           & Julia  & $54321$; $12345$ \\
PCA sensitivity                      & Julia  & $42$; per-condition seed \\
$K{=}20$ specificity                 & Julia  & $20001$ \\
Potts (plmDCA) baseline              & Python & $42$ \\
MSA Transformer baseline             & Python & $42$ \\
AlphaFold2 via ColabFold             & Python & $0$ \\
DMS position-matched null            & Python & $0$ \\
\bottomrule
\end{tabular}
\end{table}

\end{document}